\documentclass[a4paper,fleqn]{cas-dc}
\usepackage[numbers]{natbib}
\definecolor{CornflowerBlue}{RGB}{100,149,237} 
\definecolor{RoyalBlue}{RGB}{65,105,225}       
\definecolor{TealBlue}{RGB}{0,128,128}         
\definecolor{Green}{RGB}{0,128,0}
\definecolor{Black}{rgb}{0,0,0}
\definecolor{COLOR_MEAN}{RGB}{240, 240, 240}
\definecolor{LIGHT_YELLOW}{RGB}{241, 245, 138}
\usepackage{tikz}
\usepackage[edges]{forest}
\definecolor{hidden-draw}{RGB}{20,68,106}
\definecolor{hidden-pink}{RGB}{255,245,247}
\newcommand{\dec}[1]{{\small \color[HTML]{EA4335} {(-#1)}}}
\newcommand{\imp}[1]{{\small \color[HTML]{34A853} {(+#1)}}}
\usepackage{nicefrac}       
\usepackage{microtype}      
\usepackage{multirow}
\usepackage{multicol}
\usepackage{graphicx}
\usepackage{amssymb}
\usepackage{bbding}
\usepackage{color,colortbl}
\usepackage[export]{adjustbox}
\usepackage{booktabs}
\usepackage{amsmath}
\usepackage{float}
\usepackage{xcolor}
\usepackage{array}
\usepackage{booktabs}
\definecolor{myorange}{RGB}{251, 229, 214}
\definecolor{lightred}{RGB}{251,49,153}
\definecolor{skyblue}{RGB}{219, 240, 248}
\definecolor{mygreen}{RGB}{239, 254, 235}
\usepackage{tikz}
\usepackage{colortbl} 
\usepackage{geometry} 
\usepackage{bm} 
\usepackage{epsfig}
\usepackage{listings}
\usepackage{caption}
\usepackage{amsmath}
\usepackage{multirow}
\usepackage{array}
\usepackage{pifont}
\usepackage{wrapfig}
\usepackage{hhline}
\usepackage{color, colortbl}
\usepackage{bm}
\usepackage{adjustbox}
\usepackage{arydshln}
\usepackage{makecell}
\usepackage{amssymb}
\usepackage{subfiles}
\usepackage{float}

\usepackage{rotating}
\usepackage{enumitem}
\usepackage{utfsym}

\definecolor{improvement}{RGB}{225,97,78}
\definecolor{demphcolor1}{gray}{.6}
\definecolor{citecolor}{HTML}{0071bc}
\definecolor{tabhighlight}{HTML}{e5e5e5}

\newcommand{\tablestyle}[2]{\setlength{\tabcolsep}{#1}\renewcommand{\arraystretch}{#2}\centering\footnotesize}
\newlength\savewidth\newcommand\shline{\noalign{\global\savewidth\arrayrulewidth
		\global\arrayrulewidth .8pt}\hline\noalign{\global\arrayrulewidth\savewidth}}

\definecolor{COLOR_MEAN}{HTML}{f0f0f0}
\definecolor{LIGHT_BLUE}{HTML}{cce4fe}
\definecolor{LIGHT_RED}{HTML}{f1b9b8}
\definecolor{LIGHT_GREEN}{HTML}{006400}
\definecolor{LIGHT_YELLOW}{HTML}{f1f58a}

\newif\ifshowcomment
\showcommentfalse

\usepackage{hyperref}
\usepackage{url}

\usepackage{lineno}
\usepackage{amsmath,amssymb,amsfonts}
\usepackage{textcomp}
\usepackage{color}
\usepackage {hyperref}

\usepackage{amssymb}
\usepackage{amsmath}
\usepackage{mathtools}
\usepackage{amsthm}

\usepackage{pifont}       
\usepackage{fontawesome}  

\usepackage{ragged2e} 
\usepackage{booktabs,makecell, multirow, tabularx}
\usepackage{algorithm}  
\usepackage{algorithmicx}  
\usepackage{algpseudocode}

\newcommand{\colorbibs}[2][blue]%
{%
	\DeclareBibliographyCategory{ColoredBiblist#1}%
	\addtocategory{ColoredBiblist#1}{#2}%
	\AtEveryBibitem{\ifcategory{ColoredBiblist#1}{\color{#1}\bfseries}{}}
}

\begin{document}
\def\floatpagepagefraction{1}
\def\textpagefraction{.01}
\shorttitle{Wei Ai, et al, Computer Science Review}
\shortauthors{Wei Ai, et~al.}



\title[mode = title]{The Paradigm Shift: A Comprehensive Survey on Large Vision Language Models for Multimodal Fake News Detection}

\author[1]{Wei Ai}
\author[1]{Yilong Tan}
\author[1]{Yuntao Shou}
\author[1]{Tao Meng}
\cormark[1]
\author[2]{Haowen Chen}
\author[3]{Zhixiong He}

\author[4]{Keqin Li}

\cortext[1]{Corresponding author}

\address[1]{organization={College of Computer and Mathematics, Central South University of Forestry and Technology},
	postcode={410004},
	city={ Hunan, Changsha},
	country={China}}

\address[2]{organization={College of Computer Science and Electronic Engineering, Hunan University},
	postcode={410004},
	city={Hunan, Changsha},
	country={China}}

\address[3]{organization={College of Economics and Management, Central South University of Forestry and Technology},
	postcode={410004},
	city={ Hunan, Changsha},
	country={China}}

\address[4]{organization={Department of Computer Science, State University of New York}, 
	city={New Paltz, New York},
	postcode={12561},
	country={USA}}

%
%
%


%
%

\begin{abstract}
In recent years, the rapid evolution of large vision-language models (LVLMs) has driven a paradigm shift in multimodal fake news detection (MFND), transforming it from traditional feature-engineering approaches to unified, end-to-end multimodal reasoning frameworks. Early methods primarily relied on shallow fusion techniques to capture correlations between text and images, but they struggled with high-level semantic understanding and complex cross-modal interactions. The emergence of LVLMs has fundamentally changed this landscape by enabling joint modeling of vision and language with powerful representation learning, thereby enhancing the ability to detect misinformation that leverages both textual narratives and visual content. Despite these advances, the field lacks a systematic survey that traces this transition and consolidates recent developments. To address this gap, this paper provides a comprehensive review of MFND through the lens of LVLMs. We first present a historical perspective, mapping the evolution from conventional multimodal detection pipelines to foundation model-driven paradigms. Next, we establish a structured taxonomy covering model architectures, datasets, and performance benchmarks. Furthermore, we analyze the remaining technical challenges, including interpretability, temporal reasoning, and domain generalization. Finally, we outline future research directions to guide the next stage of this paradigm shift. To the best of our knowledge, this is the first comprehensive survey to systematically document and analyze the transformative role of LVLMs in combating multimodal fake news. The summary of existing methods mentioned is in our Github: \href{https://github.com/Tan-YiLong/Overview-of-Fake-News-Detection}{https://github.com/Tan-YiLong/Overview-of-Fake-News-Detection}.
\end{abstract}


\maketitle

\section{Introduction}

The rapid proliferation of fake news across online platforms has emerged as a formidable societal challenge \cite{wang2024explainable, shou2022conversational, 1011453786343, meng2024deep}, undermining public trust \cite{mohseni2021machine, meng2024multi, shou2024adversarial, shou2022object}, destabilizing democratic discourse, and exacerbating global crises such as the COVID-19 pandemic and geopolitical conflicts \cite{qian2018neural, varshini2023rdgt, shou2025masked, shou2025spegcl, meng2024masked}. Unlike early waves of misinformation, which were predominantly textual, modern fake news campaigns increasingly exploit multimodal content combining manipulated images \cite{rossler2019faceforensics++, shou2024efficient, shou2025contrastive, shou2026graph, shou2025graph}, misleading videos \cite{abdali2024multi, shou2025revisiting, shou2023graphunet}, and textually coherent yet semantically deceptive captions \cite{aneja2021cosmos, shou2025dynamic}. This growing reliance on cross-modal deception renders traditional unimodal detection techniques inadequate and motivates a shift toward Multimodal Fake News Detection (MFND) \cite{liu2023robust, hosseini2023interpretable, shou2025graph, shou2025multimodal}.

Multimodal fake news detection seeks to uncover inconsistencies between text, image, video, and other modalities to assess the veracity of online information \cite{shu2019beyond, shou2025gsdnet, ai2025revisiting}. However, the complex interplay between modalities, such as subtle semantic mismatches \cite{shang2025semantic, shou2025cilf, ai2024gcn, shou2024low}, visual entailment contradictions, and temporal or contextual incoherence, presents formidable challenges \cite{wu2023mfir, shou2025contrastive, ai2023two}. Early MFND methods typically relied on late fusion architectures or shallow cross-modal alignment, which lacked the semantic depth and reasoning capacity to capture nuanced deception strategies \cite{xie2025towards}.

Importantly, the notion of cross modal inconsistency in MFND is not ad hoc, but rooted in well documented limitations of vision language pretraining (VLP) models. Prior work shows that even large-scale VLP models exhibit systematic weaknesses in fine grained semantic alignment, particularly for linguistic negation, attribute ownership, and spatial relations \cite{wang2024can}. These limitations are especially consequential for fake news detection, where deceptive content often maintains surface level coherence while violating latent semantic constraints across modalities. As demonstrated by Wang et al. \cite{wang2024can}, such inconsistencies cannot be reliably identified through naive cross-modal similarity alone, revealing the brittleness of multimodal alignment without explicit linguistic grounding. This observation provides a theoretical basis for MFND, as many forms of multimodal misinformation intentionally exploit these alignment blind spots.

\begin{figure*}
	\centering
	\includegraphics[width=1\linewidth]{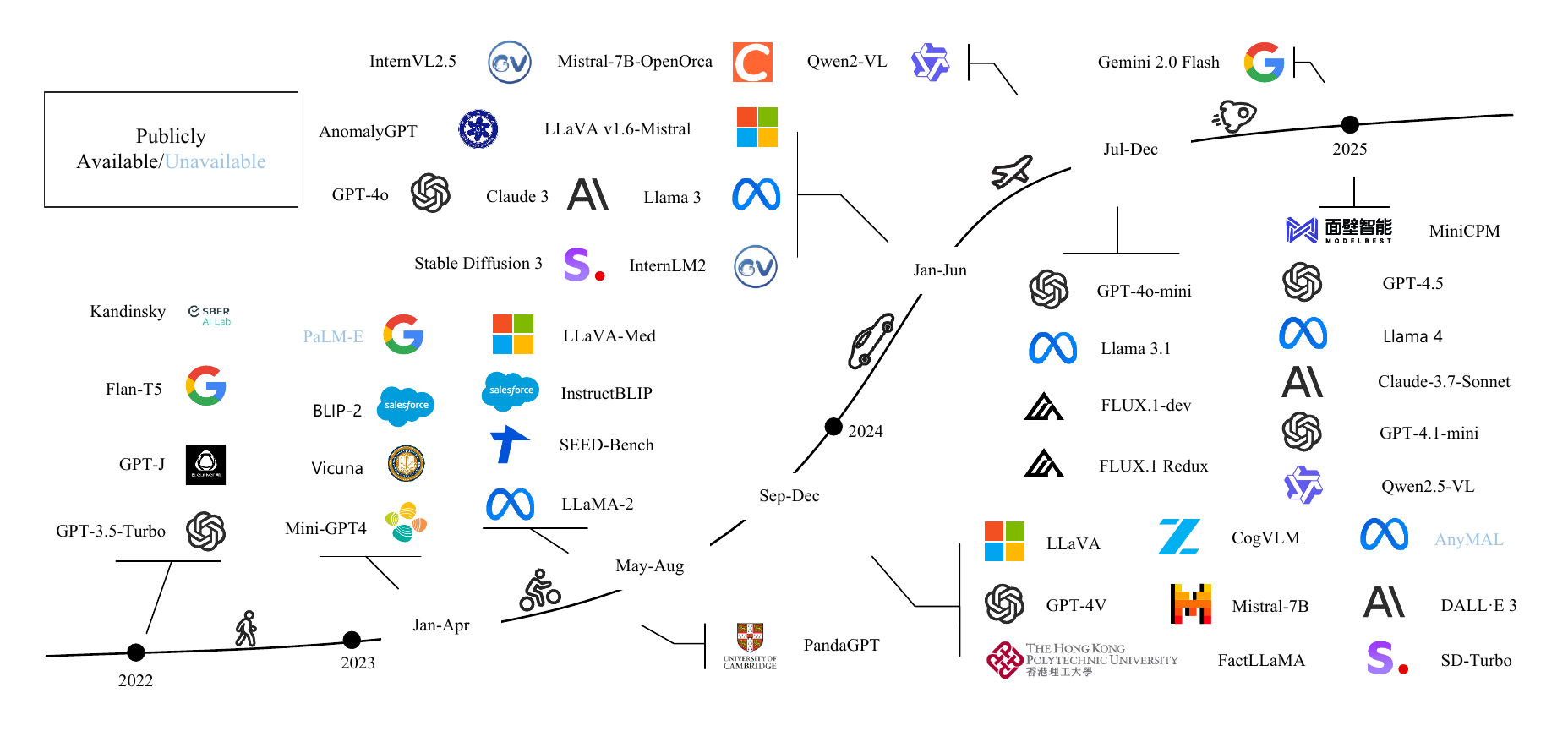}
	\caption{A chronological overview of representative LVLMs is presented, highlighting the rapid growth of this
		field.}
	\label{fig:LLM_his}
\end{figure*}

The emergence of Large Vision-Language Models (LVLMs) has fundamentally changed this landscape \cite{kuntur2024under}. Built upon advances in large language models (LLMs) and pre-trained visual encoders \cite{liu2025modality, wu2024fake}, as shown in Fig. \ref{fig:LLM_his}, LVLMs such as CLIP \cite{zhou2023multimodal}, BLIP-2 \cite{wang2024multimodal}, Flamingo \cite{alayrac2022flamingo}, Kosmos-1 \cite{huang2023language}, LLaVA \cite{liu2024improved}, and GPT-4V \cite{lyu2025gpt} offer unified architectures capable of joint cross-modal representation learning and reasoning \cite{chen2022cross}. These models demonstrate impressive zero-shot and few-shot performance on tasks such as image-text matching \cite{hu2022causal}, visual question answering \cite{gao2024knowledge}, and multimodal entailment making them well-suited for high-level fake news detection that demands semantic grounding and cross-modal verification \cite{peng2024not}.

Nevertheless, the application of LVLMs to MFND remains fragmented and lacks a systematic understanding. Existing studies vary widely in how they integrate LVLMs, differing in architectural design, supervision strategy \cite{li2025learning}, task formulation \cite{zhang2024reinforced}, and training cost \cite{zhang2025knowledge}. To unify this landscape, we propose a novel three branch taxonomy that categorizes current MFND approaches using LVLMs into three distinct paradigms:  1) \textbf{Parameter Freezing Applications.} These methods utilize pre-trained LVLMs without modifying their internal parameters \cite{zhang2024natural}. Techniques such as in-context learning \cite{qi2024sniffer}, prompt-based adaptation \cite{hu2024multi}, or lightweight classification heads are employed on top of frozen backbones \cite{xie2025towards}. This paradigm is particularly attractive for resource constrained or real-time applications, offering fast deployment and high generalization at the cost of task specific adaptation \cite{tong2024mmdfnd}.  2) \textbf{Parameter Tuning Applications.} These approaches involve full or partial fine tuning of the LVLMs to enhance task alignment \cite{shu2019beyond}. Techniques include full model fine tuning \cite{yin2025graph}, adapter insertion \cite{wang2025fakesv}, prefix tuning \cite{xu2025ample}, or low rank adaptation (e.g., LoRA) \cite{hulora}. Tuning allows the model to capture domain specific semantics \cite{silva2021embracing}, subtle modality inconsistencies \cite{niu2025exploring}, and contextual cues critical to accurate fake news detection, albeit with higher computational and data requirements \cite{abulaish2022domain}.  3) \textbf{Reasoning Paradigm Applications.} This new paradigm focuses on how LVLMs engage in structured, multi step reasoning to enhance the accuracy of fake news detection. Approaches under this paradigm emphasize the use of multi agent systems or explicit reasoning pipelines that sequentially analyze multimodal evidence and the claim. Methods such as agent-based reasoning \cite{li2024large}, which breaks down reasoning into distinct sub-tasks like evidence retrieval and contradiction analysis, and prompting-based reasoning \cite{wang2024mitigating}, which guides the model through reasoning via carefully crafted prompts, are key examples. The reasoning paradigm enables a deeper understanding of complex misinformation, leveraging the strengths of both large pre-trained models and task-specific reasoning strategies. However, it typically incurs higher inference costs and complexity, requiring more elaborate model architectures and robust reasoning pipelines \cite{he2025factguard}.

In this survey, we offer the first comprehensive and structured review of how large vision-language models are being utilized for multimodal fake news detection. The contributions made in this paper are summarized as follows:

\begin{itemize}

	\item \textbf{New Taxonomy:} We propose a novel three-branch taxonomy for applying large vision language models (LVLMs) to multimodal fake news detection (MFND), dividing existing methods into three main categories: (1) parameter-freezing applications, (2) parameter-tuning applications, and (3) reasoning-paradigm applications. This is the first comprehensive survey to systematically document and analyze the transformative role of LVLMs in combating multimodal fake news, incorporating the latest advancements in structured reasoning and agent-based strategies.

	\item \textbf{Comprehensive Review:} We systematically review the evolution from unimodal to multimodal approaches, analyze representative LVLMs-based architectures, and compare their performance and design principles across visual-textual misinformation detection benchmarks.

	\item \textbf{Unified Framework:} We build a structured analytical framework that encompasses model design, task formulation, training paradigms, and evaluation criteria. By aligning the strengths and limitations of parameter-freezing, parameter-tuning, and reasoning-based approaches, we provide guidance for selecting appropriate LVLM strategies under different resource and robustness constraints, helping to navigate the challenges of real-time and resource-constrained applications.

	\item \textbf{Benchmark and Evaluation:} We summarize and compare publicly available datasets in MFND, highlighting their modality composition, annotation strategy, and real-world complexity. In addition, we examine current evaluation metrics and identify key gaps.
	
	\item \textbf{Research Challenges and Future Directions:} We identify open problems in LVLM-based MFND, including multimodal hallucination, cross-modal bias, and data scarcity in misinformation domains. We further propose promising directions for future research, including counterfactual generation for cross-modal reasoning, and knowledge-enhanced LVLM adaptation.
\end{itemize}

This paper is organized as follows: Section 2 introduces the background and preliminary knowledge. Section 3 outlines the multimodal fake news detection paradigm based on LVLM. Section 4 introduces widely used benchmark datasets. Section 5 reviews evaluation metrics. Section 6 compares the experimental results of different methods. Section 7 discusses the advantages and challenges of LVLM in this field. Section 8 points out future research directions. Section 9 concludes the paper.

\section{Preliminary Information}

\subsection{Traditional Single-Modal Fake News Detection}

\textbf{Problem Definition.} Early approaches to fake news detection predominantly relied on a single modality \cite{wang2018eann, cui2019same}, typically focusing on either textual or visual content \cite{khattar2019mvae, singhal2019spotfake}. In these methods, fake news is formulated as a binary classification problem over a single input space \cite{kaliyar2020fndnet}. Let $x_t$ denote textual features extracted from news articles \cite{jing2023multimodal}, and $x_v$ represent visual features derived from associated images \cite{qian2021knowledge}. The single-modal fake news classifier can be generally defined as $\mathcal{F}_{sm}: x \rightarrow \{0,1\}$, where $x \in \{x_t, x_v\}$ and the output $\{0,1\}$ corresponds to the labels real and fake \cite{perez2018automatic}.

\textbf{Text-based methods.}  
Traditional text-based approaches typically leverage linguistic and semantic features \cite{verma2021welfake}, such as word $n$-grams, syntactic patterns \cite{shu2019defend}, or semantic embeddings \cite{wang2024fake}. Given a text sequence $T = \{w_1, w_2, \ldots, w_n\}$, an embedding model $\phi(\cdot)$ maps words into a continuous space \cite{ruchansky2017csi}, yielding the representation as follows:
\begin{equation}
X_T = \phi(T) \in \mathbb{R}^{d}
\end{equation}
where $d$ is the embedding dimension. A classifier $f_\theta$ (e.g., logistic regression, SVM, or neural network) then predicts the probability of being fake as follows:
\begin{equation}
P(y=1|T) = \sigma(f_\theta(X_T))
\end{equation}
with $\sigma(\cdot)$ denoting the sigmoid function. The model is trained using the standard binary cross-entropy loss as follows:
\begin{equation}
\mathcal{L}_{text} = - \sum_{i=1}^N \big[ y_i \log P(y_i|T_i) + (1-y_i)\log(1-P(y_i|T_i)) \big]
\end{equation}

\textbf{Image-based methods.}  
Visual single-modal methods focus on detecting inconsistencies or manipulations in images \cite{xue2021detecting, wang2023attacking}. A typical approach involves extracting features from an image $I$ using a convolutional neural network (CNN) \cite{qian2018neural} or vision Transformer (ViT) \cite{yuan2021tokens} as follows:
\begin{equation}
X_V = \psi(I) \in \mathbb{R}^{k}
\end{equation}
where $\psi(\cdot)$ is the feature extractor. The classifier $g_\theta$ then computes as follows:
\begin{equation}
P(y=1|I) = \sigma(g_\theta(X_V))
\end{equation}
These methods aim to capture visual artifacts such as unnatural textures \cite{liu2020global}, illumination mismatches \cite{guo2020future}, or traces of generative adversarial networks (GANs) \cite{ma2019detect}.

\textbf{Limitations.}  
Although single-modal detection has achieved early success, it suffers from several critical limitations \cite{zhou2020survey}. Text-only methods often fail when the news content appears linguistically coherent but is paired with misleading images \cite{wang2017liar}. Similarly, image-only methods cannot account for manipulative narratives conveyed by text \cite{kou2025potential}. Formally, if fake news involves a cross-modal inconsistency \cite{shou2024spegcl,hiriyannaiah2020computationally}, i.e.,
\[
(T, V) \in \mathcal{D}_{fake}, \quad \phi(T) \not\approx \psi(V)
\]
then any function $\mathcal{F}_{sm}$ defined on a single modality is inherently insufficient \cite{yu2017convolutional}. This limitation motivates the shift toward multimodal approaches, where joint reasoning over text and vision becomes essential \cite{yang2022reinforcement}.

\subsection{Multimodal Fake News Detection}

\textbf{Problem Definition.} Let $\mathcal{M}=\{t,v\}$ denote the set of available modalities (e.g., text $t$, image/video $v$). A multimodal set is $x=\{x_m\}_{m\in\mathcal{M}}$, associated with a focal claim $c$ and optional external context $\kappa$ (e.g., knowledge bases, timelines, provenance) \cite{dun2021kan}. The goal of Multimodal Fake News Detection (MFND) is to infer a veracity label $y\in\{0,1\}$ (with $1$ meaning fake) together with a structured, referenceable rationale $R$ grounding the decision in cross-modal evidence \cite{li2025survey, feng2025contradicted}. We formalize a reasoned prediction as follows:
\begin{align}
	(\hat{y}, \hat{R})
	&= \arg\max_{y \in \{0,1\},\, R \in \mathcal{R}}
	P_\theta\!\big(y, R \mid x, c, \kappa\big)
	\label{eq:reasoned-pred}
\end{align}
where $\theta$ parameterizes the large vision language model (LVLM), and $\mathcal{R}$ denotes the space of executable explanations (textual justifications with pointers to spans, frames, and regions).

\subsection{Multimodal Fake News Reasoning}

\textbf{Principled Dimensions of Deception.} We posit a principled, triadic decomposition of multimodal deception into three interacting yet distinct dimensions (i.e., Authenticity, Consistency, and Intent) that together determine veracity and the structure of model explanations \cite{hu2024bad}. Authenticity quantifies media-level integrity via forensic evidence and generative priors \cite{yang2019unsupervised}. Consistency measures cross-modal semantic agreement between claim, text, image/video, and audio embeddings \cite{xue2021detecting}. Intent captures manipulative or persuasive framing, decoupled from factuality, through stylistic, pragmatic, and source cues. These dimensions target complementary failure modes, including (i) deepfaked or altered media with otherwise aligned captions; (ii) truthful media paired with misleading text; and (iii) technically accurate content arranged to steer beliefs. Relying on any single dimension is insufficient, instead, we fuse calibrated posteriors within an interpretable surrogate to produce both the veracity label and a grounded rationale \cite{ngueajio2025decoding}. This decomposition provides a unifying lens across news articles, social posts, memes, and videos, and it aligns naturally with LVLM-based reasoning that constructs evidence graphs and referenceable explanations \cite{yang2019xfake}. Below we will introduce the principles of each dimension.

\textbf{Media Authenticity.} This dimension targets whether a medium in $x$ has been synthetically generated or altered. Let $p_{\mathrm{nat}}$ and $p_{\mathrm{alt}}$ be (implicit) likelihoods under natural and altered media manifolds, respectively (instantiated by generative/forensic priors) \cite{huh2018fighting}. A generic forgery score for a visual stream $x_v$ is as follows:
\begin{equation}
\begin{aligned}
	S_F(x_v)
	&= -\log p_{\mathrm{nat}}(x_v) + \log p_{\mathrm{alt}}(x_v) \\
	F
	&= \sigma\!\big(\phi_F(S_F(x_v))\big) \in [0,1]
	\label{eq:forgery-prob}
\end{aligned}
\end{equation}
where $\phi_F$ calibrates the score and $\sigma$ is a sigmoid. $F$ indicates tampering signals (e.g., seam/lighting inconsistencies, GAN fingerprints, lip-sync drift).

\textbf{Cross-Modal Inconsistency.} This dimension captures contradictions between modalities (e.g., the caption asserts ``Category 5 hurricane'' while the video depicts calm weather). Let $g_m:\mathcal{X}_m\to\mathbb{R}^d$ embed each modality into a shared semantic space \cite{wu2023mfir}. For any pair $(m,n) \in \mathcal{M}$:
\begin{align}
	D_{m,n}(x)
	&= 1 - \mathrm{sim}\!\big(g_m(x_m),\, g_n(x_n)\big)
	\label{eq:pairwise-inconsistency}\\
	D
	&= \max_{(m,n)\in \mathcal{M}} D_{m,n}(x) \in [0,1]
	\label{eq:global-inconsistency}
\end{align}
where $\mathrm{sim}$ can be cosine similarity or an entailment-based score from a caption$\to$NLI pipeline. Large $D$ signals cross-modal contradiction or missing supporting evidence \cite{xu2022evidence}.

\textbf{Manipulative Intent.} Distinct from factuality, this dimension evaluates whether content is crafted to steer beliefs via rhetorical devices (e.g., cherry-picked visuals, loaded language, fear appeals) \cite{zhou2020survey}. We model intent as a latent probability as follows:
\begin{equation}
\begin{aligned}
	I
	&= P_\theta\!\big(\text{manipulative}\mid x,c,\kappa\big) \\
	& = \sigma\!\Big(\phi_I\big(\psi_t(x_t),\, \psi_v(x_v),\, \psi_a(x_a),\, u\big)\Big) \in [0,1]
\end{aligned}
\label{eq:intent}
\end{equation}
where $\psi_\bullet$ extract stance, sentiment, emotional arousal, propaganda patterns, and $u$ encodes source-level priors.

\textbf{Joint Decision.} The veracity decision should couple these dimension while remaining explanation-seeking \cite{jin2022towards}. A factorized formulation is as follows:
\begin{equation}
	\begin{aligned}
	& P_\theta(y,F,D,I \mid x,c,\kappa)\propto \psi_Y(y\mid x,c,\kappa)\psi_F(F\mid x) \\
	&\psi_D(D\mid x,c)
	\psi_I(I\mid x,c,\kappa)
	\psi_C(y,F,D,I)
\end{aligned}
\label{eq:factor-graph}
\end{equation}
where $\psi_C$ enforces structural compatibilities (e.g., severe tampering or strong cross-modal contradictions raise the prior for $y{=}1$, while high $I$ alone is not conclusive) \cite{chen2022cross}. A transparent decision surrogate is as follows:
\begin{equation}
	\begin{aligned}
	s(x,c,\kappa)
	&= h_\theta(x,c,\kappa) + \lambda_F F + \lambda_D D + \lambda_I I\\
	\hat{y}
	&= \mathbb{1}\!\big[\sigma(s(x,c,\kappa)) > \tau\big]
	\label{eq:surrogate-decision}
	\end{aligned}
\end{equation}
where $h_\theta$ capturing claim-conditioning and external verification (e.g., retrieval-augmented grounding), and $\lambda_\bullet$ balancing the three dimension.

\textbf{Definition of Reasoning in MFND.} Reasoning is the construction of a \emph{grounded evidence graph} $G=(V,E)$ and a minimal rationale $R\subseteq V$ such that a verifier $V_{\mathrm{check}}$ confirms sufficiency \cite{jin2022towards} and faithfulness as follows:
\begin{equation}
	\begin{aligned}
	\underbrace{P_\theta\!\big(y \mid x,c,\kappa,R\big)}_{\text{sufficient}}
	-
	\underbrace{P_\theta\!\big(y \mid x,c,\kappa,\varnothing\big)}_{\text{baseline}}
	&\ge \epsilon\\
	\underbrace{P\!\big(R \text{ is used by } \theta\big)}_{\text{faithful}}
	&\ge 1 - \delta
	\end{aligned}
\label{eq:faithfulness}
\end{equation}
where $R$ consists of modality-aligned snippets (text spans, frames, regions) and explicit cross-modal relations (support/contradiction) that can be referenced and checked \cite{liu2023robust}.

\textbf{Learning Objective.} Supervision is naturally multi-task, promoting veracity accuracy, forensic robustness, semantic alignment, and intent recognition \cite{choudhary2021linguistic}, while rewarding concise, grounded explanations as follows:
\begin{equation}
\begin{aligned}
	\mathcal{L}
	&= \underbrace{\ell_{\mathrm{cls}}(y,\hat{y})}_{\text{veracity}}
	+ \alpha\,\underbrace{\ell_F\!\big(F,\hat{F}\big)}_{\text{authenticity}}
	+ \beta\,\underbrace{\ell_D\!\big(D,\hat{D}\big)}_{\text{consistency}} \\
	&+ \gamma\,\underbrace{\ell_I\!\big(I,\hat{I}\big)}_{\text{intent}} 
	+ \eta\,\underbrace{\ell_{\mathrm{exp}}(R,\hat{R})}_{\text{grounded rationale}}
\end{aligned}
\label{eq:loss}
\end{equation}
where $\alpha$ controls the relative importance of the content-level forgery detection loss $\ell_F$, which focuses on identifying tampered or synthetic media, $\beta$ weights the cross-modal consistency loss $\ell_D$, encouraging the model to detect semantic misalignments between modalities \cite{shen2025llm}, $\gamma$ adjusts the manipulative intent recognition loss $\ell_I$, which measures whether the content is strategically framed to mislead or persuade, $\eta$ regulates the explanation generation loss $\ell_{\mathrm{exp}}$, ensuring that the model outputs interpretable and faithful rationales $R$ grounded in multimodal evidence \cite{si2023exploring}.

\subsection{Positioning of Large Vision-Language Models}
\label{sec:positioning-lvlm}

\textbf{Unified Parameterization.} LVLMs provide a single probabilistic program to instantiate the factorization in Eq.~\eqref{eq:factor-graph} and the reasoned prediction \cite{wu2024unified} in Eq.~\eqref{eq:reasoned-pred}.
Let $\phi_t,\phi_v$ be modality encoders and let $r_\kappa(c)$ denote a retrieval operator over external context $\kappa$ conditioned on claim $c$.
We define claim-aware token streams
\begin{equation}
\tilde{t}=\text{Pack}(x_t,c),\quad
\tilde{v}=\text{Patch}(x_v),\quad
\tilde{\kappa}=r_\kappa(c)
\end{equation}
and a cross-modal aggregator $\mathcal{A}_\theta$ (multi-head cross-attention over concatenated tokens) that yields a joint representation as follows:
\begin{equation}
	Z=\mathcal{A}_\theta\Big(\phi_t(\tilde{t}),\,\phi_v(\tilde{v}),\,\phi_t(\tilde{\kappa})\Big)\in\mathbb{R}^d
	\label{eq:joint-Z}
\end{equation}

Concretely, the authenticity, consistency \cite{wu2023human}, and intent scores are realized by LVLM heads, which are mathematically expressed as follows:
\begin{equation}
\begin{aligned}
	&\hat{F} = \sigma\!\big(w_F^\top Z+b_F\big), \quad
	\hat{I} = \sigma\!\big(w_I^\top Z+b_I\big) \\
	&g_t(x_t) = W_t \phi_t(\tilde{t}),\quad g_v(x_v)=W_v \phi_v(\tilde{v}) \\
	&\hat{D} =1-\mathrm{sim}\!\big(g_t(x_t),g_v(x_v)\big)
	\label{eq:D-head}
\end{aligned}
\end{equation}

The claim-conditioned verifier $h_\theta$ is implemented to fuse content evidence with retrieved context as follows:
\begin{equation}
	h_\theta(x,c,\kappa)=w_Y^\top Z+b_Y
	\label{eq:h-theta}
\end{equation}

\textbf{Cross-modal Alignment Pretraining.}
To make $\mathrm{sim}(\cdot,\cdot)$ meaningful, LVLMs optimize contrastive alignment \cite{zheng2025predictions} between $g_t$ and $g_v$:
\begin{equation}
	\mathcal{L}_{\mathrm{align}}
	= -\sum_{i=1}^N \log
	\frac{\exp\big(\langle g_t(x_t^{(i)}),g_v(x_v^{(i)})\rangle/\tau\big)}
	{\sum_{j=1}^N \exp\big(\langle g_t(x_t^{(i)}),g_v(x_v^{(j)})\rangle/\tau\big)}
	\label{eq:align}
\end{equation}
This objective ensures that $D$ faithfully reflects cross-modal consistency.

\textbf{Claim-conditioned Verification and Retrieval Grounding.}
External context $\kappa$ is integrated by a differentiable retriever as follows:
\begin{equation}
	r_\kappa(c)=\mathrm{Top}\!K\big(\arg\max_{d\in \kappa}\;\mathrm{sim}(q(c),k(d))\big)
	\label{eq:retriever}
\end{equation}
with query encoder $q(\cdot)$ and key encoder $k(\cdot)$ sharing parameters with $\phi_t$.
The retrieved snippets are injected as memory tokens, allowing $h_\theta$ to realize the claim-conditioning term and to reduce spurious correlations in $\hat{F},\hat{D},\hat{I}$.

\textbf{Evidence-consistent Rationale Generation.}
Let $R=(r_1,\ldots,r_L)$ denote a textual rationale augmented \cite{zhang2024reinforced} with pointers $\pi$ to evidence units (text spans, frames, regions) from the evidence graph $G=(V,E)$. LVLMs generate $R$ with an autoregressive head over $Z$ as follows:
\begin{equation}
\begin{aligned}
	P_\theta(R\mid x,c,\kappa)
	&=\prod_{t=1}^{L} P_\theta(r_t\mid r_{<t},x,c,\kappa), \\
	P_\theta(\pi_t=u\mid r_{\le t},x,c,\kappa)
	&=\mathrm{softmax}\big(\langle q_t,\, \nu(u)\rangle\big),\; u\in V
	\label{eq:pointer}
\end{aligned}
\end{equation}
where $q_t$ is the decoder query at step $t$ and $\nu(u)$ is the node embedding of evidence unit $u$.
By training Eqs.~\eqref{eq:pointer} jointly, the LVLM yields rationales that satisfy the sufficiency/faithfulness criteria via explicit grounding \cite{shao2023detecting}.

\textbf{Decision Surrogate and Calibration.}
Plugging LVLM heads into Eq.~\eqref{eq:surrogate-decision} gives as follows:
\begin{equation}
	\begin{aligned}
	s(x,c,\kappa) & =h_\theta(x,c,\kappa)+\lambda_F \hat{F}+\lambda_D \hat{D}+\lambda_I \hat{I}\\
	\hat{y} & =\mathbb{1}\big[\sigma(s)>\tau\big]
	\label{eq:lvlm-surrogate}
	\end{aligned}
\end{equation}
where the weights $(\lambda_F,\lambda_D,\lambda_I)$ are learned by post-hoc calibration to respect the compatibility potential $\psi_C$ in Eq.~\eqref{eq:factor-graph}:
\begin{equation}
	\min_{\lambda_F,\lambda_D,\lambda_I,\tau}\;
	\mathbb{E}\!\left[\ell_{\mathrm{cls}}\!\big(y,\mathbb{1}[\sigma(s)>\tau]\big)\right]
	+\Omega\!\big(\lambda_F,\lambda_D,\lambda_I\big),
	\label{eq:calib}
\end{equation}
with a regularizer $\Omega$ that discourages over-reliance on any single dimension.

\textbf{Media Authenticity with Generative Priors.}
To instantiate $p_{\mathrm{nat}}$ and $p_{\mathrm{alt}}$, LVLMs either distill forensic experts into a head over $Z$, or use a learned energy surrogate \cite{yang2019unsupervised} as follows:
\begin{equation}
	\begin{aligned}
	S_F(x_v)\approx E_\theta(x_v) & =
	-\log \sum_{z} \exp\!\big(-\mathcal{E}_\theta(x_v,z)\big)\\
	\hat{F} & =\sigma\!\big(\phi_F(E_\theta(x_v))\big)
	\label{eq:energy-forgery}
	\end{aligned}
\end{equation}
where $\mathcal{E}_\theta$ is an amortized energy over latent codes $z$ (e.g., visual artifacts, frequency cues). This provides a practical bridge to Eq.~\eqref{eq:forgery-prob} within the LVLM.

\textbf{Training objective.}
The multi-task learning is realized by augmenting it with alignment and instruction-following terms \cite{wang2020weak} as follows:
\begin{equation}
	\begin{aligned}
		\mathcal{L}_{\mathrm{LVLM}}
		&=\underbrace{\ell_{\mathrm{cls}}(y,\hat{y})}_{\text{veracity}}
		+\alpha\,\underbrace{\ell_F(F,\hat{F})}_{\text{authenticity}}
		+\beta\,\underbrace{\ell_D(D,\hat{D})}_{\text{consistency}}
		+\gamma\,\underbrace{\ell_I(I,\hat{I})}_{\text{intent}}\\
		&\quad+\eta\,\underbrace{\ell_{\mathrm{exp}}(R,\hat{R})}_{\text{grounded rationale}}
		+\mathcal{L}_{\mathrm{align}}
		+\mathcal{L}_{\mathrm{LM}}
		+\mathcal{L}_{\mathrm{ptr}}
	\end{aligned}
	\label{eq:lvlm-loss}
\end{equation}
where $\mathcal{L}_{\mathrm{LM}}$ is the next-token negative log-likelihood for rationale generation and $\mathcal{L}_{\mathrm{ptr}}$ is a cross-entropy over pointers $\pi$ to evidence nodes \cite{li2022dynamic}.

\begin{figure*}[!htbp]
	\centering
	\resizebox{1\textwidth}{!}{
		\begin{forest}
			forked edges,
			for tree={
				grow=east,
				reversed=true,
				anchor=base west,
				parent anchor=east,
				child anchor=west,
				base=left,
				font=\tiny\color{Black!90},      
				rectangle,
				draw=RoyalBlue!70,                 
				fill=RoyalBlue!10,                 
				rounded corners=4pt,                
				align=left,
				minimum width=2em,
				edge+={CornflowerBlue!90, line width=1pt}, 
				s sep=20pt,
				l sep=12pt,
				inner xsep=3pt,
				inner ysep=4pt,
				line width=0.6pt,
				ver/.style={
					rotate=90,
					child anchor=north,
					parent anchor=south,
					anchor=center,
					fill=RoyalBlue!15,             
					draw=RoyalBlue!70,
				},
			},
			where level=0{                          
				fill=RoyalBlue!25,
				draw=RoyalBlue!90,
				thick,
				font=\tiny\bfseries\color{Black},
			}{},
			where level=1{                          
				text width=9em,
				font=\tiny\bfseries\color{Black!90},
				fill=CornflowerBlue!15,
				draw=CornflowerBlue!60,
			}{},
			where level=2{                          
				text width=28em,
				font=\tiny\color{Black!75},
				fill=RoyalBlue!5,
				draw=CornflowerBlue!40,
			}{},
			[
			Multimodal Fake News Detection and Veracity Reasoning, ver
			[
			Parameter-Frozen Paradigm (\S \ref{sec:Frozen})
			[
			\textbf{Zero-shot Learning:} 
			\textcolor{Green}{CLIP}~\cite{radford2021learning}{\char44}
			\textcolor{Green}{ALIGN}~\cite{jia2021scaling}{\char44}
			\textcolor{Green}{M$^3$A}~\cite{xu2024m3a}{\char44}
			\textcolor{Green}{BLIP-2}~\cite{li2023blip}{\char44}
			\textcolor{Green}{LLaVA}~\cite{liu2024improved}{\char44}
			\textcolor{Green}{InstructBLIP}~\cite{dai2023instructblip}{\char44}
			\textcolor{Green}{GPT-4V}~\cite{yang2023dawn}{\char44}\\
			\textcolor{Green}{FILIP}~\cite{yaofilip}{\char44}
			\textcolor{Green}{MCCA}~\cite{liu2018multiview}{\char44} 
			\textcolor{Green}{METER}~\cite{dou2022empirical}{\char44}
			\textcolor{Green}{UNITE-FND}~\cite{mukherjee2025unite}{\char44}
			\textcolor{Green}{Gemini-1.5}~\cite{team2024gemini}{\char44}
			\textcolor{Green}{MiniGPT-4}~\cite{zhu2023minigpt}{\char44}
			\textcolor{Green}{BLIP}~\cite{li2022blip}{\char44}
			\textcolor{Green}{KAI}~\cite{zhang2024knowledge}{\char44}\\
			\textcolor{Green}{GPT4o}~\cite{hurst2024gpt}{\char44} 
			\textcolor{Green}{FND-CLIP}~\cite{zhou2023multimodal}{\char44}
			\textcolor{Green}{Phi-3-Vision-128k-Instruct}~\cite{phi3vision2024}{\char44}
			\textcolor{Green}{LLaVA-v1.5-Vicuna-7B}~\cite{NEURIPS2023LLaVA}{\char44}
			\textcolor{Green}{Pixtral}~\cite{mistral2024pix}{\char44}\\
			\textcolor{Green}{LLaVA-v1.6-Mistral-7B}~\cite{NEURIPS2023LLaVA}{\char44}
			\textcolor{Green}{Qwen2-VL-7B-Instruct}~\cite{Qwen2VL}{\char44} 
			\textcolor{Green}{InternVL2-8B}~\cite{chen2024internvl}{\char44}
			\textcolor{Green}{DeepSeek-VL2-Small}~\cite{wu2024deepseek}{\char44}\\
			\textcolor{Green}{DeepSeek Janus-Pro-7B}~\cite{chen2025janus}{\char44} 
			\textcolor{Green}{GLM-4V-9B}~\cite{glm2024chatglm}{\char44}
			\textcolor{Green}{LLaMA-3.2-11B-Vision}~\cite{meta2024llama}{\char44}
			\textcolor{Green}{GPT 3.5 (Base)}~\cite{achiam2023gpt}{\char44}
			\textcolor{Green}{MRCD}~\cite{zhou2025collaborative}{\char44}\\ 
			\textcolor{Green}{GPT 3.5 (CoT) }~\cite{achiam2023gpt}{\char44} 
			\textcolor{Green}{GPT 4 (Base)}~\cite{achiam2023gpt}{\char44}
			\textcolor{Green}{GPT 4 (CoT)}~\cite{achiam2023gpt}{\char44} 
			\textcolor{Green}{Vicuna-7B-v1.5}~\cite{zheng2023judging}{\char44}
			\textcolor{Green}{InternLM2-7B}~\cite{cai2024internlm2}{\char44}\\
			\textcolor{Green}{Mistral-7B-Instruct-v0.3}~\cite{mistral20237b}{\char44} 
			\textcolor{Green}{Qwen2-7B-Instruct}~\cite{yang2024qwen2}{\char44}
			\textcolor{Green}{GLM-4-9B-Chat}~\cite{glm2024chatglm}{\char44}
			\textcolor{Green}{MiRAGe}~\cite{huang2024miragenews}{\char44}
			\textcolor{Green}{LEMMA}~\cite{xuan2024lemma}{\char44}\\
			\textcolor{Green}{Kosmos-1}~\cite{huang2023language}\\
			]
			[
			\textbf{Few-shot Learning:} 
			\textcolor{Green}{AAR}~\cite{zheng2025unveiling}{\char44} 
			\textcolor{Green}{SearchLVLMs}~\cite{li2024searchlvlms}{\char44} 
			\textcolor{Green}{DriftBench}~\cite{li2025drifting}{\char44}
			\textcolor{Green}{FKA-Owl}~\cite{liu2024fka}{\char44}  
			\textcolor{Green}{CMA}~\cite{jiang2025cross}{\char44}
			\textcolor{Green}{ARG-D}~\cite{hu2024bad}{\char44}\\
			\textcolor{Green}{FNDPT}~\cite{gao2023few}{\char44}
			\textcolor{Green}{Prompt-and-Align}~\cite{wu2023prompt}{\char44}
			\textcolor{Green}{COOL}~\cite{ouyang2024cool}{\char44}
			\textcolor{Green}{FSKD}~\cite{yuan2023fskd}{\char44}
			\textcolor{Green}{MPL}~\cite{hu2024multi}{\char44}
			\textcolor{Green}{DAFND}~\cite{gao2023few}{\char44}
			\textcolor{Green}{StablePT}~\cite{liu2024stablept}{\char44}\\
			\textcolor{Green}{FakeSV-VLM}~\cite{wang2025fakesv}
			]
			]
			[
			Parameter-Tuning Paradigm (\S \ref{sec:Tuning}) 
			[
			\textbf{Full-parameter Tuning:} 
			\textcolor{Green}{IMFND}~\cite{jiang2025imfnd}{\char44} 
			\textcolor{Green}{MDAM$^3$}~\cite{xuan2024lemma}{\char44} 
			\textcolor{Green}{DIFND}~\cite{yan2025debunk}{\char44}
			\textcolor{Green}{MCOT}~\cite{shen2024multimodal}{\char44}
			\textcolor{Green}{MACAW}~\cite{yindetecting}{\char44}  
			\textcolor{Green}{DPOD}~\cite{bhattacharya2025can}{\char44}  \\
			\textcolor{Green}{TRUST-VL}~\cite{yan2025trust}{\char44}
			\textcolor{Green}{EARAM}~\cite{zheng2025predictions}{\char44} 
			\textcolor{Green}{RumorLLM}~\cite{lai2024rumorllm} \\
			]
			[
			\textbf{Parameter-efficient Tuning:}
			\textcolor{Green}{E2LVLM}~\cite{wu2025e2lvlm}{\char44}
			\textcolor{Green}{LVLM4FV}~\cite{tahmasebi2024multimodal}{\char44}
			\textcolor{Green}{M-DRUM}~\cite{jin2024fake}{\char44}
			\textcolor{Green}{MMKD}~\cite{zeng2024multimodal}{\char44}
			\textcolor{Green}{LVLM4CEC}~\cite{tahmasebi2025verifying}{\char44}\\
			\textcolor{Green}{Cross-SEAN}~\cite{kumar2023optnet}{\char44}
			\textcolor{Green}{SNIFFER}~\cite{qi2024sniffer}{\char44}
			\textcolor{Green}{FakeNewsGPT4}~\cite{liu2024fakenewsgpt4}
			]
			]
			[
			Reasoning Paradigm (\S \ref{sec:Reasoning}) 
			[
			\textbf{Prompting-based Reasoning:} 
			\textcolor{Green}{ICD}~\cite{wang2024mitigating}{\char44} 
			\textcolor{Green}{CAPE-FND}~\cite{jin2025veracity}{\char44} 
			\textcolor{Green}{NRFE}~\cite{zhang2025llms}{\char44}
			\textcolor{Green}{IFAI}~\cite{Zhang2025ConfidenceBS}{\char44}
			\textcolor{Green}{DIFAR}~\cite{wan2025difar}{\char44}  
			\textcolor{Green}{MFC-Bench}~\cite{wang2025mfc}{\char44} \\
			\textcolor{Green}{LLM-GAN}~\cite{wang2025llm}{\char44}
			\textcolor{Green}{LIFE}~\cite{wang2025prompt} \\
			]
			[
			\textbf{Agent-based Reasoning:}
			\textcolor{Green}{FactAgent}~\cite{li2024large}{\char44}
			\textcolor{Green}{TED}~\cite{liu2025truth}{\char44}
			\textcolor{Green}{SheepDog}~\cite{wu2024fake}{\char44}
			\textcolor{Green}{ARG}~\cite{hu2024bad}{\char44}
			\textcolor{Green}{FACTGUARD}~\cite{he2025factguard}{\char44}
			\textcolor{Green}{RAMA}~\cite{yang2025rama}\\
			]
			]
			]
		\end{forest}
	}
	\caption{\textbf{Taxonomy of Multimodal Fake News Detection and Veracity Reasoning}. We systematically categorize multimodal fake news detection methods according to their parameter adaptation paradigms, including full fine-tuning, parameter-efficient tuning, prompting-based inference, and agent-based reasoning. This taxonomy provides a structured overview of state-of-the-art approaches and clarifies how different parameter interaction strategies are leveraged for multimodal veracity assessment.}
	\label{fig:taxonomy}
	\vspace{-4mm}
\end{figure*}
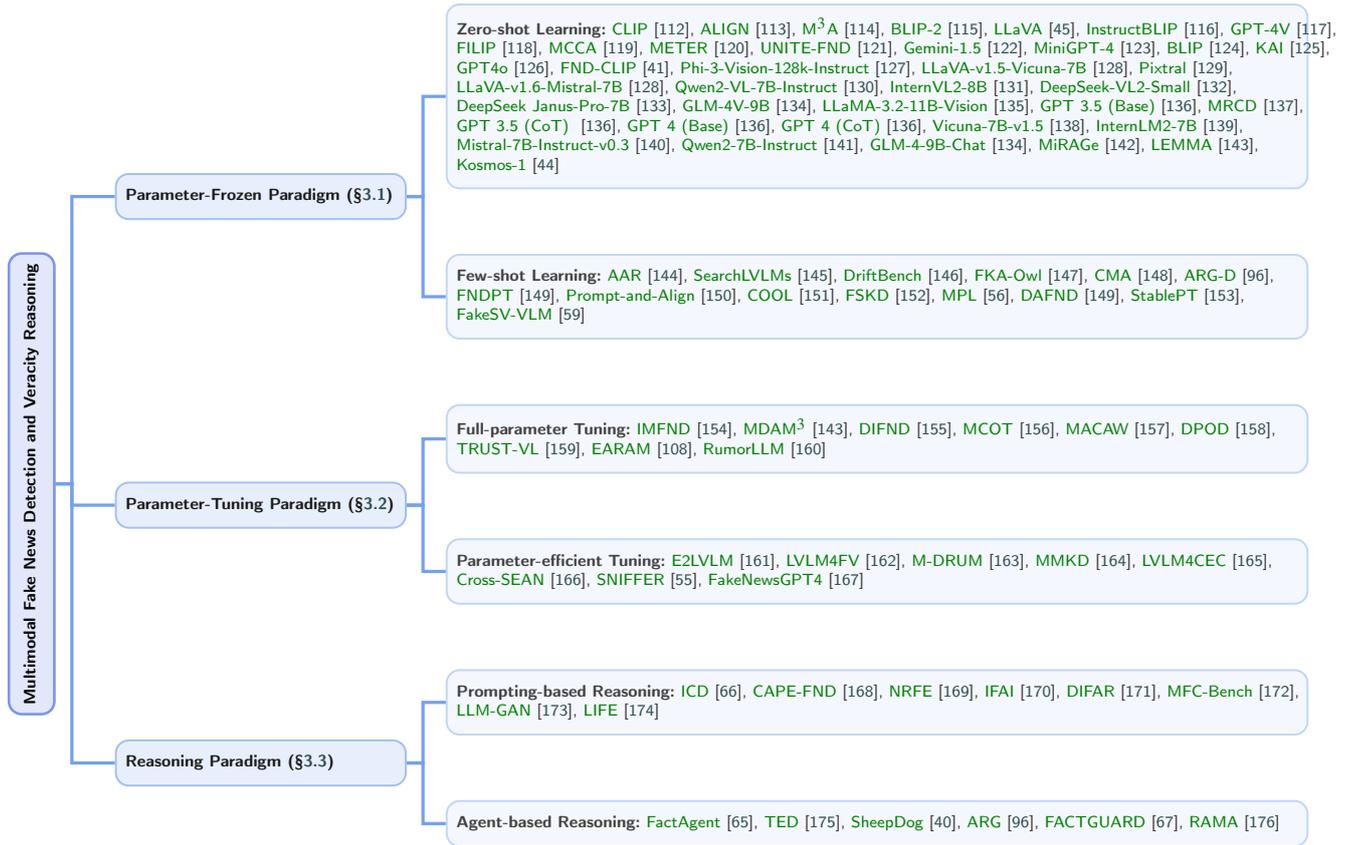

\section{Taxonomy}

We classify multimodal fake news detection methods into three paradigms, namely Parameter-Frozen Paradigm, Parameter-Tuning Paradigm, and Reasoning Paradigm. The Parameter-Frozen Paradigm relies on zero-shot or few-shot prompting without updating the model’s parameters, ensuring that the backbone model remains fixed. In contrast, the Parameter-Tuning Paradigm involves adapting the model’s parameters through full fine-tuning or parameter-efficient techniques such as adapters, prefix tuning, and Low Rank Adaptation (LoRA). Finally, the Reasoning Paradigm categorizes methods based on the type of reasoning employed, with prompting-based reasoning conducting implicit reasoning within a single inference pass and agent-based reasoning involving explicit, structured, multi step interactions for veracity reasoning. This classification is illustrated in Fig. \ref{fig:taxonomy}, providing a comprehensive framework for understanding the distinct approaches in multimodal fake news detection.

\subsection{Parameter-Frozen Paradigm}
\label{sec:Frozen}
In the parameter-frozen paradigm, the backbone LVLM remains unchanged, and task adaptation is realized purely through prompting and in-context specification \cite{fan2025generating}. This setting is particularly attractive for multimodal fake news detection because it avoids expensive fine-tuning on large models, and it enables rapid transfer across datasets and platforms with heterogeneous news formats \cite{hu2024bad}. Let $X = \{x_t, x_i, x_v\}$ denote text, image, and video inputs after modality-specific encoding or serialization into model-acceptable inputs, and let $\mathcal{I}$ denote a task instruction template that defines the goal (e.g., \textit{``Determine whether the given news is fake or real and justify the decision with cross-modal evidence.''}) \cite{mostafa2024modality}. The model produces a distribution over label strings given the constructed prompt $\Pi(X,\mathcal{I})$ as follows:
\begin{equation}
	P_{\Theta}(y \mid X, \mathcal{I}) = \sum_{s \in \nu(y)} P_{\Theta}(s \mid \Pi(X, \mathcal{I})) \label{eq:1}
\end{equation}
where $\Theta$ are frozen parameters and $\nu(y)$ is a verbalizer mapping each class $y$ (e.g., \textit{fake}, \textit{real}) to one or more label strings. The final decision is $\hat{y} = \arg\max_y P_{\Theta}(y \mid X, \mathcal{I})$.

\begin{figure*}
	\centering
	\includegraphics[width=1\linewidth]{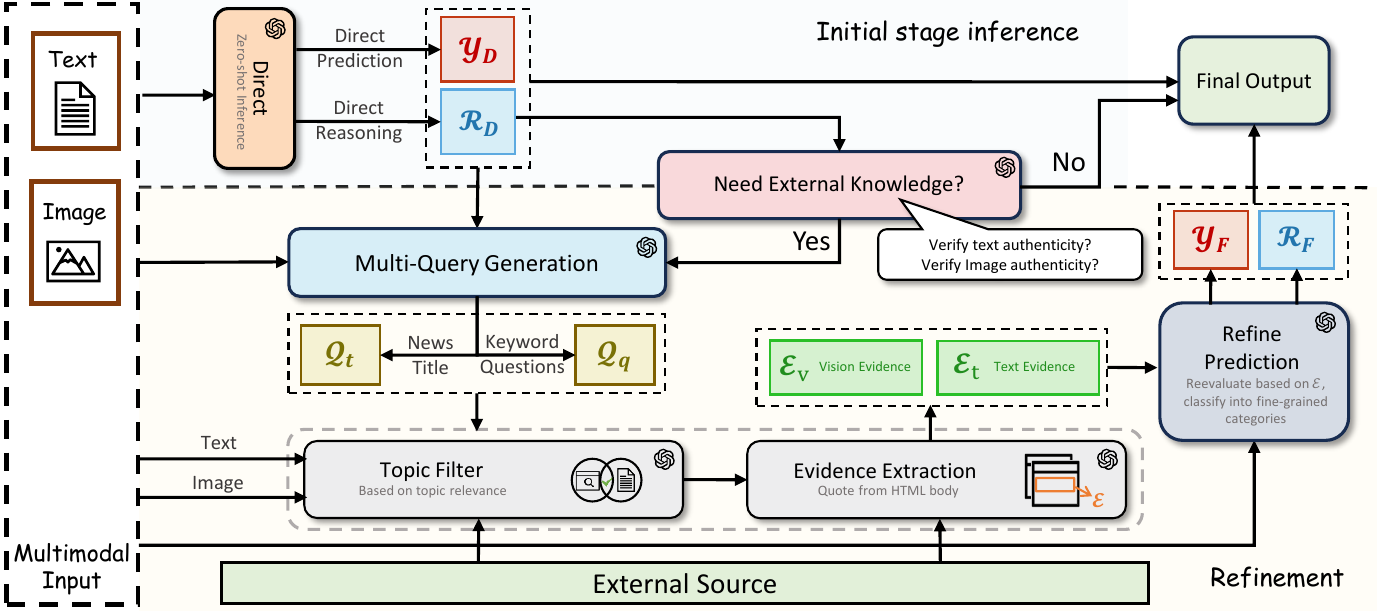}
	\caption{ Zero-shot learning framework for multimodal fake news detection. The architecture integrates direct prediction and reasoning mechanisms to process text and image inputs. It employs a multi-query generation module to formulate news-related queries based on the input's title and keywords. These queries are filtered for topical relevance and used to extract evidence from external sources. If external knowledge is deemed necessary, the system verifies the authenticity of both text and image inputs. The extracted evidence, along with the initial predictions, is refined through a reasoning process to produce the final output. This end-to-end approach enables the detection of fake news without prior training on labeled data, leveraging the complementary strengths of direct reasoning and external evidence integration \cite{xuan2024lemma}.}
	\label{fig:LEMMA}
\end{figure*}

\textbf{Mechanistic View under Frozen Parameters.}
From a mechanistic perspective, parameter-frozen LVLMs perform multimodal fake news detection by integrating visual tokens and textual instructions into a unified autoregressive reasoning process, rather than relying on explicit cross modal classifiers. Since model parameters remain unchanged, prompts act as the primary interface for controlling how multimodal evidence is attended to, decomposed, and compared. Different prompting strategies induce distinct reasoning behaviors \cite{wu2023prompt}. Direct classification prompts typically encourage holistic judgments, which may obscure fine grained semantic inconsistencies such as mismatched object attributes, quantities, or spatial relations. In contrast, structured prompts that explicitly request visual entity extraction, attribute comparison, or step by step reasoning promote finer grained alignment between visual evidence and textual claims \cite{shen2025gamed}. Chain-of-thought prompting further externalizes intermediate reasoning states, reducing over reliance on language priors and improving sensitivity to subtle cross-modal discrepancies. This mechanism distinguishes parameter-frozen LVLMs from traditional multimodal pipelines, where inconsistency detection is enforced through explicit similarity objectives or supervised heads. In frozen LVLMs, fine grained semantic inconsistency detection emerges implicitly from prompt guided reasoning trajectories, highlighting the central role of inference time design in determining detection robustness.

 \textbf{Zero-shot learning.} Zero-shot multimodal fake news detection relies solely on natural-language instructions without task-specific examples \cite{lin2023zero}. As shown in Fig. \ref{fig:LEMMA}. A well-engineered prompt serializes heterogeneous evidence into a structured context that highlights cross-modal credibility cues while minimizing spurious correlations \cite{choudhary2025beyond}. In LLM-based pipelines, non-text modalities are first converted into textual evidence snippets, e.g., image captions, entity–relation triples, and video scene summaries through auxiliary encoders, and concatenated with raw news articles using role tags (e.g., \texttt{<Text>}, \texttt{<Image-Caption>}, \texttt{<Video-Scene>}). In LVLM-based pipelines, raw embeddings are passed alongside textual tokens, but the model is still queried through instructions \cite{zhou2020survey}; 
hence Eq.(\ref{eq:1}) applies with $\Pi(\cdot)$ injecting modality tokens or connector-produced embeddings. To reduce prompt sensitivity and improve robustness, zero-shot systems commonly employ instruction variants and prompt ensembling with majority voting or probability averaging, self-consistency with
reasoning where latent rationales $r$ are sampled and marginalized as follows:
\begin{equation}
	P_{\Theta}(y \mid X, \mathcal{I}) \approx \sum_{r \in \mathcal{R}} P_{\Theta}(y \mid r, X, \mathcal{I}) P_{\Theta}(r \mid X, \mathcal{I})
\end{equation}
and contextual calibration subtracts a prior estimated from a content-free prompt to alleviate label-word frequency bias. Zero-shot decoding can output both a veracity label (fake or real) and an explanation, providing weak but valuable interpretability for auditing fake news decisions \cite{raj2023true}.

Representative methods have instantiated these principles in diverse ways. For instance, MRCD \cite{zhou2025collaborative} introduces a zero-shot multi round collaborative framework that iteratively generates hypotheses, retrieves external evidence, and refines predictions without supervised fine tuning. This approach effectively supports the detection of emerging fake news while enhancing decision explainability. Additionally, MRCD leverages collaborative evidence gathering, which improves robustness by facilitating cross source information validation and hypothesis refinement. LEMMA \cite{xuan2024lemma} adapts this paradigm by leveraging exemplar driven fusion pipelines in a zero-shot setting, showing that multimodal alignment and reasoning can be improved without task specific supervision. This model introduces an innovative approach by using exemplar based learning to guide reasoning, enabling better performance in resource constrained settings. MiRAGe \cite{huang2024miragenews} improves cross modal alignment in zero-shot scenarios through synthetic data augmentation and augmented connectors, enhancing multimodal reasoning. By synthesizing additional data, MiRAGe reduces the dependency on manually curated datasets, making it highly effective for scaling fake news detection across diverse domains. Similarly, M$^3$A \cite{xu2024m3a} integrates text, image, and domain adaptation features via large scale pre-training, achieving transferability across diverse disinformation benchmarks without task specific fine tuning. This model demonstrates the power of large scale pre-training, allowing it to generalize effectively across multiple fake news detection challenges. KAI \cite{zhang2024knowledge} proposes a knowledge enhanced interpretable network that leverages a large language model to generate target specific analytical perspectives, which are integrated through a bidirectional knowledge guided neural generation system. This enables zero-shot stance detection with improved interpretability, making KAI particularly suitable for real world applications that demand transparency and explainability in decision making processes.


\textbf{Few-shot learning.} Few-shot multimodal fake news detection augments the instruction with $k$ demonstration pairs $\{(X_i, y_i)\}_{i=1}^k$ injected into the prompt as in-context exemplars. The predictive distribution is as follows:
\begin{equation}
P_{\Theta}(y \mid X^*, \mathcal{I}, \mathcal{D}_k) = \sum_{s \in \nu(y)} P_{\Theta}(s \mid \Pi(X^*, \mathcal{I}, \mathcal{D}_k))
\end{equation}
where the selection of $D_k$ is crucial. Effective strategies balance relevance to the query and diversity across news topics and modalities \cite{sharma2019combating, abdali2024multi}, often via embedding-based retrieval with determinantal or max–min objectives to avoid redundancy \cite{nezafat2024fake}. Demonstrations should expose the model to prototypical cross-modal inconsistencies (e.g., text–image mismatch, fabricated video snippets, or misleading multimodal narratives) and discourse contexts (e.g., satire, conspiracy, and propaganda) that are underrepresented in pre-training \cite{zhang2025knowledge}. Ordering also matters: placing task definition first, then structured exemplars (“context $\rightarrow$ rationale $\rightarrow$ label”), and finally the query tends to reduce hallucination \cite{zhang2025llms}. For classification with free-form generation, a rationale-then-verbalizer template that first elicits a short explanation $r$ and then constrains the final answer to $\nu(y)$ often improves calibration and stability \cite{feng2025contradicted}. 
When label spaces differ across datasets, dynamic verbalizers provide a principled bridge by defining synonyms per class and aggregating token probabilities \cite{verma2021welfake}.

Representative few-shot systems in fake news detection operationalize these principles through diverse mechanisms. For instance, AAR \cite{zheng2025unveiling} demonstrates that incorporating adaptive exemplars enables models to align retrieved evidence with news claims under limited supervision. By dynamically selecting relevant exemplars, AAR allows for effective alignment of multimodal evidence, even in settings with scarce labeled data, thus improving robustness in detecting inconsistencies. SearchLVLMs \cite{li2024searchlvlms} shows that few-shot multimodal demonstrations can guide large vision language models to detect subtle inconsistencies between text and images by leveraging retrieval augmented exemplars. The integration of retrieval augmented learning in SearchLVLMs allows for more precise detection of fine grained inconsistencies, effectively enhancing model sensitivity to cross-modal discrepancies, such as conflicting visual cues or ambiguous textual claims. DriftBench \cite{li2025drifting} extends few-shot evaluation by systematically assessing model robustness under limited exemplars and distribution shifts, providing insights into adaptability against evolving disinformation patterns. This framework facilitates a deeper understanding of how models perform under real world conditions where training data may vary, highlighting the importance of model adaptability to shifting distributions in fake news contexts. FKA-Owl \cite{liu2024fka} further enhances robustness by integrating structured attention over curated demonstrations, revealing that hierarchical exemplar design mitigates the impact of noisy or adversarial content. This model's hierarchical attention mechanism helps filter out noise and adversarial influences, ensuring that models remain reliable even when presented with imperfect or misleading evidence. At a broader scale, FakeSV-VLM \cite{wang2025fakesv} exemplifies few-shot adaptability by exploiting cross source and cross event multimodal exemplars, improving generalization without necessitating full model fine tuning. By leveraging cross source and event specific data, FakeSV-VLM improves the model’s transferability across diverse contexts, ensuring more effective detection across different types of disinformation. Collectively, these systems highlight that few-shot learning strikes a pragmatic balance between efficiency and adaptability, making it a compelling paradigm for fake news detection in scenarios where annotated data is limited but multimodal evidence must be effectively utilized.


\subsection{Parameter-Tuning Paradigm}
\label{sec:Tuning}

Unlike the parameter-frozen paradigm that keeps $\Theta$ fixed, the parameter-tuning paradigm updates part or all of the LVLM parameters $\Theta$ to enhance domain specialization \cite{nan2021mdfend}, multimodal calibration \cite{xue2021detecting}, and interpretable reasoning \cite{guo2023interpretable}.
Let each modality encoder be $\phi_t, \phi_v$ for text, vision inputs, and let $\mathcal{A}_{\theta}$ denote the cross-modal aggregator. 
Given a multimodal dataset as follows:
\begin{equation}
	\mathcal{D}=\{(x_i,c_i,y_i,R_i)\}_{i=1}^N
\end{equation}
where $x_i=\{x_t,x_v\}$ are paired modalities \cite{liu2025modality}, $c_i$ is the claim, $y_i\in\{0,1\}$ is the veracity label \cite{silva2021embracing}, and $R_i$ the annotated rationale, 
parameter tuning seeks the optimal $\Theta$ that minimizes a task-aligned multi-objective loss as follows:
\begin{equation}
	\mathcal{L}_{tune}(\Theta)
	= \mathcal{L}_{LVLM}
	+ \lambda_{reg}\Vert \Theta-\Theta_0\Vert_2^2
	\label{eq:ltune}
\end{equation}
where $\mathcal{L}_{LVLM}$ follows Eq.~(\ref{eq:lvlm-loss}) and $\Theta_0$ is the pretrained initialization, acting as a regularizer to prevent catastrophic forgetting.

\textbf{Mechanistic View under Parameter-Tuning Paradigm.}
From a mechanistic perspective, the parameter-tuning paradigm enhances the multimodal reasoning capabilities of LVLMs by fine tuning a subset of model parameters, allowing for a more flexible integration of text and image modalities. Unlike traditional models that treat text and images as separate inputs, LVLMs in this paradigm jointly optimize the parameters of both the textual and visual encoders through cross modal aggregators. This enables the model to better capture fine grained semantic inconsistencies between text and image, such as mismatched attributes or spatial relations. The effectiveness of this mechanism is largely driven by the choice of prompting strategy. Direct classification prompts encourage holistic judgments, which may overlook subtle mismatches, whereas structured prompts (e.g., entity extraction, attribute comparison, or step-by-step reasoning) guide the model to attend to specific details, improving the alignment between the visual and textual data. Furthermore, chain-of-thought prompting externalizes intermediate reasoning steps, reducing over reliance on language priors and enhancing the model's sensitivity to cross modal discrepancies. Through the fine tuning of model parameters and strategic prompting, the parameter-tuning paradigm  allows LVLMs to implicitly detect fine grained semantic inconsistencies, offering a more adaptive and robust approach compared to traditional methods.

\textbf{Full-parameter tuning.} Full fine-tuning unfreezes all LVLM components \cite{qin2024boosting}, including modality encoders $\phi_t, \phi_v$, aggregator $\mathcal{A}_{\theta}$, and decision heads $(w_F,w_D,w_I,w_Y)$. 
The optimization jointly refines the authenticity \cite{shu2019role}, consistency \cite{wu2023human}, and intent estimators as follows:
\begin{equation}
\begin{aligned}
	&\hat{F}=\sigma(w_F^{\top}Z+b_F) \\
	&\hat{D}=1-\mathrm{sim}(g_t(x_t),g_v(x_v)) \\
	&\hat{I}=\sigma(w_I^{\top}Z+b_I) \\
	&h_{\theta}(x,c,\xi)=w_Y^{\top}Z+b_Y
	\label{eq:fine_heads}
\end{aligned}
\end{equation}
and the calibrated decision surrogate remains as follows:
\begin{equation}
	s(x,c,\xi)=h_{\theta}(x,c,\xi)+\alpha_F\hat{F}+\alpha_D\hat{D}+\alpha_I\hat{I}
	\label{eq:s_finetune}
\end{equation}
During training, all parameters receive gradients \cite{zhang2024early} from $\mathcal{L}_{LVLM}$, enabling domain-specific adaptation \cite{mosallanezhad2022domain} as follows:
\begin{equation}
\begin{aligned}
	\mathcal{L}_{LVLM} 
	&= l_{cls}(y,\hat{y})
	+\lambda_F l_F(F,\hat{F})
	+\lambda_D l_D(D,\hat{D})
	+\lambda_I l_I(I,\hat{I}) \\
	&
	+\lambda_R l_{exp}(R,\hat{R})
	+ \mathcal{L}_{align}
	+ \mathcal{L}_{LM}
	+ \mathcal{L}_{ptr}
	\label{eq:llvlm_finetune}
\end{aligned}
\end{equation}
To stabilize optimization \cite{kumar2023optnet}, practical implementations adopt stage-wise unfreezing, mixed-precision gradient checkpointing, 
and curriculum fine-tuning (progressing from single-claim to multi-evidence reasoning) \cite{kao2024we}. 
Full tuning allows LVLMs to internalize complex cross-modal contradictions and high-level intents but requires significant computational and data resources \cite{chen2022cross}.

\begin{figure*}
	\centering
	\includegraphics[width=1\linewidth]{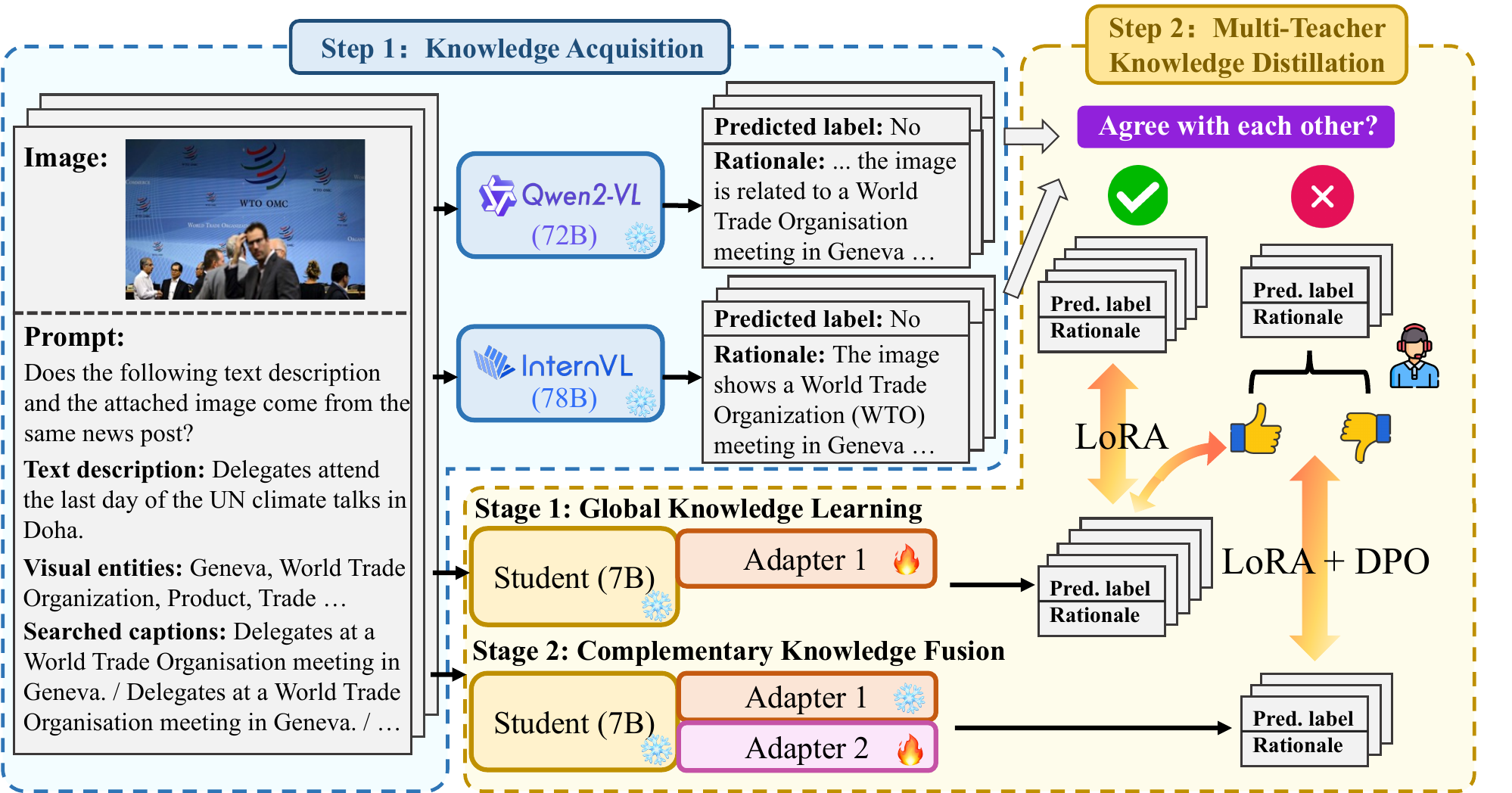}
	\caption{Parameter-efficient tuning architecture for multimodal fake news detection. The framework operates in two stages: global knowledge learning and complementary knowledge fusion. In the first stage, a student model with 7 billion parameters is trained alongside two adapters, processing visual and textual inputs to identify relevant entities and search options. The second stage integrates the student model with the adapters to enhance reasoning. Knowledge acquisition is achieved through two vision-language models, Qwen2-VL and InternVL, which provide predictions and rationales for the input image. Multi-teacher knowledge distillation follows, where the student model and adapters align their predictions and rationales, facilitated by LoRA and DPO techniques to refine the model's outputs \cite{zeng2024multimodal}.}
	\label{fig:MMKD}
\end{figure*}

Recent full-tuning approaches in multimodal fake news detection have demonstrated both the strengths and limitations of this paradigm \cite{shao2025deepfake}. For instance, IMFND \cite{jiang2025imfnd} established the foundation by fully fine tuning pretrained multimodal transformers on paired text image datasets \cite{zhang2023multimodal}, achieving strong results in capturing cross modal inconsistencies in news content. By optimizing all parameters, IMFND improves the model’s ability to learn fine grained interactions between text and images, significantly enhancing its capacity to identify misleading content. Building on these successes, video oriented systems such as MDAM$^3$ \cite{xuan2024lemma} extended full parameter tuning to spatiotemporal news verification, showing that large scale video pretraining coupled with end-to-end optimization yields significant gains in detecting deepfake based misinformation \cite{zong2025text}. MDAM$^3$ pushes the boundaries of video based fake news detection by effectively incorporating both temporal and spatial cues, improving the model's robustness against dynamic manipulations often found in deepfake content. More recent models such as DIFND \cite{yan2025debunk} and TRUST-VL \cite{yan2025trust} scale this strategy by integrating multimodal evidence with reasoning rich supervision, demonstrating that fully updating all parameters enhances both interpretability and robustness against adversarial news \cite{guo2023interpretable}. These models exemplify how reasoning guided updates allow for better disambiguation of complex multimodal evidence, improving both the model's generalization and its ability to handle adversarial inputs in fake news scenarios. Similarly, MACAW \cite{yindetecting} leverages parameter intensive optimization to jointly model discourse level attention and multimodal alignment \cite{li2021entity}, highlighting the effectiveness of full-tuning for handling long context and linguistically nuanced misinformation \cite{verma2021welfake}. By modeling discourse level attention, MACAW improves the model’s ability to capture long range dependencies and subtle language nuances, which are critical in understanding and detecting misinformation embedded in lengthy articles or complex narratives.

While these systems highlight the versatility and strong performance of full-parameter tuning in fake news detection, they also expose its substantial computational costs \cite{sharma2019combating}, which motivates ongoing exploration \cite{apuke2021user} of more efficient alternatives such as parameter-efficient tuning.

\textbf{Parameter-efficient tuning.} To alleviate the prohibitive cost of full fine-tuning while preserving most of its performance benefits \cite{hamed2025enhanced}, parameter-efficient tuning (PET) has become a practical strategy for multimodal fake news detection \cite{zhang2024early}. In this paradigm, the backbone of large vision–language or video–language models is kept frozen, while lightweight modules, such as Adapter-tuning \cite{houlsby2019parameter}, LoRA \cite{hulora}, Prefix/Prompt-tuning \cite{li2021prefix,lester2021power}, or gating mechanisms (e.g., IA$^3$ \cite{liu2022few}, BitFit \cite{ben2022bitfit}) are optimized to capture domain-specific patterns. For multimodal news involving text, images, and videos, PET allows detectors to emphasize subtle lexical manipulations, visual tampering artifacts, and temporal inconsistencies without incurring the high computational overhead of updating the entire model. Recent advances such as QLoRA \cite{dettmers2023qlora} further improve scalability by quantizing the frozen backbone to 4-bit while training LoRA adapters in higher precision, enabling efficient deployment at scale. Moreover, PET is often combined with connector tuning or selective unfreezing of higher Transformer layers to enhance discourse-level reasoning \cite{zhou2019network} and cross-modal alignment under constrained resources \cite{qiao2025improving}. Specifically, parameter-efficient tuning keeps most of $\Theta$ frozen and introduces a small trainable subset $\Theta'$ through lightweight modules \cite{shan2021poligraph} as follows:
\begin{equation}
	\Theta'=\{A_{adapter}, P_{prefix}, L_{LoRA}\}
	\label{eq:pet_module}
\end{equation}
Each module injects learnable low-rank or prefix parameters into the transformer blocks of $\mathcal{A}_{\theta}$ or modality encoders as follows:
\begin{equation}
	Z'=\mathcal{A}_{\theta+\Theta'}(\phi_t(\tilde{t}),\phi_v(\tilde{v}),\phi_t(\tilde{\xi}))
	\label{eq:pet_z}
\end{equation}
The downstream heads $(w_F,w_D,w_I,w_Y)$ are fine-tuned on $Z'$, while the frozen backbone retains general cross-modal alignment from pre-training. 
The loss function remains Eq.~(\ref{eq:llvlm_finetune}) but with gradient flow restricted to $\Theta'$. 
Regularization such as orthogonality or low-rank constraints on LoRA matrices (rank $r \ll d$) ensures stable adaptation:
\begin{equation}
	\min_{\Theta'} 
	\mathcal{L}_{LVLM}(\Theta')
	+\beta \Vert A_{LoRA}A_{LoRA}^{\top}-I \Vert_F^2
	\label{eq:lora_reg}
\end{equation}

Recent PET-based systems have demonstrated the practicality of parameter-efficient tuning in multimodal fake news detection \cite{wu2024fake}. For instance, E2LVLM \cite{wu2025e2lvlm} employs selective adaptation to integrate retrieved textual and visual evidence, significantly reducing training overhead while maintaining cross domain generalization. E2LVLM's selective adaptation approach allows the model to adapt only relevant parts of the pretrained model, ensuring efficient use of resources while still achieving strong performance across diverse domains and multimodal data. LVLM4FV \cite{tahmasebi2024multimodal} and M-DRUM \cite{jin2024fake} extend PET strategies to evidence driven verification, incorporating lightweight adapters into frozen LVLMs to efficiently capture cross modal reasoning under limited supervision. These models demonstrate that by incorporating lightweight adapters, it is possible to enhance model efficiency and generalization without the need for extensive fine tuning, offering a scalable solution for multimodal fake news detection. As shown in Fig. \ref{fig:MMKD}, MMKD \cite{zeng2024multimodal} illustrates how PET can unify multimodal representation learning with knowledge guided reasoning, showing that adapter based modules can approximate the gains of full tuning while remaining computationally efficient. MMKD’s approach of combining knowledge guided reasoning with PET allows the system to maintain interpretability and efficiency, providing a lightweight yet powerful solution for multimodal reasoning. Domain specific adaptations, such as LVLM4CEC \cite{tahmasebi2025verifying}, demonstrate that parameter efficient designs can preserve sensitivity to subtle context event correlations when annotated multimodal corpora are scarce. LVLM4CEC’s ability to adapt to domain specific nuances without relying heavily on large annotated datasets highlights its effectiveness in specialized fake news detection tasks, particularly in resource constrained environments. Cross-SEAN \cite{kumar2023optnet} illustrates how adapter based architectures can be tuned alongside frozen backbones to support cross lingual and rationale grounded debunking. Cross-SEAN’s cross lingual capabilities enhance its versatility in addressing fake news in multiple languages, while rationale-grounded debunking adds a layer of interpretability by explaining the model’s reasoning process. Collectively, these systems exemplify how PET techniques balance the high performance of full parameter tuning with the scalability required for real-world deployment, underscoring their growing role in advancing multimodal fake news detection \cite{mosallanezhad2022domain}.


	\subsection{Reasoning Paradigm}
	\label{sec:Reasoning}
	At both the system and cognitive levels, multimodal fake news detection methods employ distinct reasoning strategies. Prompting-based reasoning operates at the system level by implicitly guiding the model’s reasoning process through instructions or prompts. In contrast, agent-based reasoning functions at the cognitive level by structuring reasoning as an explicit sequence of coordinated sub-tasks, simulating a more deliberate and human like approach to decision making.
	
	\textbf{Mechanistic View under Reasoning Paradigm.} From a mechanistic perspective, the reasoning paradigm in LVLMs offers a significant advancement over traditional models in detecting fine grained semantic inconsistencies between text and images. In prompting-based reasoning, LVLMs use task-specific prompts to guide the model's reasoning in a single inference pass, which contrasts with traditional methods that typically rely on separate pipelines for text and image processing. However, while prompting-based methods enable efficient, zero-shot generalization, they may obscure subtle mismatches between textual and visual evidence, especially when using direct classification prompts. In contrast, structured prompts such as those for entity extraction or step-by-step reasoning allow LVLMs to more explicitly align and detect discrepancies in object attributes, quantities, and spatial relationships, improving the model's sensitivity to fine grained inconsistencies. On the other hand,  agent-based reasoning decomposes the detection process into a sequence of explicit tasks performed by specialized agents, which interact to refine the reasoning process. This approach provides greater transparency and interpretability compared to prompting-based reasoning, allowing for more deliberate, human like decision making. It also facilitates error localization and enables strategies like counter argumentation, which are essential for handling complex, long context misinformation. By combining these two paradigms, LVLMs are able to surpass traditional models in robustness and interpretability, providing a more flexible and effective approach to multimodal fake news detection.

	\textbf{Prompting-based Reasoning.} Prompting-based reasoning methods treat multimodal fake news detection as a single pass inference problem, where reasoning is implicitly induced through carefully designed prompts, instructions, or constraints, without the need for explicit intermediate decision states or tool driven interactions. Given multimodal evidence \( x = \{x_t, x_v\} \) and a claim \( c \), the model directly produces a veracity prediction:
	\begin{equation}
		\hat{y} = f_{\Theta}(x_t, x_v, c; \mathcal{P})
	\end{equation}
	where \( x_t \) and \( x_v \) represent the textual and visual modalities of the multimodal evidence, respectively, and \( c \) denotes the claim whose veracity is being assessed. \( \mathcal{P} \) is a task specific prompt encoding reasoning cues, such as cross modal consistency, credibility, or relevance of the evidence.
	
	This paradigm leverages the emergent reasoning capabilities of large vision language models (LVLMs) or language models, using prompt engineering to guide the model toward detecting semantic inconsistencies, fabricated visual evidence, or misleading narratives. Representative approaches include ICD \cite{wang2024mitigating} and CAPE-FND \cite{jin2025veracity}, which mitigate multimodal hallucination and calibration errors through instruction level constraints, ensuring that models provide more accurate interpretations by imposing explicit guidance on reasoning pathways. These systems enhance the reliability of LVLMs by constraining the model's outputs to adhere more strictly to logical reasoning patterns, reducing the occurrence of hallucinations in multimodal content. NRFE \cite{zhang2025llms} and IFAI \cite{Zhang2025ConfidenceBS}, which employ confidence aware or feedback enhanced prompting to improve robustness, further address the challenge of multimodal inconsistency by adjusting the confidence levels in the model's output. NRFE utilizes feedback loops to refine reasoning accuracy, while IFAI focuses on boosting robustness in video based fake news detection through the strategic use of feedback enhanced prompts that guide the model’s attention toward more reliable features. Specifically, IFAI addresses challenges in fake news video detection by constructing multimodal prompts through prompt engineering to semantically understand news videos and generate auxiliary information from the perspectives of video style, content, and information matching. The video information interactor enables small models to effectively learn supplementary knowledge from LVLMs, while the key information selector evaluates the importance of inference rationales, thus improving the efficiency of utilizing knowledge from LVLMs. This mechanism significantly improves the model’s capacity to discern subtle discrepancies between video content and textual narratives, enhancing detection performance in dynamic, real world contexts. Recent works, such as DIFAR \cite{wan2025difar} and LIFE \cite{wang2025prompt}, demonstrate that structured prompts can replicate reasoning behaviors typically associated with explicit reasoning pipelines. DIFAR introduces a robust framework for prompt guided reasoning, leveraging structured prompts to replicate the reasoning steps commonly used in traditional, explicit pipelines, while LIFE focuses on long term effectiveness by applying structured prompt templates that align multimodal evidence over extended interactions. Benchmarks like MFC-Bench \cite{wang2025mfc} and generative frameworks such as LLM-GAN \cite{wang2025llm} highlight both the potential and limitations of prompt induced reasoning, especially under adversarial or distribution shifted conditions. These benchmarks demonstrate that, while prompt based methods show significant promise in handling a wide range of multimodal scenarios, challenges remain in maintaining consistent reasoning under adversarial settings or when the distribution of data shifts significantly from the training conditions.

	Despite their efficiency and strong zero-shot generalization, prompting-based methods inherently combine perception, reasoning, and decision making within a single inference pass. This monolithic approach limits controllability and interpretability, particularly when dealing with complex, multi hop misinformation.
	
	\textbf{Agent-based Reasoning.} Unlike prompting-based reasoning, which relies on implicit reasoning in a single inference step, agent-based reasoning explicitly decomposes the multimodal fake news detection process into a sequence of structured, multi step tasks. This paradigm involves multiple specialized agents, each performing a distinct sub-task such as evidence retrieval, cross modal verification, contradiction detection, and final adjudication. These agents interact with one another to refine the reasoning process, ensuring that the detection system can handle complex scenarios requiring in-depth analysis. Formally, the inference process is modeled as a sequence of agent interactions:
	
	\begin{equation}
		\mathcal{S}_{t+1} = \mathcal{A}_k(\mathcal{S}_t, x, c), \quad k \in \{1, \dots, K\}
	\end{equation}
	
	where \( \mathcal{S}_t \) represents the intermediate reasoning state at time step \( t \), and \( \mathcal{A}_k \) denotes a task-specific agent responsible for executing a particular sub-task.
	
	Systems like FactAgent \cite{li2024large} and TED \cite{liu2025truth} exemplify the agent-based reasoning paradigm by coordinating multiple agents that iteratively verify claims using multimodal evidence, allowing for explicit control over the reasoning order and the selection of relevant evidence at each step. This coordination of agents enables more flexible and interpretable reasoning processes, as each agent can specialize in evaluating a particular type of evidence, whether textual, visual, or audio, and can adaptively select the most relevant information for the task at hand. These systems enable better traceability of the reasoning process, ensuring transparency and accountability in decision making by offering a detailed log of the steps each agent took in reaching its conclusion. Furthermore, agent-based frameworks like SheepDog \cite{wu2024fake} and ARG \cite{hu2024bad} take this approach further by introducing adversarial or debate-style agents, which can engage in counter-argumentation, helping the system improve robustness against false or manipulated narratives. The adversarial agents engage in a structured exchange of opposing viewpoints, forcing the system to critically evaluate evidence from different perspectives and refine its reasoning, thus enhancing its ability to detect inconsistencies or biases in the evidence. By simulating a back-and-forth debate, these agents can identify subtle inconsistencies in the evidence, improving the system's ability to handle misleading or deceptive content. More recent architectures, such as FACTGUARD \cite{he2025factguard}, have integrated tool-augmented agents that not only reason about the evidence but also maintain structured memory and rationale supervision. FACTGUARD introduces a memory mechanism that allows agents to retain and reference previous reasoning steps, making it possible to build a more coherent and structured argument over time. This approach enhances interpretability by providing clear, human readable reasoning traces that explain how conclusions are reached. Furthermore, these systems demonstrate improved generalization under complex, long-context misinformation scenarios, where the model must synthesize and analyze large volumes of data over extended reasoning chains. By incorporating memory and rationale supervision, these systems can maintain contextual continuity across longer interactions, improving the accuracy of fake news detection in scenarios where misinformation evolves gradually.

	While agent-based reasoning systems are computationally more demanding and involve greater system complexity, they provide several key advantages. These include enhanced transparency, the ability to modularize reasoning processes, and improved error localization, all of which make agent-based methods highly suited for high stakes applications such as judicial decision-making, legal analysis, and evidence intensive fake news verification. By ensuring that each step in the reasoning process is explicitly documented and accountable, agent-based systems offer a level of trustworthiness and flexibility that is critical in real world verification tasks.

\section{Popular Benchmark Dataset}

In recent years, the research progress in multimodal fake news detection has benefited greatly from the continuous construction and opening of high-quality datasets. Datasets not only provide a basis for model training and evaluation, but also largely define the boundaries and difficulty of research problems. Starting from early social media rumor detection datasets such as Twitter15 \cite{liu2015real} and Twitter16 \cite{ma2016detecting}, researchers have gradually expanded from single text or image modalities to multimodal resources integrating text, images, videos and social context, and continuously improved data scale, modality diversity and annotation precision, thereby promoting the development of more complex detection tasks that are closer to real-world scenarios. As shown in Fig. \ref{fig:datasets_tree}, Looking back at the related work from 2015 to 2025, we can observe that dataset construction shows the following four development trends as follows:

\begin{figure*}
	\centering
	\includegraphics[width=1\linewidth]{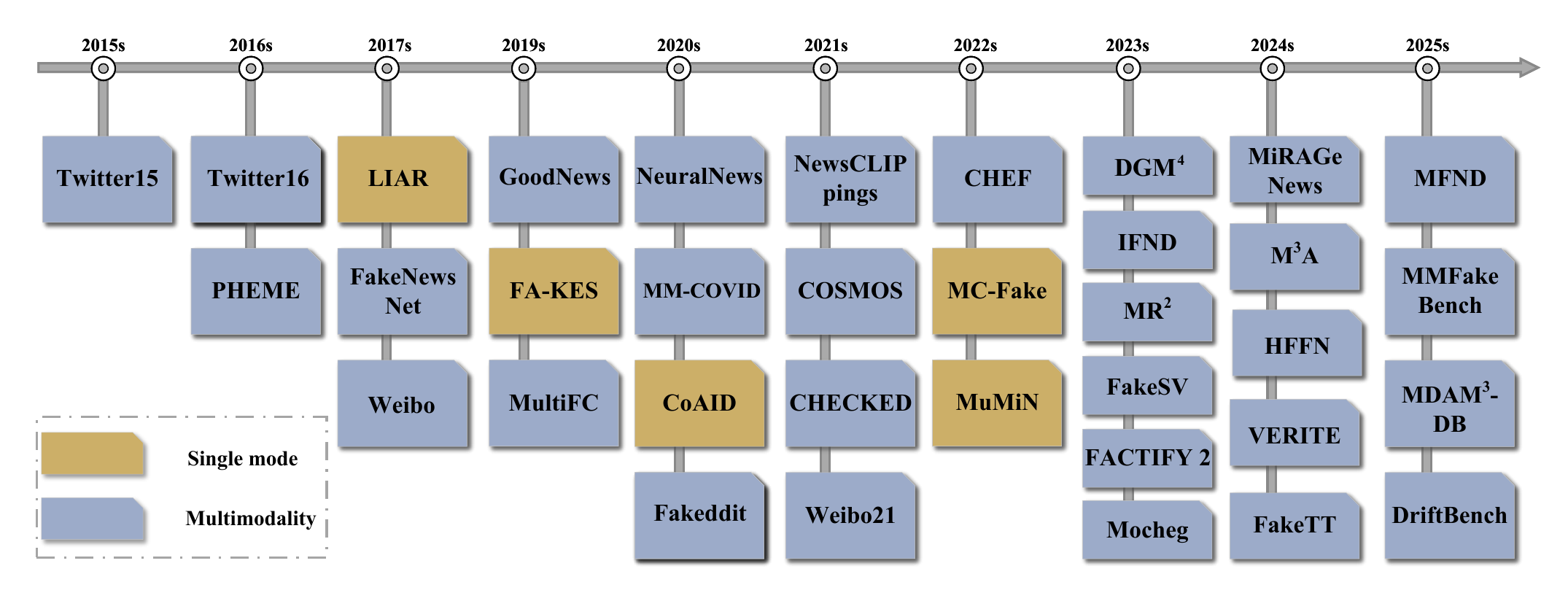}
	\caption{Timeline of multimodal fake news detection datasets.}
	\label{fig:datasets_tree}
\end{figure*}

\begin{itemize}
	\item \textbf{Data Scale Expansion:} Early datasets such as LIAR \cite{wang2017liar}, Weibo \cite{jin2017multimodal}, and PHEME \cite{zubiaga2016analysing} typically contained only a few thousand to tens of thousands of examples and were primarily used for training and validating small-scale models. As research deepened, subsequent datasets (such as M$^{3}$A \cite{xu2024m3a}, Fakeddit \cite{nakamura2020fakeddit}, and NewsCLIPpings \cite{luo2021newsclippings}) expanded to hundreds of thousands or even millions of examples, facilitating the application of deep learning and large-scale pre-trained models, and supporting the systematic evaluation of complex models at varying data scales.
	
	\item \textbf{Modal Diversity:} Initially, data was primarily text-based. Subsequently, images (such as Weibo \cite{jin2017multimodal} and GoodNews \cite{biten2019good}), image-text alignment data (such as NeuralNews \cite{tan2020detecting} and COSMOS \cite{aneja2021cosmos}), and multilingual content (such as MM-COVID \cite{li2020mm} and CHECKED \cite{yang2021checked}) were introduced, and further expanded to include audio, video, and social interaction information (such as M$^{3}$A \cite{xu2024m3a} and MDAM$^{3}$-DB \cite{xu2025mdam3}). This increased modal diversity not only provides the model with richer learning signals but also better reflects the complex spread of disinformation on real social platforms.
	
	\item \textbf{Task Dimension Expansion:} Early research focused primarily on binary classification tasks (authenticity and fake identification), but recent datasets have gradually expanded to support multi-class labeling (such as LIAR \cite{wang2017liar} and Fakeddit \cite{nakamura2020fakeddit}), tampered region localization (such as DGM$^{4}$ \cite{shao2023detecting} and MFND \cite{zhu2025multimodal}), image-text inconsistency detection (such as COSMOS \cite{aneja2021cosmos} and NewsCLIPpings \cite{luo2021newsclippings}), and AI-generated content recognition (such as MiRAGeNews \cite{huang2024miragenews} and MMFakeBench). This expansion of task dimensions has greatly increased the challenge and application value of detection, enabling research to cover diverse objectives, from coarse-grained to fine-grained, and from discriminative to generative.
	
	\item \textbf{Cross-language and cross-domain adaptation:} With the widespread global spread of disinformation, the need for cross-language and cross-domain detection has become increasingly prominent. Some datasets (such as MM-COVID \cite{li2020mm}, Weibo21 \cite{nan2021mdfend}, and CHEF) focus on multilingual, cross-topic, and cross-domain authenticity detection. These datasets not only enrich the training corpus but also provide important support for research on model transferability and cross-cultural adaptability.
\end{itemize}


Overall, dataset evolution over the past decade has seen a gradual transition from early rumor detection to complex cross-modal disinformation identification. These datasets have continuously improved in terms of authenticity annotation granularity, modal coverage, cross-lingual adaptability, and generative content detection. These datasets not only provide a solid experimental benchmark for the research community but also drive the field's progress from traditional classification tasks to more complex multi-task, multi-modal, and multi-lingual approaches, forming a relatively complete evolutionary trajectory and research ecosystem. Table \ref{dataset_link}. lists 32 fake news detection benchmark datasets. We analyze the release time, modality, and open-source URL of each dataset. Next, we briefly introduce each dataset.

\begin{table*}[H]
	\centering
	\caption{Publicly available benchmark datasets in multimodal fake news detection.}
	\begin{tabular}{@{}lllll@{}}
		\toprule
		\textbf{Datasets} & \textbf{Year} & \textbf{Modality} & \textbf{Available at} \\ \midrule
		Twitter15 \cite{liu2015real} & 2015 & Text,Image & \url{https://github.com/majingCUHK/Rumor_RvNN?tab=readme-ov-file} \\
		Twitter16 \cite{ma2016detecting} & 2016 & Text,Image & \url{https://github.com/majingCUHK/Rumor_RvNN?tab=readme-ov-file} \\
		PHEME \cite{zubiaga2016analysing} & 2016 & Text,Image & \url{https://www.kaggle.com/datasets/usharengaraju/pheme-dataset} \\
		LIAR \cite{wang2017liar} & 2017 & Text & \url{https://www.kaggle.com/datasets/yuktibishambu/liar-dataset-labeled} \\
		FakeNewsNet \cite{shu2017fake} & 2017 & Text,Image & \url{https://github.com/KaiDMML/FakeNewsNet} \\
		Weibo \cite{jin2017multimodal} & 2017 & Text,Image & \url{https://github.com/plw-study/MRML?tab=readme-ov-file} \\
		GoodNews \cite{biten2019good} & 2019 & Text,Image & \url{https://github.com/furkanbiten/GoodNews} \\
		FA-KES \cite{salem2019fa} & 2019 & Text & \url{https://www.kaggle.com/datasets/mohamadalhasan}\\
		MultiFC \cite{salem2019fa} & 2019 & Text,Image & \url{https://huggingface.co/datasets/pszemraj/multi_fc}\\
		Fakeddit \cite{nakamura2020fakeddit} & 2020 & Text,Image & \url{https://github.com/entitize/fakeddit} \\
		NeuralNews \cite{tan2020detecting} & 2020 & Text,Image & \url{https://drive.google.com/file/d/1vD4DtyJOIjRzchPtCQu-KPrUjgTiWSmo/view} \\
		MM-COVID \cite{li2020mm} & 2020 & Text,Image & \url{https://drive.google.com/drive/folders/1gd4AvT6BxPRtymmNd9Z7ukyaVhae5s7U} \\
		CoAID \cite{cui2020coaid} & 2020 & Text & \url{https://github.com/cuilimeng/CoAID} \\
		NewsCLIPpings \cite{luo2021newsclippings} & 2021 & Text,Image & \url{https://github.com/g-luo/news_clippings?tab=readme-ov-file} \\
		COSMOS \cite{aneja2021cosmos}  & 2021 & Text,Image & \url{https://shivangi-aneja.github.io/projects/cosmos/} \\
		CHECKED \cite{yang2021checked}  & 2021 & Text,Image & \url{https://github.com/cyang03/CHECKED/tree/master/dataset} \\
		Weibo21 \cite{nan2021mdfend} & 2021 & Text,Image & \url{https://github.com/kennqiang/MDFEND-Weibo21} \\
		CHEF \cite{hu2022chef} & 2022 & Text & \url{https://github.com/THU-BPM/CHEF?tab=readme-ov-file} \\
		MC-Fake \cite{min2022divide} & 2022 & Text & \url{https://github.com/qwerfdsaplking/MC-Fake} \\
		MuMiN \cite{nielsen2022mumin} & 2022 & Text,Image & \url{https://mumin-dataset.github.io/} \\
		DGM$^{4}$ \cite{shao2023detecting} & 2023 & Text,Image & \url{https://huggingface.co/datasets/rshaojimmy/DGM4} \\
		IFND \cite{sharma2023ifnd} & 2023 & Text,Image & \url{https://www.kaggle.com/datasets/sonalgarg174/ifnd-dataset} \\
		MR$^{2}$ \cite{hu2023mr2} & 2023 & Text,Image & \url{https://github.com/THU-BPM/MR2} \\
		Mocheg \cite{yao2023end} & 2023 & Text,Image & \url{https://github.com/PLUM-Lab/Mocheg} \\
		FakeSV \cite{qi2023fakesv} & 2023 & Text,Vedio & \url{https://github.com/ICTMCG/FakeSV} \\
		FACTIFY 2\cite{suryavardan2023factify} & 2023 & Text,Image & \url{https://github.com/surya1701/Factify-2.0?tab=readme-ov-file} \\
		MiRAGeNews \cite{huang2024miragenews} & 2024 & Text,Image & \url{https://huggingface.co/datasets/anson-huang/mirage-news} \\
		M$^{3}$A \cite{xu2024m3a} & 2024 & Text,Image,Audio,Vedio & \url{https://github.com/FinalYou/M3A?tab=readme-ov-file} \\
		HFFN \cite{jin2024fake} & 2024 & Text,Image & - \\
		VERITE \cite{papadopoulos2024verite} & 2024 & Text,Image & \url{https://github.com/stevejpapad/image-text-verification}\\
		FakeTT \cite{bu2024fakingrecipe} & 2024 & Text,Vedio & \url{https://github.com/ICTMCG/FakingRecipe/tree/main?tab=readme-ov-file} \\
		MFND \cite{zhu2025multimodal} & 2025 & Text,Image & \url{https://github.com/yunan-wang33/sdml} \\
		MMFakeBench \cite{liummfakebench} & 2025 & Text,Image & \url{https://huggingface.co/datasets/liuxuannan/MMFakeBench} \\
		MDAM$^{3}$-DB \cite{xu2025mdam3} & 2025 & Text,Image,Audio,Vedio & - \\
		DriftBench \cite{li2025drifting} & 2025 & Text,Image & -\\
		\bottomrule
	\end{tabular}
	\label{dataset_link}
\end{table*}

\subsection{Overview of Benchmark Datasets}

\textbf{Twitter15.}
The Twitter15 dataset \cite{liu2015real}, created by crawling two rumor-tracking websites, Snopes\footnote{snopes.com} and Emergent\footnote{emergent.info}, collects 2,299 news items published up to March 2015. After screening, it contains 94 true news items and 446 false news items. To obtain tweets related to an event, a keyword-based query was constructed, and a web crawler was used to obtain the complete history. This was then sampled and cross-checked by researchers. Furthermore, real events were collected using Twitter's free data stream and a clustering algorithm. The final dataset contains 421 true events and 421 false events.

\textbf{Twitter16.}
The Twitter16 dataset \cite{ma2016detecting}, based on Snopes\footnote{snopes.com}, an online rumor-debunking service, collects 778 events from March to December 2015, 64\% of which were rumors. Researchers extracted and optimized keywords to obtain relevant tweets, supplemented with public datasets, ultimately creating a balanced dataset containing 498 rumors and 494 non-rumors, which is widely used in rumor detection research.

\textbf{PHEME.}
The PHEME dataset \cite{zubiaga2016analysing}, constructed through a combination of automated crawling and manual verification, covers tweets, comments, and user interaction data, such as retweets and likes, related to specific events. Its multi-dimensional information structure and high-quality annotations, including the differentiation of rumor types, provide a rich resource for rumor detection, information dissemination analysis, and user behavior research, supporting machine learning model training and cross-event comparative research.

\textbf{LIAR.}
LIAR \cite{wang2017liar} is a multimodal fake news detection dataset constructed in 2017. It contains 12,836 manually annotated sentences from PolitiFact, covering a variety of contexts including news, speeches, interviews, advertisements, and social media. The dataset provides fine-grained authenticity annotations (six-category labels) and rich metadata, including speaker identity, context, and historical credibility. It is suitable for research in disinformation detection, credibility analysis, and multimodal modeling, providing an important benchmark resource in the field.

\textbf{FakeNewsNet.}
The FakeNewsNet dataset \cite{shu2017fake} is a multidimensional data repository containing two fact-checking datasets based on Politifact and Gossipcop, covering news content, social context, and spatiotemporal information. Constructed by the FakeNewsTracker system, this dataset aims to advance open research questions in the field of fake news research.

\textbf{Weibo.}
The Weibo dataset \cite{jin2017multimodal}, built from Weibo's official rumor-busting system and Xinhua News Agency news sources, covers verified rumor and non-rumor posts from 2012-2016. The dataset contains approximately 40,000 tweets with images, including text, images, and social context. The dataset is deduplicated and quality-screened using the LSH algorithm, and the training/test set is split into an 8:2 ratio based on event clustering to minimize data leakage. This dataset provides an authoritative and high-quality benchmark for multimodal rumor detection.

\textbf{GoodNews.}
The GoodNews dataset \cite{biten2019good}, sourced from The New York Times, contains over 460,000 news images, corresponding articles, and headlines, making it one of the largest multimodal news resources available. The average article length exceeds 650 words, 97\% of headlines contain named entities, and 68\% contain human names, fully demonstrating the complexity and semantic relevance of news text. This dataset not only supports news image and text generation and alignment tasks but also provides a core data source for building multimodal fake news detection benchmarks such as NeuralNews. Because approximately half of the news texts exceed 512 tokens, GoodNews presents significant challenges for long-text modeling, multimodal alignment, and named entity recognition.

\textbf{FA-KES.}
The FA-KES dataset \cite{salem2019fa}, constructed for diverse media coverage of the Syrian war, utilizes a semi-supervised annotation process and fact-checking mechanism to collect approximately 804 English news articles (with a near-balanced distribution of true and false news). Its rich structured information (including title, date, source, etc.) and reliable label generation process provide a solid foundation for fake news detection, especially for research on meta-learning, weak supervision, and multi-feature fusion models in few-shot scenarios. Its significant generalizability makes it suitable for dataset construction and algorithm validation in other military conflict scenarios.

\textbf{MultiFC.}
The MultiFC dataset \cite{salem2019fa}, collected from 26 English-language fact-checking websites, covers 34,918 naturally occurring factual claims, accompanied by supporting evidence, context, and rich metadata, all annotated by professional journalists. Its core features include the entities involved in the claims, contextual information, and multi-dimensional metadata. These additional attributes significantly improve model performance in automated claim verification tasks. MultiFC provides a solid foundation for the development and evaluation of fact-checking models.

\textbf{Fakeddit.}
The Fakeddit dataset \cite{nakamura2020fakeddit}, constructed from 22 Reddit subreddits, covers topics from politics to everyday life and contains over one million posts. It undergoes multi-stage review and employs distant supervision to provide 2/3/6 classification labels. The data includes text, images, metadata, and comments, with approximately 64\% of the samples being multimodal, supporting research on multimodal and hierarchical fake news detection.

\textbf{NeuralNews.}
The NeuralNews dataset \cite{tan2020detecting}, built on the GoodNews dataset, covers approximately 128,000 real and machine-generated news articles, divided into four categories (combinations of real/generated articles and headlines). The machine-generated content, generated by models such as Grover and entity-aware image caption models, includes text, images, and headline information. Its multimodal and fine-grained design provides a more realistic and challenging benchmark for detecting machine-generated news, and is of great significance for advancing research in multimodal fake news detection.

\textbf{MM-COVID.}
MM-COVID \cite{li2020mm} is a multilingual, multimodal dataset for COVID-19 fake news detection. It covers six languages and contains 3,981 fake news items and 7,192 real news samples. The data was collected from February to July 2020 and comes from sources including social media, traditional media, blogs, and fact-checking organizations. This dataset integrates multimodal content, preserves user interactions, and preserves timestamp information. It supports cutting-edge tasks such as cross-language detection, multimodal fusion, and social and spatiotemporal feature analysis, providing key support for building efficient and generalizable fake news detection models.

\textbf{CoAID.}
CoAID \cite{cui2020coaid} is a multimodal dataset for detecting COVID-19 health disinformation. It covers 4,251 news articles and statements, 296,000 user tweets and replies, and 926 social media posts collected from real-world online environments between December 2019 and July 2020. CoAID supports a variety of tasks, including multimodal fake news detection and social contextual propagation analysis, providing a critical data foundation for building efficient and generalizable disinformation identification models in real-world scenarios.

\textbf{Weibo21.}
The Weibo 21 dataset \cite{nan2021mdfend} was released in 2021 and covers 4,488 fake news and 4,640 real news from 9 different fields(i.e., Science, Military, Education, Disasters, Politics, Health, Finance, Entertainment, Society). The dataset was created by collecting fake news and real news on Sina Weibo between December 2014 and March 2021.

\textbf{NewsCLIPpings.}
The NewsCLIPpings dataset \cite{luo2021newsclippings}, built on VisualNews, addresses the problem of mismatched images and misleading headlines. By generating forged samples using multiple strategies and combining them with CLIP filtering, we have generated 1,111,828 balanced image-text pairs. This dataset is challenging and avoids unimodal bias, making it a popular tool for evaluating the performance of multimodal models in image-text inconsistency detection.

\textbf{COSMOS.}
COSMOS \cite{aneja2021cosmos} is a typical multimodal dataset focused on detecting contextual inconsistencies between images and captions to support research on disinformation detection. The dataset consists of 160,000 training samples, 40,000 validation samples, and 1,700 test samples. Each image is assigned up to 10 bounding boxes, and the data is primarily sourced from news websites and the Snopes platform. Unlike traditional annotation methods, COSMOS does not directly use contextual misuse annotations during training, but only incorporates them during evaluation. This significantly increases the challenge of the task and enhances the generalization capabilities of the model. This dataset provides a high-quality benchmark for multimodal disinformation detection, and is of great significance for comparative studies and method evaluation.

\textbf{CHECKED.}
CHECKED \cite{yang2021checked} is the first Chinese multimodal dataset for COVID-19-related fake news detection, covering 2,104 samples collected from Weibo between December 2019 and August 2020, including 344 fake news and 1,760 real news. The dataset integrates multimodal information such as text, images, and videos, and provides large-scale social context data, including more than 1.86 million reposts, 1.18 million comments, and 56.85 million likes. CHECKED provides important benchmark experimental support for multimodal fake news detection, propagation behavior analysis, and early identification.

\textbf{CHEF.}
CHEF \cite{hu2022chef} is a multi-domain Chinese dataset for evidence-based fact-checking, covering politics, public health, science, society, and culture. The dataset contains 10,000 manually verified true claims, each accompanied by manually curated and annotated evidence collected from the internet, including text and some images and videos, to ensure the reliability and accuracy of the annotations. The data source comes from six Chinese fact-checking and news websites (e.g., Piyao, TFC, MyGoPen, Jiaozhen, and Cnews), and is divided into a training set of 5,754 claims, a validation set of 666 claims, and a test set of 666 claims. Multiple rounds of annotation and verification ensure annotation consistency, with a Fleiss Kappa of 0.74.

\textbf{MC-Fake.}
The MC-Fake dataset \cite{min2022divide}, collected from Twitter, covers 28,334 news events across five major themes: politics, entertainment, health, COVID-19, and the Syrian War. This dataset is unique in that it provides both news text and rich social context, including tweets, retweets, replies, user attributes, and their social connections. Compared to existing content-based fake news benchmarks, MC-Fake complements existing content-based fake news benchmarks by incorporating social context and relationship networks. This provides a new experimental platform for studying the cross-modal characteristics of fake information and holds significant potential for open research.

\textbf{MuMiN.}
The MuMiN dataset \cite{nielsen2022mumin}, built on Twitter, covers 41 languages. It integrates multimodal information, including tweets, replies, users, images, articles, and hashtags, spanning over a decade and covering a wide range of topics and events. This dataset offers significant advantages in scale, diversity, and cross-linguality, providing a large-scale benchmark for studying cross-modal modeling, cross-lingual propagation, and long-term evolution of disinformation.

\textbf{DGM$^{4}$.}
The DGM$^{4}$ dataset ( \cite{shao2023detecting}) is a large-scale dataset for detecting and localizing multimodal media manipulation. Built on VisualNews, it covers real-world sources from The Guardian, BBC, USA Today, and The Washington Post. It contains 230,000 news image and text samples, including 77,426 pairs of original images, 152,574 pairs of manipulated images (e.g., face swaps, attribute edits, and text replacements), and 32,693 pairs of hybrid images. The dataset is constructed with a distribution of nine manipulation categories. All samples are annotated with fine-grained and sentimentally balanced annotations, making it more challenging than existing datasets and suitable for multimodal forgery detection and localization research.


\textbf{IFND.}
The IFND dataset \cite{sharma2023ifnd} covers multimodal news related to India between 2013 and 2021, totaling 56,868 text and image samples. Real news comes from mainstream media outlets such as Times Now News and The Indian Express, while false news is collected by authoritative fact-checking platforms such as Alt News and Boom Live and manually verified for labeling accuracy. To alleviate the category imbalance, researchers introduced an intelligent enhancement algorithm to generate semantically plausible false statements and, based on LDA topic modeling, categorized the news into five major categories: elections, politics, COVID-19, violence, and others. IFND is widely used for performance evaluation of machine learning and deep learning models, expanding the research boundaries of multimodal disinformation detection.

\textbf{MR$^{2}$.}
The MR$^{2}$ dataset \cite{hu2023mr2} consists of two subsets, Weibo and Twitter, covering both text and image news, and provides external evidence retrieved from the internet for both modalities. The dataset supports both Chinese and English, covers multiple fields including politics, society, technology, and entertainment, and contains a rich collection of text, images, and webpage information, enabling a more realistic reflection of the cross-platform rumor propagation and verification process. With its bilingual nature and multimodal design, MR2 provides an important benchmark for the training and evaluation of multimodal rumor detection models, making it particularly suitable for lightweight CNN experiments and content moderation scenarios.

\textbf{Mocheg.}
The Mocheg dataset \cite{yao2023end}, constructed from fact-checked claims from PolitiFact and Snopes, contains 15,601 truth-labeled claims, along with 33,880 text paragraphs and 12,112 images as evidence. This dataset is unique in that it provides both textual and image evidence, supporting, for the first time, evidence retrieval, multimodal fact-checking, and explanation generation within a single, end-to-end framework. As a key benchmark for multimodal fact-checking, Mocheg is not only suitable for validating model detection capabilities but also provides a systematic evaluation platform for interpretability research.

\textbf{FakeSV.}
The FakeSV dataset \cite{qi2023fakesv} is the largest Chinese short-video benchmark for multimodal fake news detection, containing rich social context information. It includes news video content, user comments, and publisher profiles, providing a comprehensive view of the dissemination environment of short video news. FakeSV enables multimodal analysis by jointly leveraging visual, textual, and social modalities, addressing the limited exploitation of multimodal correlations in prior works. In addition to fake/real labels, the dataset supports exploratory analysis of fake news propagation characteristics across content and social dimensions. To establish baseline performance, the authors further introduce a multimodal detection model named SV-FEND, which adaptively exploits cross-modal correlations and social contextual cues to enhance detection accuracy. Overall, FakeSV bridges the gap between multimodal content understanding and social behavior analysis, offering a valuable resource for future research on fake news detection in short video platforms.

\textbf{FACTIFY 2.}
The FACTIFY 2 dataset \cite{suryavardan2023factify} is a large-scale multimodal fact-checking benchmark that extends FACTIFY 1 by incorporating new data sources and adding satire news articles, resulting in 50,000 additional data instances. Each sample consists of textual claims paired with visual content, enabling research on multimodal verification. Similar to its predecessor, FACTIFY 2 categorizes samples into three broad labels—support, no-evidence, and refute—with sub-categories that reflect the entailment between text and image. The dataset provides a foundation for developing and evaluating models that jointly reason over textual and visual information. Baseline experiments using BERT and Vision Transformer architectures achieve a test F1 score of 65\%, highlighting the challenge of multimodal fact verification.

\textbf{MiRAGeNews.}
The MiRAGeNews dataset \cite{huang2024miragenews} is a multimodal benchmark designed specifically for AI-generated fake news detection. It contains 12,500 pairs of real and generated image-headline examples. Real examples are sourced from The New York Times articles in the TARA dataset, while the fake ones are generated using GPT-4 headlines and Midjourney V5.2 images, ensuring the content is highly realistic and misleading. This dataset includes training, validation, and cross-generator/publisher test sets, emphasizing the model's generalization capabilities.

\textbf{HFFN.}
Guided by the core principles of "human-centeredness" and "factual relevance," the HFFN dataset \cite{jin2024fake} comprises multimodal samples consisting of image-text pairs covering entertainment, sports, politics, and other fields. These samples are generated through image manipulation, text manipulation, and fact manipulation, and are accompanied by detailed human annotations. This significantly enhances the research value of fake news detection models in both authenticity judgment and manipulation reasoning.

\textbf{VERITE.}
VERITE \cite{papadopoulos2024verite} is a dataset designed specifically for multimodal disinformation detection, aiming to address the unimodal bias commonly found in existing datasets. Constructed from real-world news and social media image and text pairs, VERITE ensures that detection relies on cross-modal information rather than unimodal shortcuts by balancing modalities and eliminating asymmetric multimodal disinformation.

\textbf{M$^{3}$A.}
M$^{3}$A \cite{xu2024m3a} is a large-scale multimodal disinformation dataset covering text, images, audio, and video. Collected from 60 leading news outlets worldwide, the dataset contains 708,425 real news items and 6,566,386 fake news items. It provides multi-category, fine-grained topic and sentiment annotations and serves as a unified benchmark for various disinformation detection tasks, such as out-of-context detection and deepfake detection. This dataset aims to promote the development of robust multimodal disinformation analysis techniques.

\textbf{FakeTT.}
The FakeTT dataset \cite{bu2024fakingrecipe} is an English-language benchmark specifically designed for fake news detection on short video platforms. It contains a large collection of short videos paired with textual descriptions, user comments, and video metadata, capturing the rich but heterogeneous multimodal information inherent in short-form content. Unlike previous datasets that focus mainly on content analysis, FakeTT emphasizes the creative process behind video production, providing insights into material selection and editing patterns commonly found in fake news videos. In addition to real/fake labels, FakeTT supports research on creative-process-aware detection, enabling models to learn from sentimental, semantic, spatial, and temporal cues in video content. Together with the existing Chinese FakeSV dataset, FakeTT provides a comprehensive resource for developing and evaluating multimodal models that consider both content and production characteristics in fake news detection on short video platforms.

\textbf{MFND.}
The MFND dataset \cite{zhu2025multimodal} (released on May 11, 2025) contains 125,000 multimodal news samples across four combinations: real image-real text (RIRT), fake image-real text (FIRT), real image-fake text (RIFT), and fake image-fake text (FIFT). The fabricated content is constructed using 11 advanced generative techniques, including StyleGAN3 for synthesized images and multimodal large models (LVLMs) for text generation. In addition to true and fake labels, MFND also provides labels for image and text manipulation detection and precise annotation of manipulated image regions. Compared to existing datasets, MFND is more authentic to real-world communication scenarios and supports multi-task research such as detection and localization.

\begin{figure*}
	\centering
	\includegraphics[width=1\linewidth]{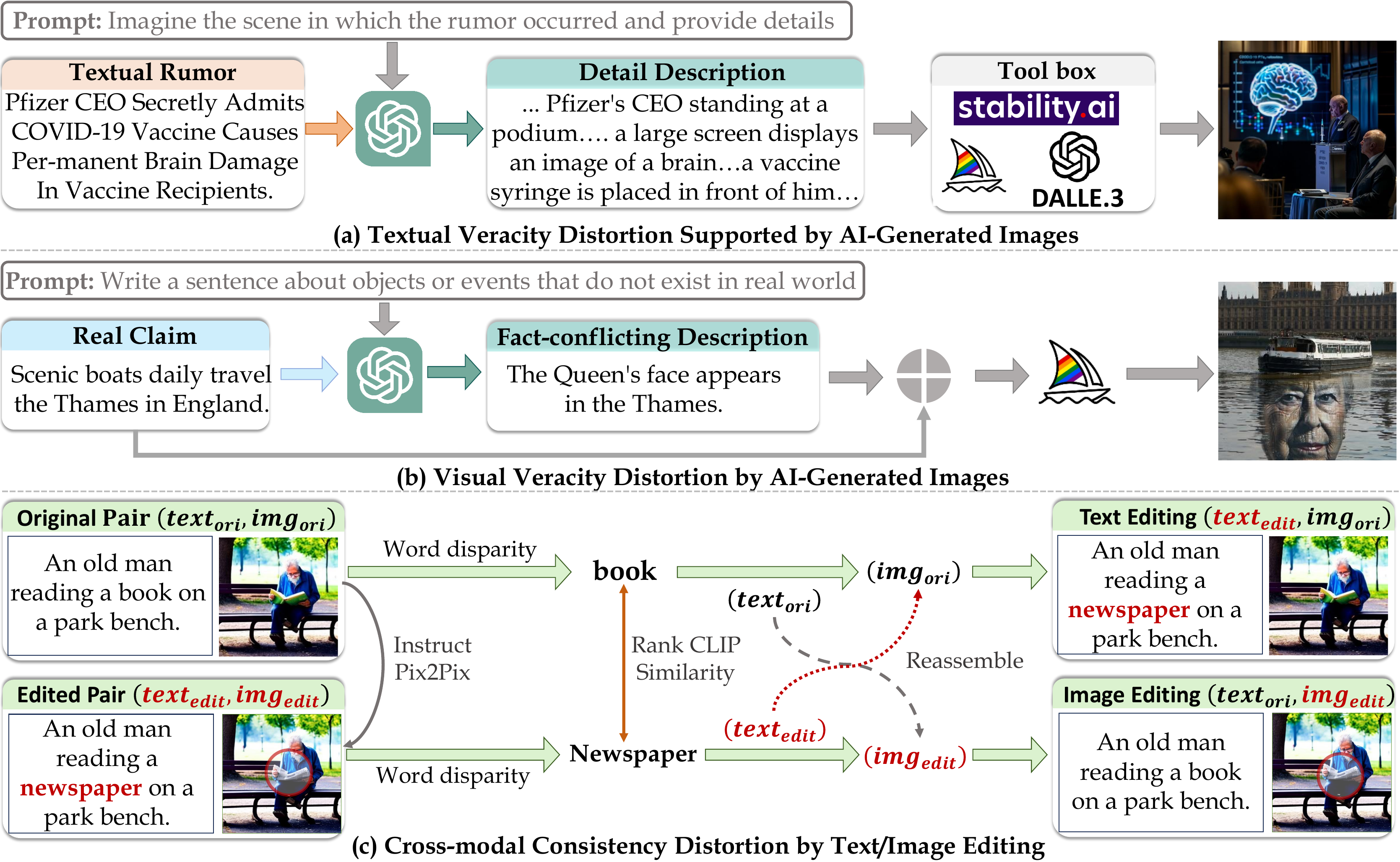}
	\caption{Instructions for constructing the MMFakeBench dataset. (a) AI-generated images support text authenticity distortion. (b) Visual authenticity distortion is caused by generating images that conflict with facts. (c) Cross-modal consistency distortion is caused by text/image editing \cite{liummfakebench}.}
	\label{fig:mmfakebench}
\end{figure*}

\textbf{MMFakeBench.}
As shown in Fig. \ref{fig:mmfakebench}, the MMFakeBench dataset \cite{liummfakebench} (released in 2025) collects multimodal disinformation samples from various sources, covering both image and text modalities. The dataset provides complete annotations for each sample, including text content, image path, source information, and binary and multi-class labels. As a mixed-source multimodal disinformation detection benchmark, MMFakeBench is particularly suitable for evaluating the performance of large-scale visual-language models in multimodal fake news detection.

\textbf{MDAM$^{3}$-DB.}
MDAM$^{3}$-DB \cite{xu2025mdam3} is a comprehensive multimodal disinformation detection benchmark consisting of 90,000 text, image, video, and audio samples. This dataset not only supports disinformation detection and modal inconsistency modeling across multimodal inputs, but also covers various deception scenarios, including AI-generated content, factual conflicts, and cross-modal mismatches. MDAM³-DB was built by integrating an interpretable analysis module driven by a large-scale visual-language model (LVLM). Its usability and practical performance were validated through systematic user studies. MDAM$^{3}$-DB provides a solid foundation for promoting more comprehensive, transparent, and trustworthy disinformation detection research.

\textbf{DriftBench.}
The DriftBench dataset \cite{li2025drifting} contains 16,000 news articles with images and text, covering six major topics. It focuses on simulating two typical scenarios: genre drift and evidence drift. Its tasks cover authenticity verification, adversarial evidence detection, and cross-variant consistency reasoning. DriftBench not only reveals the vulnerabilities of LVLM in the GenAI era but also provides a key benchmark for research on robustness and generalization methods.

\begin{table*}[H]
	\centering
	\caption{Comparison of misinformation datasets across three categories of distortion: textual veracity, visual veracity, and cross-modal consistency.}
	\resizebox{\linewidth}{!}{
		\tablestyle{5.5pt}{1.4}
		\begin{tabular}{lcccccccc}
			\shline
			\multirow{3}{*}{\textbf{Dataset}} & \multicolumn{3}{c}{\textbf{Textual Veracity Distortion}}             & \multicolumn{3}{c}{\textbf{Visual Veracity Distortion}}                                            & \multicolumn{2}{c}{\textbf{Cross-modal Consistency Distortion}}                                                                                                  \\ \cline{2-9} 
			& \multirow{2}{*}{\begin{tabular}[c]{@{}c@{}}Text\\ (Rumor)\end{tabular}} & \multicolumn{2}{c}{Supporting Image} & \multirow{2}{*}{\begin{tabular}[c]{@{}c@{}}Text\\ (Veracity)\end{tabular}} & \multicolumn{2}{c}{Fact-conflicting Image} & \multirow{2}{*}{\begin{tabular}[c]{@{}c@{}}Image/Text\\ Repurposing\end{tabular}} & \multirow{2}{*}{\begin{tabular}[c]{@{}c@{}}Image/Text\\ Editing\end{tabular}} \\ \cline{3-4} \cline{6-7}
			&                                & Repurposed      & AI-generated      &                                                                        & PS-edited  & AI-generated &                                                                                   &                                                                               \\ \shline
			Twitter15 \cite{liu2015real} & \textcolor{red}{\CheckmarkBold} & \XSolidBrush & \XSolidBrush & \XSolidBrush & \XSolidBrush & \XSolidBrush & \XSolidBrush & \XSolidBrush \\
			Twitter16 \cite{ma2016detecting} & \textcolor{red}{\CheckmarkBold} & \XSolidBrush & \XSolidBrush & \XSolidBrush  & \XSolidBrush  & \XSolidBrush & \XSolidBrush  & \XSolidBrush \\
			PHEME \cite{zubiaga2016analysing}                         & \textcolor{red}{\CheckmarkBold}                     & \XSolidBrush              & \XSolidBrush            & \XSolidBrush                                                                   & \XSolidBrush       & \XSolidBrush         & \XSolidBrush                                                                              & \XSolidBrush                                                                          \\
			LIAR \cite{wang2017liar}                            & \textcolor{red}{\CheckmarkBold}                  & \XSolidBrush              & \XSolidBrush            & \XSolidBrush                                                                   & \XSolidBrush       & \XSolidBrush         & \XSolidBrush                                                                              & \XSolidBrush                                                                          \\
			FakeNewsNet \cite{shu2017fake}                            & \textcolor{red}{\CheckmarkBold}                      & \XSolidBrush              & \XSolidBrush            & \XSolidBrush                                                                   & \XSolidBrush       & \XSolidBrush         & \XSolidBrush                                                                              & \XSolidBrush                                                                          \\
			Weibo \cite{jin2017multimodal}                           & \textcolor{red}{\CheckmarkBold}                     & \XSolidBrush              & \XSolidBrush            & \XSolidBrush                                                                   & \XSolidBrush       & \XSolidBrush         & \XSolidBrush                                                                              & \XSolidBrush                                                                          \\ 
			GoodNews \cite{biten2019good}                               & \textcolor{red}{\CheckmarkBold}                           & \XSolidBrush              & \XSolidBrush            & \XSolidBrush                                                                   & \XSolidBrush      & \XSolidBrush         & \XSolidBrush                                                                              & \XSolidBrush                                                                          \\
			FA-KES \cite{salem2019fa}                               & \textcolor{red}{\CheckmarkBold}                          & \XSolidBrush              & \XSolidBrush            & \XSolidBrush                                                                   & \XSolidBrush       & \XSolidBrush         & \XSolidBrush                                                                              & \XSolidBrush                                                                          \\
			MultiFC \cite{salem2019fa}                               & \textcolor{red}{\CheckmarkBold}                         & \XSolidBrush              & \XSolidBrush            & \XSolidBrush                                                                   & \XSolidBrush       & \XSolidBrush         & \XSolidBrush                                                                              & \XSolidBrush                                                                          \\
			Fakeddit \cite{nakamura2020fakeddit}                          & \XSolidBrush                           & \XSolidBrush              & \XSolidBrush            & \textcolor{red}{\CheckmarkBold}                                                                   & \textcolor{red}{\CheckmarkBold}       & \XSolidBrush         & \XSolidBrush                                                                              & \XSolidBrush                                                                          \\
			NeuralNews \cite{tan2020detecting}                          & \textcolor{red}{\CheckmarkBold}                        & \XSolidBrush              & \XSolidBrush            & \XSolidBrush                                                                  & \XSolidBrush       & \XSolidBrush         & \XSolidBrush                                                                              & \XSolidBrush                                                                          \\ 
			MM-COVID \cite{li2020mm}                         & \textcolor{red}{\CheckmarkBold}                         & \XSolidBrush              & \XSolidBrush            & \XSolidBrush                                                                  & \XSolidBrush       & \XSolidBrush         & \XSolidBrush                                                                              & \XSolidBrush                                                                          \\
			CoAID \cite{cui2020coaid}                         & \textcolor{red}{\CheckmarkBold}                           & \XSolidBrush              & \XSolidBrush            & \XSolidBrush                                                                    & \XSolidBrush        & \XSolidBrush         & \XSolidBrush                                                                              & \XSolidBrush                                                                          \\
			NewsCLIPpings \cite{luo2021newsclippings}                     & \XSolidBrush                           & \XSolidBrush              & \XSolidBrush            & \XSolidBrush                                                                   & \XSolidBrush       & \XSolidBrush         & \textcolor{red}{\CheckmarkBold}                                                                              & \XSolidBrush                                                                          \\
			COSMOS \cite{aneja2021cosmos}                            & \XSolidBrush                           & \XSolidBrush              & \XSolidBrush            & \XSolidBrush                                                                   & \XSolidBrush       & \XSolidBrush         & \textcolor{red}{\CheckmarkBold}                                                                               & \XSolidBrush                                                                          \\
			CHECKED \cite{yang2021checked}                              & \textcolor{red}{\CheckmarkBold}                            & \XSolidBrush              & \XSolidBrush            & \XSolidBrush                                                                   & \XSolidBrush       & \XSolidBrush         & \XSolidBrush                                                                              & \XSolidBrush                                                                          \\
			Weibo21 \cite{nan2021mdfend}                              & \textcolor{red}{\CheckmarkBold}                            & \XSolidBrush              & \XSolidBrush            & \XSolidBrush                                                                   & \XSolidBrush       & \XSolidBrush         & \XSolidBrush                                                                             & \XSolidBrush                                                                          \\
			CHEF \cite{hu2022chef}                              & \textcolor{red}{\CheckmarkBold}                          & \XSolidBrush              & \XSolidBrush            & \XSolidBrush                                                                   & \XSolidBrush       & \XSolidBrush         & \textcolor{red}{\CheckmarkBold}                                                                              & \XSolidBrush                                                                          \\
			MC-Fake \cite{min2022divide}                              & \textcolor{red}{\CheckmarkBold}                           & \XSolidBrush              & \XSolidBrush            & \XSolidBrush                                                                   & \XSolidBrush       & \XSolidBrush         & \textcolor{red}{\CheckmarkBold}                                                                              & \XSolidBrush                                                                          \\
			MuMiN \cite{nielsen2022mumin}                              & \textcolor{red}{\CheckmarkBold}                           & \XSolidBrush              & \XSolidBrush            & \XSolidBrush                                                                   & \XSolidBrush       & \XSolidBrush         & \textcolor{red}{\CheckmarkBold}                                                                              & \XSolidBrush                                                                          \\
			DGM$^4$ \cite{shao2023detecting}                              &  \XSolidBrush                         & \XSolidBrush              & \XSolidBrush            & \XSolidBrush                                                                                                                   & \XSolidBrush       & \XSolidBrush         & \XSolidBrush                                                                              & \textcolor{red}{\CheckmarkBold}                                                                          \\ 
			IFND \cite{sharma2023ifnd}                              & \textcolor{red}{\CheckmarkBold}                            & \XSolidBrush              & \XSolidBrush            & \XSolidBrush                                                                   & \XSolidBrush       & \XSolidBrush         & \XSolidBrush                                                                              & \textcolor{red}{\CheckmarkBold}                                                                          \\
			MR$^2$ \cite{hu2023mr2}                              & \textcolor{red}{\CheckmarkBold}                             & \XSolidBrush              & \XSolidBrush            & \XSolidBrush                                                                   & \XSolidBrush       & \XSolidBrush         & \XSolidBrush                                                                              & \XSolidBrush                                                                         \\
			Mocheg \cite{yao2023end}                              & \textcolor{red}{\CheckmarkBold}                            & \XSolidBrush              & \XSolidBrush            & \XSolidBrush                                                                   & \XSolidBrush       & \XSolidBrush         & \XSolidBrush                                                                              & \XSolidBrush                                                                         \\
			FakeSV \cite{qi2023fakesv}                              & \textcolor{red}{\CheckmarkBold}                            & \XSolidBrush              & \XSolidBrush            & \XSolidBrush                                                                   & \XSolidBrush       & \XSolidBrush         & \XSolidBrush                                                                              & \XSolidBrush                                                                         \\
			FACTIFY 2 \cite{suryavardan2023factify}                              & \textcolor{red}{\CheckmarkBold}                            & \XSolidBrush              & \XSolidBrush            & \XSolidBrush                                                                   & \XSolidBrush       & \XSolidBrush         & \XSolidBrush                                                                              & \XSolidBrush                                                                         \\
			MiRAGeNews \cite{huang2024miragenews}                              & \XSolidBrush                           & \textcolor{red}{\CheckmarkBold}              & \textcolor{red}{\CheckmarkBold}            & \textcolor{red}{\CheckmarkBold}                                                                   & \XSolidBrush        & \textcolor{red}{\CheckmarkBold}        & \textcolor{red}{\CheckmarkBold}                                                                              & \textcolor{red}{\CheckmarkBold}                                                                          \\
			M$^3$A \cite{xu2024m3a}                              & \XSolidBrush                           & \textcolor{red}{\CheckmarkBold}              & \textcolor{red}{\CheckmarkBold}            & \textcolor{red}{\CheckmarkBold}                                                                   & \textcolor{red}{\CheckmarkBold}       & \textcolor{red}{\CheckmarkBold}        & \textcolor{red}{\CheckmarkBold}                                                                              & \textcolor{red}{\CheckmarkBold}                                                                          \\
			HFFN \cite{jin2024fake}                              & \XSolidBrush                           & \textcolor{red}{\CheckmarkBold}              & \textcolor{red}{\CheckmarkBold}            & \textcolor{red}{\CheckmarkBold}                                                                   & \XSolidBrush       & \textcolor{red}{\CheckmarkBold}         & \textcolor{red}{\CheckmarkBold}                                                                              & \textcolor{red}{\CheckmarkBold}                                                                          \\
			VERITE \cite{papadopoulos2024verite}                              & \textcolor{red}{\CheckmarkBold}                          & \XSolidBrush              & \XSolidBrush            & \XSolidBrush                                                                   & \XSolidBrush       & \XSolidBrush         & \XSolidBrush                                                                              & \XSolidBrush  
			\\
			FakeTT \cite{bu2024fakingrecipe}                              & \textcolor{red}{\CheckmarkBold}                          & \XSolidBrush              & \XSolidBrush            & \XSolidBrush                                                                   & \XSolidBrush       & \XSolidBrush         & \XSolidBrush                                                                              & \XSolidBrush  \\
			MFND \cite{zhu2025multimodal}                              & \textcolor{red}{\CheckmarkBold}                           & \textcolor{red}{\CheckmarkBold}              & \textcolor{red}{\CheckmarkBold}            & \textcolor{red}{\CheckmarkBold}                                                                   & \textcolor{red}{\CheckmarkBold}       & \textcolor{red}{\CheckmarkBold}        & \textcolor{red}{\CheckmarkBold}                                                                             & \textcolor{red}{\CheckmarkBold}                                                                          \\
			MMFakeBench \cite{liummfakebench}                      & \textcolor{red}{\CheckmarkBold}                      & \textcolor{red}{\CheckmarkBold}              & \textcolor{red}{\CheckmarkBold}            & \textcolor{red}{\CheckmarkBold}                                                                   & \textcolor{red}{\CheckmarkBold}       & \textcolor{red}{\CheckmarkBold}         & \textcolor{red}{\CheckmarkBold}                                                                              & \textcolor{red}{\CheckmarkBold}                                                                          \\ 
			MDAM$^3$-DB \cite{xu2025mdam3}                              & \textcolor{red}{\CheckmarkBold}                           & \textcolor{red}{\CheckmarkBold}              & \textcolor{red}{\CheckmarkBold}            & \textcolor{red}{\CheckmarkBold}                                                                   & \textcolor{red}{\CheckmarkBold}       & \textcolor{red}{\CheckmarkBold}         & \textcolor{red}{\CheckmarkBold}                                                                              & \textcolor{red}{\CheckmarkBold}                                                                          \\
			DriftBench \cite{li2025drifting}                   &           \textcolor{red}{\CheckmarkBold}                           & \textcolor{red}{\CheckmarkBold}              & \textcolor{red}{\CheckmarkBold}            & \textcolor{red}{\CheckmarkBold}                                                                   & \textcolor{red}{\CheckmarkBold}       & \textcolor{red}{\CheckmarkBold}         & \textcolor{red}{\CheckmarkBold}                                                                              & \textcolor{red}{\CheckmarkBold}   \\\shline
		\end{tabular}
	}
	\label{dataset_comparision}%
	
\end{table*}%

\subsection{Comparison and Taxonomy of Misinformation Datasets}

To comprehensively understand the landscape of misinformation datasets, 
Table~\ref{dataset_comparision} categorizes 32 representative benchmarks 
across three major axes of veracity distortion: (i) Textual Veracity Distortion, 
(ii) Visual Veracity Distortion, and (iii) Cross-modal Consistency Distortion. 
Each axis captures different aspects of misinformation generation, ranging from textual fabrication to visual manipulation and semantic inconsistency between modalities. Specifically:

\textbf{(i) Textual Veracity Distortion.} This category involves falsified or manipulated textual content that misrepresents facts. 
Datasets such as Twitter15/16, LIAR, and FakeNewsNet primarily focus on this dimension, 
providing annotated claims and news statements labeled as true, false, or unverified. 
Some recent datasets, including FA-KES and GoodNews, extend this by incorporating 
AI-generated or repurposed text samples to simulate emerging misinformation patterns. 
These resources enable fine-grained studies on linguistic cues, contextual framing, and source reliability.

\textbf{(ii) Visual Veracity Distortion.}
The second dimension focuses on visual manipulations, including 
photo editing, compositional blending, and AI-synthesized imagery. 
Datasets such as Fakeddit, NeuralNews, and MM-COVID explicitly mark 
images that have been Photoshop-edited (PS-edited) or generated by diffusion and GAN-based models. 
Unlike textual datasets, these visual corpora are often multimodal, pairing each image with a corresponding textual caption or headline, 
thus supporting research in multimodal forgery detection and image-text alignment. 
This line of datasets becomes increasingly important with the rise of generative models producing hyper-realistic yet fabricated content.

\textbf{(iii) Cross-modal Consistency Distortion.}
Cross-modal distortion refers to semantic or factual mismatches between modalities, for example, when the text claims an event that the image does not depict. 
Datasets such as Weibo, FakeNewsNet, and COSMOS 
explicitly annotate image–text repurposing or editing inconsistencies, 
making them valuable for multimodal reasoning tasks. 
More recent datasets like M3FD and VERITE
further introduce controlled repurposing and image-text editing tasks to support explainable evaluation.

In summary, as summarized in Table~\ref{dataset_comparision},  most early rumor datasets (e.g., Twitter15/16, PHEME, LIAR) emphasize 
textual veracity classification, whereas later benchmarks gradually incorporate 
visual and cross-modal distortions to reflect real-world misinformation complexity. 
Notably, datasets such as MM-COVID, M3FD, and DriftBench 
cover all three dimensions, providing a holistic platform for evaluating multimodal reasoning and robustness. 
The trend indicates a paradigm shift from single-modality rumor detection 
toward comprehensive multimodal misinformation understanding.


\section{Evaluation Metrics}

In multimodal fake news detection, evaluation metrics not only serve as fundamental tools to assess model performance but also guide research progress. Unlike traditional text classification tasks, fake news detection involves high-risk, cross-modal, and adversarial characteristics. Therefore, its evaluation framework must systematically consider classification performance, robustness and generalization, cross-modal consistency, and interpretability and user perception. The following subsections provide a structured overview of commonly used metrics, along with mathematical definitions.

In the evaluation of multimodal fake news detection systems, particularly those framed as classification problems, standard performance metrics provide essential insights into model behavior across diverse and often imbalanced datasets. Given the binary or multi-class nature of fake news categorization (e.g., real vs.~fake, or fine-grained labels such as satire, misleading, fabricated), the following metrics are widely adopted to quantify predictive efficacy.

\textbf{Accuracy} measures the proportion of correctly classified instances among all predictions:
\begin{equation}
	\text{Accuracy} = \frac{\text{TP} + \text{TN}}{\text{TP} + \text{TN} + \text{FP} + \text{FN}}
\end{equation}
where TP, TN, FP, and FN denote true positives, true negatives, false positives, and false negatives, respectively. More specifically, TP is the number of samples that are actually "fake news" and are correctly judged as "fake news" by the model, TN is the number of samples that are actually "real news" and correctly judged as "real news" by the model, FP is the number of samples that are actually "real news" but are mistakenly judged as "fake news" by the model, FN is the number of samples that are actually "fake news" but are mistakenly judged as "real news" by the model. While intuitive, accuracy can be misleading under class imbalance, a common scenario in fake news datasets, where genuine news often dominates.

To address this limitation, Precision and Recall offer class-specific perspectives as follows:
\begin{equation}
	\text{Precision} = \frac{\text{TP}}{\text{TP} + \text{FP}}, \quad
	\text{Recall} = \frac{\text{TP}}{\text{TP} + \text{FN}}
\end{equation}
Precision reflects the reliability of positive predictions (i.e., how many flagged items are truly fake), whereas recall indicates coverage (i.e., how many actual fake instances were detected). In safety-critical applications like misinformation mitigation, high recall is often prioritized to minimize undetected disinformation.

\textbf{F1 Score} harmonizes precision and recall via their harmonic mean as follows:
\begin{equation}
	\text{F1} = 2 \cdot \frac{\text{Precision} \cdot \text{Recall}}{\text{Precision} + \text{Recall}}
\end{equation}
This single-value metric is especially informative for imbalanced settings, as it penalizes extreme disparities between precision and recall.

For multi-class fake news taxonomies (e.g., distinguishing clickbait, deepfake, conspiracy), Macro-F1 and Micro-F1 extend the F1 score to aggregate performance across all classes. \textbf{Macro-F1} computes the unweighted mean of per-class F1 scores, treating all classes equally regardless of size as follows:
\begin{equation}
	\text{Macro-F1} = \frac{1}{C} \sum_{i=1}^{C} \text{F1}_i
\end{equation}
where $C$ is the number of classes. This metric highlights model performance on minority classes, which is crucial when rare but high-impact misinformation types must be identified.

Conversely, \textbf{Micro-F1} aggregates TP, FP, and FN globally before computing precision and recall, effectively weighting each class by its frequency:
\begin{equation}
	\text{Micro-F1} = \frac{2 \cdot \sum_{i=1}^{C} \text{TP}_i}{2 \cdot \sum_{i=1}^{C} \text{TP}_i + \sum_{i=1}^{C} \text{FP}_i + \sum_{i=1}^{C} \text{FN}_i}
\end{equation}
Micro-F1 thus reflects overall instance-level performance and aligns closely with accuracy in balanced scenarios, but remains robust to label skew through its grounding in contingency counts.

\section{Experimental Performance}

	Due to the heterogeneous nature of multimodal misinformation detection benchmarks, this survey follows the original evaluation protocols, baseline selections, and metrics defined in the corresponding benchmark papers, rather than enforcing a unified setting across all datasets. The use of different proprietary LVLM baselines (e.g., GPT-4V in Tables \ref{MMD-Agent}–\ref{MDAM} and GPT-4o-mini in Table \ref{tab: main_results}) reflects dataset-specific evaluation objectives and historical contexts. In particular, GPT-4V is adopted in earlier benchmarks as a representative high-capacity LVLM, whereas GPT-4o-mini is used in DriftBench to reflect deployment-oriented evaluation under controlled diversity and distribution shift, consistent with the original benchmark design. Each table therefore constitutes a self-contained evaluation environment, where all compared methods share the same reference baseline, ensuring fair within-benchmark comparison. Similarly, evaluation metrics are retained as defined by each dataset to account for differences in task formulation, class imbalance, and annotation granularity. Metrics such as macro-F1, accuracy, AUC, and AP are selected to best reflect the primary evaluation goals of the corresponding benchmarks, and unifying them would risk obscuring task-specific performance characteristics. Accordingly, this survey does not aim to conduct direct numerical comparisons across different tables. Instead, cross-dataset analysis is performed at the level of modeling paradigms and reasoning mechanisms, while numerical comparisons are restricted to within-dataset settings under consistent protocols.

\subsection{Results on MM-FakeBench}

Table \ref{MMD-Agent} provides a comprehensive evaluation of binary detection performance for various large vision language models (LVLMs) on the MM-FakeBench benchmark, comparing both validation and large scale test settings. The results highlight the significant role of model size and inference strategies in detecting different types of misinformation distortions. 

Smaller models with 7B parameters generally struggle with performance, as reflected by macro-F1 scores below 50. These models exhibit limitations in addressing cross modal inconsistencies, which are critical for distinguishing between authentic and fabricated content. For instance, Otter-Image, MiniGPT4, and InstructBLIP show poor precision and recall, indicating their difficulty in recognizing relevant evidence across text and image modalities. This can be attributed to their inability to handle complex textual veracity distortion, which often involves rumor laden text that is challenging to separate from supporting images. When the model size increases to 13B parameters, performance improves significantly. Models like VILA and InstructBLIP achieve macro-F1 scores above 50 in validation, although they still fall short of human-level performance (F1 = 54.9). The introduction of the MMD-Agent framework notably enhances the models' ability to handle visual veracity distortions, such as fact conflicting images. The MMD-Agent framework employs a structured reasoning process that combines multi step reasoning and confidence calibration, helping the models to better differentiate genuine content from misleading visual cues. For example, VILA's F1 score increases from 51.1 to 56.5, and InstructBLIP’s score rises from 51.3 to 56.1, showing the substantial impact of these inference enhancements. At the 34B scale, Nous-Hermes-2 demonstrates strong performance with a macro-F1 of 62.9 in validation, which increases to 67.2 when paired with MMD-Agent. This shows that while model size contributes to performance, it is the targeted application of inference strategies like those in MMD-Agent that enables models to effectively address cross modal consistency distortion. MMD-Agent refines the model’s reasoning by enhancing the alignment between text and images, particularly when there are discrepancies such as image/text repurposing or AI-generated content. These refinements lead to more robust detection performance across various distortion types, ensuring the model better handles inconsistencies between the modalities. Finally, GPT-4V, a proprietary LVLM, sets a high benchmark with a macro-F1 of 72.3 in validation and 74.2 on the test set under standard prompting. When enhanced with the MMD-Agent framework, it reaches a macro-F1 of 74.0 in validation and 72.8 in testing, narrowly outperforming human evaluation. The improvements in both precision and recall reflect how MMD-Agent fine tunes the reasoning process to address specific distortions, particularly in visual textual misalignments, making the model more reliable and accurate in multimodal misinformation detection.

The results underscore that simply increasing model capacity is not enough. The integration of the MMD-Agent framework, with its advanced inference strategies such as multi step reasoning, confidence calibration, and adaptive visual amplification, consistently enhances model performance. This framework allows the models to effectively address different types of distortions, highlighting the importance of structured reasoning and adaptive refinement in achieving high accuracy in multimodal misinformation detection.

\begin{table*}[H]
	\centering
	\caption{Binary overall results of different models on the MM-FakeBench validation and test set with the comparison of standard prompting (Standard) and proposed MMD-Agent framework. The best results are \textbf{bolded}.}
	\label{model_comparision_bina}
	
	\resizebox{\linewidth}{!}{
		\tablestyle{5.0pt}{1.3}
		\begin{tabular}{lcccccccccc}
			\hline
			\multicolumn{1}{c}{}                                                                                &                                                                                     &                                                                                    & \multicolumn{4}{c}{\textbf{Validation (1000)}}                                            & \multicolumn{4}{c}{\textbf{Test (10000)}}                                                                                                     \\ \cline{4-11} 
			\multicolumn{1}{c}{\multirow{-2}{*}{\textbf{\begin{tabular}[l]{@{}c@{}}Model\\ Name\end{tabular}}}} & \multirow{-2}{*}{\textbf{\begin{tabular}[c]{@{}c@{}}Language\\ Model\end{tabular}}} & \multirow{-2}{*}{\textbf{\begin{tabular}[c]{@{}c@{}}Prompt\\ Method\end{tabular}}} & \textbf{F1}   & \textbf{Precision} & \textbf{Recall} & \textbf{ACC}                       & \textbf{F1}                          & \textbf{Precision}                   & \textbf{Recall}                      & \textbf{ACC}                \\ \hline
			\multicolumn{3}{c|}{\textit{Human Evaluation}}                                                                                                                                                                                                                                          & 54.9          & 56.6               & 57.8            & \multicolumn{1}{c|}{56.8}          & -                                    & -                                    & -                                    & -  \\
			\rowcolor{COLOR_MEAN}
			\multicolumn{11}{c}{\textbf{LVLMs with 7B Parameter}}                                                                                                                                                                                                                                                                                                                                                                                                                                                                \\
			Otter-Image \cite{liotter}                                                                                          & MPT-7B                                                                              & \multicolumn{1}{c|}{Standard}                                                            & 7.9           & 4.1                & 4.5             & \multicolumn{1}{c|}{7.9}           & 8.6                                  & 32.4                                 & 5.0                                  & 8.6                         \\
			MiniGPT4 \cite{zhu2024minigpt}                                                   & Vicuna-7B                                                                           & \multicolumn{1}{c|}{Standard}                                                            & 40.4          & 38.2               & 45.7            & \multicolumn{1}{c|}{63.1}          & 41.7                                 & 41.0                                 & 47.4                                 & 65.2                        \\
			InstructBLIP \cite{dai2023instructblip}                                      & Vicuna-7B                                                                           & \multicolumn{1}{c|}{Standard}                                                            & 14.7          & 30.8               & 13.2            & \multicolumn{1}{c|}{8.1}           & 16.1                                 & 40.5                                 & 14.2                                 & 8.8                         \\
			Qwen-VL \cite{liang2025damage}                          & Qwen-7B                                                                             & \multicolumn{1}{c|}{Standard}                                                            & 43.6          & 50.6               & 44.9            & \multicolumn{1}{c|}{60.3}          & 44.0                                 & 51.6                                 & 45.2                                 & 60.5                        \\
			VILA \cite{lin2024vila}          & LLaMA2-7B                                                                           & \multicolumn{1}{c|}{Standard}                                                            & 41.2          & 35.0               & 50.0            & \multicolumn{1}{c|}{70.0}          & 41.2                                 & 35.0                                 & 50.0                                 & 70.0                        \\
			PandaGPT \cite{su2023pandagpt}                            & Vicuna-7B                                                                           & \multicolumn{1}{c|}{Standard}                                                            & 24.6          & 60.6               & 50.5            & \multicolumn{1}{c|}{30.9}          & 24.1                                 & 61.7                                 & 50.4                                 & 30.6                        \\
			mPLUG-Owl2 \cite{ye2024mplug}                                        & LLaMA2-7B                                                                           & \multicolumn{1}{c|}{Standard}                                                            & 47.2          & 64.9               & 52.3            & \multicolumn{1}{c|}{70.6}          & 48.7                                 & 71.1                                 & 53.3                                 & 71.4                        \\
			BLIP2 \cite{li2023blip}                                    & FlanT5-XL                                                                           & \multicolumn{1}{c|}{Standard}                                                            & 41.2          & 35.0               & 50.0            & \multicolumn{1}{c|}{70.0}          & 41.2                                 & 35.0                                 & 50.0                                 & 70.0                        \\
			LLaVA-1.6 \cite{liu2023visual}                                     & Vicuna-7B                                                                           & \multicolumn{1}{c|}{Standard}                                                            & 48.1          & 48.2               & 48.5            & \multicolumn{1}{c|}{59.5}          & 52.5                                 & 53.0                                 & 52.6                                 & 62.5                        \\ \hline
			\rowcolor{COLOR_MEAN}
			\multicolumn{11}{c}{\textbf{LVLMs with 13B Parameter}}                                                                                                                                                                                                                                                                                                                                                                                                                                                               \\
			&                                                                                     & \multicolumn{1}{c|}{Standard}                                                            & 41.1          & 35.0               & 50.0            & \multicolumn{1}{c|}{70.0}          & 41.1                                 & 35.0                                 & 50.0                                 & 70.0                        \\
			\multirow{-2}{*}{VILA \cite{lin2024vila}}                                                                          & \multirow{-2}{*}{LLaMA2-13B}                                                        & \multicolumn{1}{c|}{MMD-Agent}                                                          &  \textbf{56.5} & \textbf{62.2}      & \textbf{56.9}   & \multicolumn{1}{c|}{\textbf{70.3}} & \textbf{56.6}                        & \textbf{64.3}                        & \textbf{57.2}                        & \textbf{71.2}               \\ \hline
			&                                                                                     & \multicolumn{1}{c|}{Standard}                                                            & 41.1          & 35.0               & 49.9            & \multicolumn{1}{c|}{\textbf{69.9}} & 41.1                                 & 35.0                                 & 49.9                                 & \textbf{69.8}               \\ 
			\multirow{-2}{*}{InstructBLIP \cite{dai2023instructblip}}                                 & \multirow{-2}{*}{Vicuna-13B}                                                        & \multicolumn{1}{c|}{MMD-Agent}                                                          & \textbf{51.3} & \textbf{53.4}      & \textbf{54.0}   & \multicolumn{1}{c|}{53.1}          & { \textbf{47.9}} & { \textbf{50.1}} & { \textbf{50.1}} & {49.9} \\ \hline
			&                                                                                     & \multicolumn{1}{c|}{Standard}                                                            & 31.6          & \textbf{63.4}      & 53.6            & \multicolumn{1}{c|}{35.5}          & 30.6                                 & \textbf{64.9}                        & 53.4                                 & 34.9                        \\
			\multirow{-2}{*}{BLIP2  \cite{li2023blip}}                                                                             & \multirow{-2}{*}{FlanT5-XXL}                                                        & \multicolumn{1}{c|}{MMD-Agent}                                                          & \textbf{51.5} & 53.4               & \textbf{54.0}   & \multicolumn{1}{c|}{\textbf{53.6}} & \textbf{51.8}                        & 54.0                                 & \textbf{54.7}                        & \textbf{53.5}               \\ \hline
			&                                                                                     & \multicolumn{1}{c|}{Standard}                                                            & 41.1          & 35.0               & 50.0            & \multicolumn{1}{c|}{69.7}          & 42.3                                 & 57.3                                 & 50.1                                 & 69.5                        \\
			\multirow{-2}{*}{LLaVA-1.6 \cite{liu2023visual}}                                                                         & \multirow{-2}{*}{Vicuna-13B}                                                        & \multicolumn{1}{c|}{MMD-Agent}                                                          & \textbf{51.8} & \textbf{66.7}      & \textbf{54.6}   & \multicolumn{1}{c|}{\textbf{71.4}} & \textbf{50.2}                        & \textbf{67.3}                        & \textbf{53.9}                        & \textbf{71.3}               \\ \hline
			\rowcolor{COLOR_MEAN}
			\multicolumn{11}{c}{\textbf{LVLMs with 34B Parameter}}                                                                                                                                                                                                                                                                                                                                                                                                                                                               \\
			&                                                                                     & \multicolumn{1}{c|}{Standard}                                                            & 62.9          & 67.1               & \textbf{70.0}   & \multicolumn{1}{c|}{63.4}          & 64.3                                 & 68.8                                 & \textbf{71.7}                        & 64.8                        \\
			\multirow{-2}{*}{LLaVA-1.6 \cite{liu2023visual}}                                                                         & \multirow{-2}{*}{\begin{tabular}[c]{@{}c@{}}Nous-Hermes-2\\ -Yi-34B\end{tabular}}   & \multicolumn{1}{c|}{MMD-Agent}                                                          & \textbf{67.2} & \textbf{70.4}      & 66.0            & \multicolumn{1}{c|}{\textbf{75.1}} & \textbf{68.1}                        & \textbf{71.1}                        & 67.0                                 & \textbf{75.6}               \\ \hline
			\rowcolor{COLOR_MEAN}
			\multicolumn{11}{c}{\textbf{Proprietary LVLMs}}                                                                                                                                                                                                                                                                                                                                                                                                                                                                               \\
			&                                                                                     & \multicolumn{1}{c|}{Standard}                                                            & 72.3          & 72.1               & 72.8            & \multicolumn{1}{c|}{75.6}          & \textbf{74.2}                        & \textbf{73.5}                        & \textbf{76.9}                        & \textbf{76.4}               \\
			\multirow{-2}{*}{GPT-4V \cite{yang2023dawn}}                                                                     & \multirow{-2}{*}{ChatGPT}                                                           & \multicolumn{1}{c|}{MMD-Agent}                                                          & \textbf{74.0} & \textbf{73.4}      & \textbf{75.5}   & \multicolumn{1}{c|}{\textbf{76.8}} & 72.8                                 & 72.4                                 & 75.4                                 & 75.0                        \\ \hline
		\end{tabular}
	}
	\label{MMD-Agent}
\end{table*}

\subsection{Results on Twitter and Fakeddit}

Table \ref{table:comparison} compares the performance of LEMMA with several baseline methods on the Twitter and Fakeddit datasets, both of which are widely used benchmarks for multimodal misinformation detection. The results underscore the importance of structured reasoning and modular architectures in improving performance across these datasets.

On the Twitter dataset, LEMMA achieves an accuracy of 0.824 and an F1 score of 0.816, significantly outperforming the best performing baseline, GPT-4V with Chain-of-Thought (CoT), which attains an accuracy of 0.757 and an F1 score of 0.758. This improvement is observed consistently across both rumor and non-rumor classification tasks. Specifically, in the rumor category, LEMMA achieves a precision of 0.943 and a recall of 0.741, compared to GPT-4V’s 0.866 and 0.670, respectively. In the non-rumor class, LEMMA maintains a precision of 0.721 and a recall of 0.937, demonstrating its superior balance and robustness. The results also highlight a common issue with baseline methods, such as FacTool, which show high recall but suffer from low precision. This indicates a tendency to overfit to superficial cues, such as sensational language or common visual patterns, leading to a high rate of false positives.

On the Fakeddit dataset, LEMMA maintains strong generalization performance with an accuracy of 0.828 and an F1 score of 0.857, once again outperforming all baseline methods. Even the powerful GPT-4V model, under direct prompting, achieves only an accuracy of 0.734 and an F1 score of 0.740. The CoT variant of GPT-4V improves slightly but still lags behind LEMMA by a significant margin. Notably, LEMMA’s performance remains stable across both datasets, with minimal variance in precision and recall, suggesting its effectiveness in capturing cross modal evidence without over-relying on domain-specific artifacts or superficial cues.

The ablation studies further validate LEMMA's design choices. Removing the initial stage inference module reduces accuracy to 0.781 on Twitter and 0.803 on Fakeddit, underscoring the critical role of early filtering in improving efficiency and reducing noise. Similarly, excluding visual retrieval results in a performance drop, highlighting the importance of integrating external visual knowledge to support decision making. These results confirm that LEMMA’s strength lies not only in its underlying model capacity but also in its modular architecture, which enables systematic evidence assessment and calibrated decision making. Compared to prior work, LEMMA avoids the instability observed in models like FacTool, which often exhibit sharp fluctuations in precision recall trade offs due to sensitivity to input phrasing or image composition. LEMMA, on the other hand, consistently maintains high scores across multiple metrics, demonstrating its ability to balance confidence and completeness in explanation generation. This balanced performance ensures that it minimizes both false positives and false negatives, making it particularly well suited for real world applications where reliability and fairness are paramount.

\begin{table*}[H]
	\centering
	\caption{Performance comparison of baseline methods Twitter and Fakeddit dataset. We show the result of eight different baseline methods. Additionally, we present the results of two ablation studies: one without initial-stage inference, and the other without resource distillation and evidence extraction. The best two results are \textbf{bolded} and \underline{underlined}.}
	\resizebox{1\textwidth}{!}{
		\resizebox{!}{0.8\textheight}{
			\begin{tabular}{clccccccc}
				\hline\hline
				\multirow{2}{*}{\textbf{Dataset}} & \multirow{2}{*}{\textbf{Method}} & \multirow{2}{*}{\textbf{Accuracy}} & \multicolumn{3}{c}{\textbf{Rumor}} & \multicolumn{3}{c}{\textbf{Non-Rumor}} \\
				\cmidrule(lr){4-6} \cmidrule(lr){7-9}
				& & & \textbf{Precision} & \textbf{Recall} & \textbf{F1} & \textbf{Precision} & \textbf{Recall} & \textbf{F1} \\
				\hline
				\multirow{12}{*}{\textit{Twitter}} &
				{\small Direct (LLaVA \cite{liu2023visual})} & 0.605 & 0.688 & 0.590 & 0.635 & 0.522 & 0.626 & 0.569 \\
				& {\small CoT (LLaVA \cite{liu2023visual})} & 0.468 & 0.563 & 0.231 & 0.635 & 0.441 & 0.765 & 0.560 \\
				& {\small Direct (InstructBLIP \cite{dai2023instructblip})} & 0.494 & 0.751 & 0.171 & 0.277 & 0.443 & 0.902 & 0.599 \\
				& {\small CoT (InstructBLIP \cite{dai2023instructblip})} & 0.455 & 0.813 & 0.067 & 0.112 & 0.428 & 0.921 & 0.596 \\
				& {\small Direct (GPT-4 \cite{achiam2023gpt})} & 0.637 & 0.747 & 0.578 & 0.651 & 0.529 & 0.421 & 0.469 \\
				& {\small CoT (GPT-4 \cite{achiam2023gpt})} & 0.667 & 0.899 & 0.508 & 0.649 & 0.545 & 0.911 & 0.682 \\
				& {\small FacTool (GPT-4 \cite{achiam2023gpt})} &  0.548 & 0.585 & \textbf{0.857} & 0.696 & 0.273 & 0.082 & 0.125 \\
				& {\small Direct (GPT-4V \cite{yang2023dawn})} & 0.757 & 0.866 & 0.670 & 0.756 & 0.673 & 0.867 & 0.758 \\
				& {\small CoT (GPT-4V \cite{yang2023dawn})} & 0.678  & 0.927 & 0.485 & 0.637 & 0.567 & \underline{0.946}& 0.709  \\
				\cline{2-9}
				& {\small \textbf{LEMMA} \cite{xuan2024lemma}} & \textbf{0.824} & \underline{0.943}& \underline{0.741} & \textbf{0.830} & \textbf{0.721} & 0.937& \textbf{0.816} \\
				
				& {\small \ w/o \textit{initial-stage infer}} & \underline{0.809}& 0.932& 0.736& \underline{0.823}& \underline{0.699}& 0.919& \underline{0.794}\\
				& {\small \ w/o \textit{visual retrieval}} & 0.781& \bf{0.953}& 0.672& 0.788& 0.652& \bf{0.949}& 0.773\\
				\hline
				\multirow{12}{*}{\textit{Fakeddit}} & {\small Direct (LLaVA)} & 0.663 & 0.588 & \underline{0.797} & 0.677 & 0.777 & 0.558 & 0.649 \\
				& {\small CoT (LLaVA \cite{liu2023visual})} & 0.673 & 0.612 & 0.400 & 0.484 & 0.694 & 0.843 & 0.761 \\
				& {\small Direct (InstructBLIP \cite{dai2023instructblip})} & 0.726 & 0.760 & 0.489 & 0.595 & 0.715 & 0.892 & 0.793 \\
				& {\small CoT (InstructBLIP \cite{dai2023instructblip})} & 0.610 & 0.685 & 0.190 & 0.202 & 0.604 & 0.901 & 0.742 \\
				& {\small Direct (GPT-4 \cite{achiam2023gpt})} & 0.677 & 0.598 & 0.771 & 0.674 & 0.776 & 0.606 & 0.680 \\
				& {\small CoT (GPT-4 \cite{achiam2023gpt})} & 0.691 & 0.662 & 0.573 & 0.614 & 0.708 & 0.779 & 0.742 \\
				& {\small FacTool (GPT-4 \cite{achiam2023gpt})} & 0.506 & 0.476 & \textbf{0.834} & 0.606 & 0.624 & 0.232 & 0.339 \\
				& {\small Direct (GPT-4V \cite{yang2023dawn})} & 0.734 & 0.673 & 0.723 & 0.697 & 0.771 & 0.742 & 0.764 \\
				& {\small CoT (GPT-4V \cite{yang2023dawn})} & 0.754 & \underline{0.858} & 0.513 & 0.642 & 0.720 & \bf{0.937}& 0.814 \\
				\cline{2-9}
				& {\small \textbf{LEMMA} \cite{xuan2024lemma}} & \textbf{0.828} & \textbf{0.881} & 0.706 & \textbf{0.784} & \bf{0.800}& \underline{0.925}& \textbf{0.857} \\
				
				& {\small \ w/o \textit{initial-stage infer}} & \underline{0.803}& 0.857 & 0.692 & \underline{0.766}& \underline{0.786}& 0.891 & 0.830 \\
				& {\small \ w/o \textit{visual retrieval}} & 0.792& 0.818& 0.675& 0.740& 0.778& 0.883& \underline{0.854} \\
				\hline\hline
			\end{tabular}
		}
	}
	\label{table:comparison}
\end{table*}

\subsection{Results on MDAM$^{3}$-DB}

Table \ref{MDAM} presents a comparison of various models on the MDAM$^{3}$-DB dataset, highlighting the performance improvements achieved by integrating the MDAM$^{3}$ framework across different misinformation types, including fact conflicting content, AI-generated content, offensive content, and out-of-context (OOC) information.

When analyzing the direct application of large visual-language models (LVLMs) without the MDAM$^{3}$ framework, we observe that models such as BLIP2 and InstructBLIP perform poorly on certain types of misinformation, particularly in detecting factual conflicts and AI-generated content. For instance, BLIP2 achieves an accuracy of 0.467 in detecting fact conflicting content and 0.355 for AI-generated content. Similarly, InstructBLIP shows limited performance, with an accuracy of 0.481 for fact conflicting content and 0.360 for AI-generated content. These results suggest that while these models perform reasonably well in identifying offensive content and cross modal inconsistencies, they struggle to verify factual authenticity reliably, often misclassifying synthetic content as real. This can be attributed to their lack of structured reasoning capabilities, which hinders their ability to make accurate distinctions in complex multimodal contexts. In contrast, GPT-4V demonstrates the strongest baseline performance across all tasks, particularly in handling complex semantics and multimodal information. However, the introduction of the MDAM$^{3}$ framework leads to substantial improvements. For example, in AI-generated content detection, GPT-4V's accuracy rises from 0.483 to 0.886, and its AUC improves from 0.491 to 0.924. These results underscore the critical role of MDAM$^{3}$ in correcting model biases, enhancing fact checking capabilities, and improving classification accuracy. The framework's ability to integrate external knowledge and apply structured reasoning is particularly effective in overcoming the challenges faced by models like BLIP2 and InstructBLIP, leading to notable improvements in their detection capabilities. For example, InstructBLIP’s accuracy for detecting fact conflicting content improves from 0.481 to 0.580, and BLIP2 sees a similar improvement, with its accuracy rising from 0.467 to 0.589. Additionally, MDAM$^{3}$ demonstrates remarkable versatility across multiple types of misinformation. Models like LLaVA and VILA show considerable improvements in both accuracy and AUC for detecting offensive and fact conflicting content after integrating MDAM$^{3}$. Notably, VILA's performance jumps from an accuracy of 0.484 to 0.730 in fact conflicting content detection, and LLaVA's accuracy increases from 0.503 to 0.712 for the same task. These improvements highlight MDAM$^{3}$'s ability to enhance model performance across a range of challenging misinformation types, reinforcing its adaptability and robustness. The ablation studies also provide insight into the specific contributions of MDAM$^{3}$. For instance, the removal of the MDAM$^{3}$ framework leads to significant drops in performance across all models, particularly in complex tasks like AI-generated content detection and fact conflicting content. These findings emphasize the importance of structured reasoning and external knowledge integration in improving detection accuracy and reducing model bias.

The MDAM$^{3}$ framework not only addresses the limitations of individual models in specific tasks but also incorporates external information through a structured reasoning process, resulting in significant improvements in the performance of multimodal misinformation detection systems. The observed gains across a variety of metrics, including accuracy, AUC, and AP, highlight that MDAM$^{3}$ offers a comprehensive and reliable solution for detecting and mitigating different types of misleading information.

\begin{table*}[H]
	\centering
	\renewcommand{\arraystretch}{1.3}
	\setlength{\tabcolsep}{5pt}
	\caption{Performance metrics for different misinformation types in MDAM$^{3}$-DB, comparing direct prompt queries with results obtained using the proposed MDAM$^{3}$ framework. The best two results are \textbf{bolded} and \underline{underlined}.}
	\begin{tabular}{clccccccccccccc}
		\toprule
		\multirow{2}{*}{Model} & \multirow{2}{*}{Process} 
		& \multicolumn{3}{c}{Fact-conflicting} 
		& \multicolumn{3}{c}{AI-generated} 
		& \multicolumn{3}{c}{Offensive} 
		& \multicolumn{3}{c}{OOC} \\ 
		\cmidrule(lr){3-5} \cmidrule(lr){6-8} \cmidrule(lr){9-11} \cmidrule(lr){12-14}
		& & Acc & AUC & AP & Acc & AUC & AP & Acc & AUC & AP & Acc & AUC & AP \\ 
		\midrule
		\midrule
		\multirow{2}{*}{InstructBLIP \cite{dai2023instructblip}}  & Direct & 0.481 & 0.492 & 0.498 & 0.360 & 0.389 & 0.385 & 0.612 & 0.639 & 0.634 & 0.493 & 0.496 & 0.496 \\
		& MDAM$^{3}$ & 0.580 & 0.644 & 0.638 & 0.766 & 0.798 & 0.797 & 0.746 & 0.783 & 0.779 & 0.607 & 0.654 & 0.652 \\ 
		\midrule
		\multirow{2}{*}{BLIP2 \cite{li2023blip}}  & Direct & 0.467 & 0.484 & 0.479 & 0.355 & 0.367 & 0.359 & 0.621 & 0.688 & 0.685 & 0.472 & 0.484 & 0.480 \\
		& MDAM$^{3}$ & 0.589 & 0.651 & 0.646 & 0.764 & 0.823 & 0.814 & 0.747 & 0.778 & 0.768 & 0.617 & 0.662 & 0.654 \\ 
		\midrule
		\multirow{2}{*}{LLaVA \cite{liu2024improved}}  & Direct & 0.503 & 0.512 & 0.509 & 0.398 & 0.402 & 0.399 & 0.631 & 0.696 & 0.688 & 0.622 & 0.639 & 0.635 \\
		& MDAM$^{3}$ & 0.712 & 0.761 & 0.760 & 0.787 & 0.852 & 0.849 & 0.795 & 0.848 & 0.839 & 0.728 & 0.773 & 0.761 \\ 
		\midrule
		\multirow{2}{*}{VILA \cite{lin2024vila}}  & Direct & 0.484 & 0.490 & 0.489 & 0.323 & 0.365 & 0.358 & 0.611 & 0.656 & 0.652 & 0.615 & 0.626 & 0.624 \\
		& MDAM$^{3}$ & 0.730 & 0.756 & 0.755 & 0.754 & 0.796 & 0.774 & 0.750 & 0.768 & 0.764 & 0.712 & 0.769 & 0.758 \\ 
		\midrule
		\multirow{2}{*}{GPT-4V \cite{yang2023dawn}}  & Direct & 0.611 & 0.684 & 0.667 & 0.483 & 0.491 & 0.488 & 0.744 & 0.754 & 0.747 & 0.637 & 0.664 & 0.646 \\
		& MDAM$^{3}$ & \textbf{0.853} & \textbf{0.912} & \textbf{0.908} & \textbf{0.886} & \textbf{0.924} & \textbf{0.919} & \textbf{0.891} & \textbf{0.895} & \textbf{0.894} & \textbf{0.729} & \textbf{0.784} & \textbf{0.776} \\ 
		\bottomrule
	\end{tabular}
	\label{MDAM}
\end{table*}

\subsection{Results on Pheme and Twitter16}

Table \ref{tab:mrcr-results} presents a performance comparison across different misinformation types on the Pheme and Twitter16 datasets, using various models including small language models (SLMs), large language models (LLMs) in both zero shot and few shot settings, and the hybrid LLM+SLM framework of MRCD.

SLM-based approaches, such as RoBERTa and FTT, demonstrate moderate performance on both datasets, with accuracy values of 0.754 on Pheme and 0.651 on Twitter16. These models struggle to generalize across newly emerging events due to their limited contextual understanding, particularly when facing dynamic misinformation that requires up-to-date knowledge. This is especially evident in their suboptimal performance on complex misinformation types, such as emerging rumors or evolving fake narratives, where deeper reasoning and contextual adaptation are essential. In contrast, LLM based methods, like Llama2-7B and GPT-3.5, even under few shot settings, exhibit limited adaptability, with accuracies remaining below 0.63 across both datasets. Their performance underscores the challenges LLMs face in rapidly evolving domains without explicit domain-specific supervision. Although these models show some improvement over SLMs, they remain less effective at handling nuanced and domain-specific misinformation compared to hybrid approaches that combine the strengths of both model types. The MRCD framework, by integrating LLMs and SLMs through multi round collaboration, achieves a substantial performance boost, particularly in the detection of dynamic and heterogeneous misinformation. For instance, integrating Llama3 with FTT results in the best performance, achieving an accuracy of 0.814 on Pheme and 0.794 on Twitter16, surpassing the strongest standalone SLM baseline by 7.4\% and 12.8\%, respectively. This improvement is not only observed in accuracy but also across precision, recall, and F1 scores, demonstrating MRCD’s balanced enhancement in both detection reliability and robustness. The synergy between LLMs’ broad generalization and reasoning capacity and SLMs’ domain specific precision allows MRCD to excel in detecting both emerging and context specific misinformation types, which single model paradigms struggle to address. Further analysis of the results shows that MRCD’s multi round collaborative learning mechanism enables dynamic retrieval and application of relevant knowledge, refining the detection process iteratively. This enables the system to continuously adapt to evolving misinformation, a crucial factor in real world fake news detection scenarios. The improvement of over 20\% in performance metrics, such as precision and recall, especially in the context of domain specific fake news detection, underscores the importance of such an approach. Additionally, the hybrid approach mitigates the shortcomings of single model methods, such as overfitting to superficial cues or failing to capture emerging trends in misinformation.

These results confirm that MRCD’s collaborative architecture, combining LLMs with SLMs, offers a powerful solution for tackling the complexities of dynamic and evolving misinformation. By dynamically retrieving and integrating the most relevant knowledge, MRCD not only improves detection performance but also ensures that fake news detection systems remain adaptable to new and emerging threats in rapidly changing information environments.

\begin{table*}[H]
	\centering
	\small
	\setlength{\tabcolsep}{5.5pt} 
	\renewcommand{\arraystretch}{1.2} 
	\caption{Performance comparisons for different types of misinformation on the Pheme and Twitter16 datasets are performed. Results are presented for various methods, including SLM, zero-shot and few-shot LLM, and a hybrid LLM+SLM approach. The best performance value for each metric is denoted by \textbf{bolded}.}
	\begin{tabular}{l|l|cccc|cccc}
		\toprule
		\multirow{2}{*}{\textbf{Category}} & \multirow{2}{*}{\textbf{Method}} & 
		\multicolumn{4}{c|}{\textbf{Pheme}} & 
		\multicolumn{4}{c}{\textbf{Twitter16}} \\
		\cmidrule(lr){3-6} \cmidrule(lr){7-10}
		& & Acc & Precision & Recall & F1 & Acc & Precision & Recall & F1 \\
		\midrule
		\multirow{4}{*}{SLM} 
		& RoBERTa \cite{liu2019roberta} & 0.714 & 0.777 & 0.695 & 0.734 & 0.644 & 0.649 & 0.641 & 0.645 \\
		& EANN \cite{wang2018eann} & 0.744 & 0.738 & 0.745 & 0.741 & 0.641 & 0.621 & 0.741 & 0.676 \\
		& M$^3$FEND \cite{zhu2023memory} & 0.746 & 0.747 & 0.746 & 0.746 & 0.642 & 0.608 & \textbf{0.816} & 0.697 \\
		& FTT \cite{zhou2025collaborative} & 0.754 & 0.748 & 0.764 & 0.756 & 0.651 & 0.649 & 0.720 & 0.683 \\
		\midrule
		\multirow{3}{*}{\makecell[l]{LLM\\(zero-shot)}} 
		& Llama2-7B \cite{wu2024fake} & 0.505 & 0.498 & 0.697 & 0.581 & 0.496 & 0.501 & 0.494 & 0.498 \\
		& Llama3-8B \cite{zhou2025collaborative} & 0.535 & 0.518 & 0.770 & 0.620 & 0.562 & 0.574 & 0.531 & 0.552 \\
		& GPT3.5 \cite{hong2024leveraging} & 0.503 & 0.500 & 0.714 & 0.586 & 0.583 & 0.585 & 0.571 & 0.578 \\
		\midrule
		\multirow{3}{*}{\makecell[l]{LLM\\(few-shot)}} 
		& Llama2-7B \cite{wu2024fake} & 0.528 & 0.511 & 0.894 & 0.650 & 0.590 & 0.598 & 0.584 & 0.591 \\
		& Llama3-8B \cite{maggini2024leveraging} & 0.549 & 0.524 & \textbf{0.961} & 0.679 & 0.622 & 0.607 & 0.717 & 0.658 \\
		& GPT3.5 \cite{hong2024leveraging} & 0.520 & 0.507 & 0.850 & 0.635 & 0.621 & 0.609 & 0.705 & 0.653 \\
		\midrule
		LLM+SLM & ARG\cite{hu2024bad} & 0.743 & 0.741 & 0.779 & 0.760 & 0.705 & 0.698 & 0.710 & 0.704 \\
		\midrule
		\multirow{4}{*}{MRCD}
		& Llama2+RoBERTa \cite{zhou2025collaborative} & 0.772 & 0.765 & 0.775 & 0.770 & 0.732 & 0.717 & 0.619 & 0.664 \\
		& GPT3.5+RoBERTa \cite{zhou2025collaborative} & 0.781 & 0.735 & 0.821 & 0.778 & 0.768 & 0.752 & 0.734 & 0.743 \\
		& Llama3+RoBERTa \cite{zhou2025collaborative} & 0.788 & 0.700 & 0.900 & 0.786 & 0.772 & 0.765 & 0.775 & 0.770 \\
		& \textbf{Llama3+FTT} \cite{zhou2025collaborative} & \textbf{0.814} & \textbf{0.788} & 0.841 & \textbf{0.814} & \textbf{0.794} & \textbf{0.768} & 0.782 & \textbf{0.774} \\
		\midrule
		\multirow{2}{*}{Improvements} 
		& \textit{Impr. RoBERTa \cite{liu2019roberta}} & +7.4\% & / & +20.5\% & +5.2\% & +12.8\% & +11.6\% & +13.4\% & +12.5\% \\
		& \textit{Impr. FTT \cite{zhou2025collaborative}} & +6.0\% & +4.0\% & +7.7\% & +5.8\% & +14.3\% & +11.9\% & +13.2\% & +12.1\% \\
		\bottomrule
	\end{tabular}
	\label{tab:mrcr-results}
\end{table*}

\subsection{Results on DriftBench}

Table \ref{tab: main_results} presents a comparison of various LVLM based misinformation detection methods under the DI\_OT setting within Controlled News Diversity. The results underscore the vulnerability of current systems to GenAI-driven content variation, which can significantly affect model performance, particularly when models are exposed to diversified news content with stylistic and semantic shifts.

Models like GPT-4o-mini and Claude-3.7-Sonnet, which perform well under realistic conditions, suffer notable declines in performance when exposed to diversified content. For example, GPT-4o-mini's accuracy drops from 83.3\% to 64.4\%, and its recall for real news falls sharply from 75.6\% to 35.7\%, illustrating how stylistic and semantic changes can severely impair factual verification. Similarly, Claude-3.7-Sonnet experiences a decline in accuracy from 88.3\% to 72.4\%, with a substantial drop in recall for fake news. These results emphasize a common challenge in current LVLMs, which is their inability to maintain reliable reasoning across diverse multimodal content, especially when the content deviates from the distribution encountered during training. In contrast, models like SNIFFER and LEMMA exhibit more stability in the face of content diversification, though their F1 scores still decline by more than 10\% across both real and fake categories. This indicates that while these models are somewhat resilient to content variation, they still struggle with reasoning consistency, particularly when handling complex distortions such as those induced by GenAI driven content heterogeneity. The integration of multilevel drift scenarios in DriftBench reveals a critical gap in current detection systems, which is their inability to effectively generalize across evolving distributions. Models that excel in controlled conditions such as GPT-4o-mini and Claude-3.7-Sonnet fail to maintain their performance when confronted with content that deviates stylistically or semantically. This underscores the need for new frameworks capable of adapting to such shifts, stabilizing cross-modal reasoning, and ensuring consistent misinformation detection in rapidly changing information environments. Notably, LEMMA and SNIFFER demonstrate relatively stable performance in these challenging scenarios. This resilience can be attributed to their architectural focus on robust evidence retrieval and adaptive reasoning processes. However, even these models face limitations when dealing with the broader variations introduced by GenAI, underscoring the necessity for frameworks that integrate multi modal, multi round reasoning and continuous adaptation. These findings point to the need for more robust multimodal misinformation detection systems capable of addressing the challenges posed by GenAI driven content diversification.

The results from DriftBench underscore the need for developing frameworks capable of mitigating multi level drift, stabilizing reasoning across diverse contexts, and enhancing the reliability of fake news detection in the face of dynamic, GenAI induced misinformation. While models like LEMMA outperform most others, they still highlight the gaps that must be addressed to achieve robust and generalizable misinformation detection across a wide range of content types.

\begin{table*}[H]
	\renewcommand\arraystretch{1}
	\setlength{\tabcolsep}{3 pt} 
	\small
	\begin{center}
		\caption{Performance comparison of LVLM-based multimodal misinformation detection methods under the \textit{DI\_OT} setting within Controlled News Diversity. Since \textit{DI\_OT} applies to both real and fake instances, we report results (in percentage) for this category as a representative example.}
		\begin{tabular}{cclllllll}
			\hline \hline
			\multirow{2}{*}{\textbf{Infer Type}} & \multirow{2}{*}{\textbf{Data Type}} & \multirow{2}{*}{\textbf{Accuracy}} &  \multicolumn{3}{c}{\textbf{Real}} & \multicolumn{3}{c}{\textbf{Fake}} \\ \cline{4-9}
			& & & \multicolumn{1}{c}{\textbf{Precision}} & \multicolumn{1}{c}{\textbf{Recall}} & \multicolumn{1}{c}{\textbf{F1}} & \multicolumn{1}{c}{\textbf{Precision}} & \multicolumn{1}{c}{\textbf{Recall}} & \multicolumn{1}{c}{\textbf{F1}} \\ \hline
			
			\multirow{2}{*}{\textbf{GPT-4o-mini} \cite{roumeliotis2025fake}} 
			& Realistic & 83.3 & 89.4 & 75.6 & 81.9 & 78.9 & 91.1 & 84.5 \\
			& Diversified & 64.4 \dec{18.9} & 83.7 \dec{6.0} & 35.7 \dec{39.9} & 50.0 \dec{31.9} & 59.1 \dec{19.8} & 93.1 \imp{2.0} & 72.3 \dec{12.2} \\ \hline
			
			\multirow{2}{*}{\textbf{Claude-3.7-Sonnet} \cite{yang2025realfactbench}} 
			& Realistic & 88.3 & 87.7 & 89.2 & 88.4 & 89.0 & 87.5 & 88.2 \\
			& Diversified & 72.4 \dec{15.9} & 86.1 \dec{1.6} & 53.5 \dec{35.7} & 66.0 \dec{22.4} & 66.3 \dec{22.7} & 91.4 \imp{3.9} & 76.8 \dec{11.4} \\ \hline
			
			\multirow{2}{*}{\textbf{Qwen-VL} \cite{liang2025damage}} 
			& Realistic & 73.7 & 67.6 & 91.1 & 77.6 & 86.3 & 56.3 & 68.2 \\
			& Diversified & 63.7 \dec{10.0} & 63.9 \dec{3.7} & 62.7 \dec{28.4} & 63.3 \dec{14.3} & 63.5 \dec{22.8} & 64.7 \imp{8.4} & 64.1 \dec{4.1} \\ \hline
			
			\multirow{2}{*}{\textbf{CMIE} \cite{li2025cmie}} 
			& Realistic & 90.9 & 88.8 & 93.6 & 91.1 & 93.2 & 88.2 & 90.6 \\
			& Diversified & 72.4 \dec{18.5} & 72.2\dec{16.6} & 72.8\dec{20.8} & 72.5\dec{18.6} & 72.6\dec{20.6} & 72.1\dec{16.1} & 72.3\dec{18.3} \\  \hline
			
			\multirow{2}{*}{\textbf{SNIFFER} \cite{qi2024sniffer}} 
			& Realistic & 84.3 & 78.4 & 93.2 & 85.1 & 92.3 & 76.2 & 83.5 \\
			& Diversified & 73.5 \dec{10.8} & 71.2 \dec{7.2} & 76.0 \dec{17.2} & 73.5 \dec{11.6} & 75.8 \dec{16.5} & 71.1 \dec{5.1} & 73.4 \dec{10.1} \\ \hline
			
			\multirow{2}{*}{\textbf{LEMMA} \cite{xuan2024lemma}} 
			& Realistic & 79.6 & 73.4 & 92.8 & 82.0 & 90.2 & 66.5 & 76.5 \\
			& Diversified &  68.9 \dec{10.7} & 64.7 \dec{8.7} & 83.1 \dec{9.7} & 72.7 \dec{9.3} & 76.3 \dec{13.9} & 54.7 \dec{11.8} & 63.7 \dec{12.8} \\ \hline \hline
			
		\end{tabular}
		\label{tab: main_results}
	\end{center}
\end{table*}

\subsection{Results on FakeSV and FakeTT}

Table \ref{table1} presents a performance comparison of various models, including both traditional unimodal and state of the art multimodal frameworks, on the FakeSV and FakeTT datasets. The results emphasize the importance of robust multimodal reasoning and the challenges posed by different types of distortions, such as visual manipulation, cross modal inconsistency, and AI generated content.

Models such as BLIP2 and InstructBLIP, which are general purpose vision language models (VLMs), demonstrate limited capability in handling complex cross modal reasoning. These models achieve accuracies of 70.85\% and 78.41\% on FakeSV, respectively, which reflects their difficulty in capturing subtle semantic and visual inconsistencies between text and images. The root cause of these limitations lies in their architecture, which lacks the specialized modules required for fine grained multimodal alignment. Specifically, these models struggle with detecting visual distortions, such as manipulated images, or reconciling discrepancies between textual claims and visual evidence, making them less effective at handling nuanced misinformation types. In contrast, more sophisticated multimodal models like TikTec and FANVM introduce explicit feature alignment mechanisms, improving their ability to detect cross modal inconsistencies. These models achieve better performance with accuracy scores of 73.06\% and 79.88\%, respectively, on FakeSV. However, they still fall short in addressing the challenges of fine grained factual verification. Their performance plateaus at around 86.90\% accuracy on FakeSV and 84.28\% on FakeTT, indicating that while they can capture basic cross modal correlations, they still struggle with more complex distortions like those introduced by generative adversarial techniques or cross modal contradictions. The FakeSV-VLM framework introduces a significant advancement with its Progressive Mixture of Experts (PMOE) architecture, which enables adaptive expert collaboration for multimodal authenticity verification. By leveraging specialized experts for different tasks, FakeSV-VLM demonstrates superior performance, achieving 90.22\% accuracy and 89.97\% macro-F1 on FakeSV, and 89.30\% accuracy with 87.98\% macro-F1 on FakeTT. These gains can be attributed to the system’s ability to dynamically select the most relevant experts for each specific misinformation task. The framework also incorporates an Alignment-driven Event Checking (ADEC) module, which captures subtle semantic and visual inconsistencies that are crucial for detecting fabricated content, particularly in short videos. This combination of expert reasoning and cross modal alignment allows FakeSV-VLM to outperform both traditional and LVLM based models by a large margin. The performance improvements in FakeSV-VLM can be explained by the integration of adaptive expert collaboration and the ability to handle multimodal distortions across different types of misinformation. The PMOE architecture allows the system to select specialized experts based on the task at hand, providing a tailored approach to each misinformation type, whether it involves visual manipulation, cross modal inconsistency, or semantic distortions. The ADEC module further enhances the system’s robustness by identifying and correcting subtle inconsistencies that traditional models fail to capture.

The results from FakeSV-VLM indicate that the combination of structured reasoning, dynamic expert collaboration, and deep multimodal alignment is crucial for achieving state of the art performance in multimodal misinformation detection. While general-purpose models like BLIP2 and InstructBLIP perform well in simpler scenarios, they struggle with the complexities introduced by evolving misinformation tactics. The success of FakeSV-VLM highlights the importance of integrating task-specific adaptability and fine-grained reasoning capabilities in future misinformation detection systems, ensuring that they can handle a wide range of distortion types with high accuracy and interpretability

\begin{table*}[H]
	\centering
	\renewcommand{\arraystretch}{1.3}
	\setlength{\tabcolsep}{12 pt}
	\caption{Performance comparison on two datasets. Best results are shown in \textbf{bold}.}
	\begin{tabular}{lcccccccc}
		\toprule[1.5pt]
		\textbf{Model} & \multicolumn{4}{c}{\textbf{FakeSV}} & \multicolumn{4}{c}{\textbf{FakeTT}} \\
		\cmidrule(lr){2-5} \cmidrule(lr){6-9}
		& ACC & M-F1 & M-P & M-R & ACC & M-F1 & M-P & M-R \\
		\midrule
		GPT-4o-mini \cite{roumeliotis2025fake} & 68.08 & 68.05 & 69.88 & 69.49 & 61.54 & 61.20 & 64.41 & 65.89 \\
		GPT-4.1-mini \cite{roumeliotis2025fake} & 70.30 & 70.25 & 70.61 & 70.87 & 49.16 & 48.54 & 62.50 & 59.70 \\
		Qwen2.5-VL \cite{bai2025qwen2} & 64.21 & 60.79 & 64.55 & 61.52 & 45.82 & 45.31 & 56.69 & 55.42 \\
		InternVL2.5 \cite{chen2024internvl} & 64.39 & 57.89 & 68.52 & 60.50 & 46.82 & 45.29 & 64.92 & 59.23 \\
		InternVL2.5-MPO \cite{chen2024internvl} & 65.13 & 61.07 & 66.46 & 62.12 & 43.14 & 40.84 & 61.90 & 56.23 \\
		\midrule
		ViT \cite{yuan2021tokens} & 70.85 & 70.66 & 70.64 & 70.91 & 64.88 & 62.59 & 62.54 & 63.80 \\
		BERT \cite{devlin2019bert} & 78.41 & 78.25 & 78.17 & 78.52 & 70.90 & 69.00 & 68.71 & 70.60 \\
		\midrule
		TikTec \cite{shang2021multimodal} & 73.06 & 72.79 & 72.73 & 72.93 & 66.56 & 65.55 & 66.50 & 68.62 \\
		FANVM \cite{choi2021topic} & 79.88 & 78.91 & 80.98 & 78.42 & 71.91 & 70.85 & 71.21 & 73.90 \\
		SV-FEND \cite{qi2023fakesv} & 80.81 & 80.19 & 81.08 & 79.84 & 77.26 & 75.55 & 74.94 & 77.13 \\
		FakingRecipe \cite{bu2024fakingrecipe} & 84.69 & 84.39 & 84.57 & 84.25 & 79.26 & 77.53 & 76.86 & 78.89 \\
		CA-FVD \cite{wang2025consistency} & 85.79 & 85.28 & 86.57 & 84.78 & 81.61 & 80.26 & 79.50 & 82.17 \\
		ExMRD \cite{hong2025following} & 86.90 & 86.52 & 87.31 & 86.13 & 84.28 & 83.13 & 82.27 & 85.19 \\
		\midrule
		FakeSV-VLM \cite{wang2025fakesv} & \textbf{90.22} & \textbf{89.97} & \textbf{90.55} & \textbf{89.64} & \textbf{89.30} & \textbf{87.98} & \textbf{87.80} & \textbf{88.17} \\
		\bottomrule[1.5pt]
	\end{tabular}
	\label{table1}
\end{table*}

\subsection{Results on Multiple In-Domain and Out-of-Domain Datasets}

Table~\ref{tab:main_results} compares the performance of TRUST-VL and other baseline VLMs across both in-domain and out-of-domain datasets, highlighting the ability of various models to handle different types of misinformation distortions, including textual, visual, and cross-modal inconsistencies. The results clearly demonstrate that while general purpose VLMs exhibit certain strengths, they face significant limitations in effectively addressing complex multimodal misinformation, particularly in the presence of advanced distortions like cross-modal inconsistencies or AI-generated manipulations.

Models such as BLIP2 and InstructBLIP, despite showing relatively high performance in some cases, struggle significantly with specific distortion types. These models often misclassify cross modal or AI-generated distortions as genuine content, leading to substantial drops in accuracy. For example, BLIP2’s performance on datasets like MOCHEG and VERITE is markedly lower, as it fails to capture the nuanced inconsistencies between text and supporting visual content, especially in cases where visual veracity is distorted. These results suggest that the standard training and prompting strategies used by these models are insufficient for tackling complex visual and cross modal distortions, highlighting a clear gap in their ability to adapt across varied distortion types. In contrast, TRUST-VL, which introduces the Question Aware Visual Amplifier (QAVA) module for adaptive visual representation enhancement, shows substantial improvements across all datasets, including both in-domain and out-of-domain settings. By dynamically adjusting the visual representations based on contextual textual cues, TRUST-VL excels in addressing both textual veracity distortions (e.g., rumor-laden text) and visual veracity distortions (e.g., fact conflicting images). Its performance on Factify2, achieving near perfect accuracy, underscores its ability to handle specific types of distortion, particularly in scenarios involving high level visual manipulations. The model’s average accuracy of 86.16\%, surpassing the second best model by +8.42\%, reflects the strength of its unified reasoning approach, which effectively integrates structured textual and visual reasoning. Moreover, TRUST-VL’s robustness is evident in out-of-domain datasets like MOCHEG and Fakeddit-M, where traditional models like GPT-4o show severe performance degradation. These models fail to generalize effectively when faced with data that deviate from their training distribution, particularly in domains with cross-modal inconsistencies. TRUST-VL, however, maintains strong performance in these settings, demonstrating that its training paradigm, which combines reasoning based instruction with multimodal alignment, allows it to generalize better to unseen data. This performance is further amplified by its integration of the QAVA module, which optimally balances textual and visual cues for more accurate cross modal reasoning, ensuring that the model adapts well across a wide range of distortions.

The stark contrast in performance between TRUST-VL and other models reveals the limitations of traditional approaches, particularly in their inability to maintain robustness across distortion types. The success of TRUST-VL highlights the necessity of integrated, adaptive models that not only scale across model sizes but also incorporate more sophisticated reasoning strategies. This work establishes a new benchmark for multimodal misinformation detection, setting a precedent for future models that aim to achieve both high accuracy and generalization capabilities across diverse distortion types.

\begin{table*}[H]
	\renewcommand{\arraystretch}{1.1}
	\setlength{\tabcolsep}{1.5 pt}  
	\small
	\centering
	\caption{Performance (\%) comparison between TRUST-VL and other baseline VLMs across in-domain and out-of-domain datasets. The best score is bolded, and the second-best score is underlined.}
	\label{tab:main_results}
	
	\begin{tabular}{lccccccccccccccc}
		\toprule
		\multirow{3}{*}{\textbf{Methods}} & \multirow{3}{*}{\textbf{Avg. Acc.}} 
		& \multicolumn{8}{c}{\textbf{In-Domain}} 
		& \multicolumn{6}{c}{\textbf{Out-of-Domain}} \\
		\cmidrule(lr){3-10} \cmidrule(lr){11-16}
		& & \multicolumn{2}{c}{\textbf{MMFakeBench}} & \multicolumn{2}{c}{\textbf{Factify2}} & \multicolumn{2}{c}{\textbf{DGM$^4$-Face}} & \multicolumn{2}{c}{\textbf{NewsCLIPpings}} 
		& \multicolumn{2}{c}{\textbf{MOCHEG}} & \multicolumn{2}{c}{\textbf{Fakeddit-M}} & \multicolumn{2}{c}{\textbf{VERITE}} \\
		\cmidrule(lr){3-4} \cmidrule(lr){5-6} \cmidrule(lr){7-8} \cmidrule(lr){9-10}
		\cmidrule(lr){11-12} \cmidrule(lr){13-14} \cmidrule(lr){15-16}
		& & Acc. & F1 & Acc. & F1 & Acc. & F1 & Acc. & F1 & Acc. & F1 & Acc. & F1 & Acc. & F1 \\
		\midrule
		
		\multicolumn{10}{l}{\textcolor{gray!80}{\textit{\textbf{General-purpose VLMs}}}} \\
		
		BLIP2 \cite{li2023blip} & 53.36 & 37.40 & 34.45 & 54.30 & 42.38 & 47.70 & 34.35 & 50.14 & 34.28 & 62.50 & 57.16 & 70.75 & 70.19 & 50.75 & 37.35 \\
		InstructBLIP \cite{dai2023instructblip} & 58.41 & 57.30 & 56.38 & 66.83 & 66.48 & 50.40 & 48.66 & 53.85 & 50.71 & 63.25 & 60.85 & 64.75 & 62.83 & 52.50 & 49.60 \\
		LLaVA \cite{liu2024improved} & 60.25 & 62.60 & 61.72 & 79.59 & 79.10 & 46.41 & 38.14 & 45.87 & 48.54 & 66.50 & 64.71 & 68.00 & 66.67 & 52.75 & 49.80 \\
		xGen-MM & 62.20 & 65.40 & 62.77 & 86.03 & 86.04 & 50.10 & 49.68 & 59.87 & 59.18 & 59.50 & 56.32 & 60.00 & 53.45 & 54.50 & 54.41 \\
		LLaVA-NeXT \cite{zheng2025predictions} & 62.35 & 71.60 & 65.99 & 79.60 & 79.09 & 53.40 & \underline{52.21} & 59.86 & 59.37 & 58.25 & 52.52 & 59.00 & 52.36 & 54.75 & 54.57 \\
		Qwen2-VL \cite{Qwen2VL} & 69.85 & 67.00 & 66.28 & 89.40 & 89.37 & 48.10 & 41.63 & 70.94 & 69.91 & 66.25 & 64.57 & \underline{77.25} & \underline{76.96} & 70.00 & 68.94 \\
		GPT-4o \cite{roumeliotis2025fake} & 76.16 & 83.10 & 80.88 & 88.37 & 88.21 & \underline{57.14} & 49.24 & 86.51 & 86.51 & 77.00 & 76.81 & 73.50 & 73.12 & 67.50 & 67.57 \\
		o1 \cite{jaech2024openai} & \underline{77.74} & \underline{83.90} & \underline{82.41} & \underline{96.90} & \underline{96.90} & 50.06 & 38.06 & 86.80 & 86.54 & \underline{81.50} & \underline{81.38} & 73.25 & 73.07 & 71.75 & 71.66 \\
		\midrule
		
		\multicolumn{10}{l}{\textcolor{gray!80}{\textit{\textbf{Misinformation Detectors}}}} \\
		
		MMD-Agent \cite{liummfakebench} & 56.11 & 69.10 & 48.68 & 71.03 & 69.35 & 48.30 & 48.29 & 53.06 & 41.12 & 54.25 & 43.72 & 42.25 & 42.24 & 54.75 & 47.00 \\
		SNIFFER \cite{qi2024sniffer} & 61.17 & 51.40 & 51.33 & 61.00 & 55.97 & 47.20 & 37.96 & \underline{88.85} & \underline{88.85} & 53.75 & 50.73 & 53.50 & 51.13 & \underline{72.50} & \underline{72.02} \\
		LRQ-FACT \cite{beigi2024lrq} & 66.60 & 71.30 & 74.00 & 86.63 & 89.79 & 41.80 & 44.14 & 68.19 & 73.45 & 66.25 & 69.25 & 67.25 & 71.77 & 64.75 & 68.32 \\
		TRUST-VL \cite{yan2025trust} & \textbf{86.16} & \textbf{87.30} & \textbf{85.42} & \textbf{99.50} & \textbf{99.50} & \textbf{88.50} & \textbf{88.39} & \textbf{90.35} & \textbf{90.35} & \textbf{82.75} & \textbf{82.58} & \textbf{82.50} & \textbf{82.20} & \textbf{73.75} & \textbf{73.61} \\
		\bottomrule
	\end{tabular}
\end{table*}

\section{Future Directions}

While Large Vision-Language Models (LVLMs) have demonstrated remarkable capabilities in understanding and reasoning over multimodal content, their application to the nuanced and adversarial domain of fake news detection remains nascent and fraught with unresolved challenges. Drawing from the limitations and gaps identified in this survey, we outline several high-impact research directions that are critical to advancing the field toward robust, trustworthy, and deployable systems.

\subsection{Causal and Counterfactual Reasoning for Explainable Detection}

Current large vision-language models (LVLMs) detect fake news by learning statistical patterns from massive image–text datasets. While effective on benchmark tasks, these models often rely on superficial correlations rather than genuine evidential reasoning. For example, an LVLM may incorrectly associate certain visual styles, such as low-resolution images, particular color tones, or recurring backgrounds, with “fakeness” merely because these features frequently appear in the false examples within its training data. Similarly, it may link emotionally charged language (e.g., “shocking,” “you won’t believe”) or particular news sources to misinformation, even when the underlying claim is factually accurate. This reliance on non-causal cues leads to two critical failures. First, the model’s decisions become brittle: when confronted with novel manipulation tactics or content from underrepresented domains, performance degrades sharply. Second, and more importantly, the model lacks the ability to explain why a piece of content is classified as false based on concrete inconsistencies between visual and textual evidence, such as mismatched timestamps, manipulated objects, or claims that contradict the visible context. Without such reasoning, the system remains a black box, unsuitable for high-stakes applications like journalism, legal review, or public policy. Standard LVLMs are optimized to maximize predictive accuracy on observed data without distinguishing between causal features that make a claim false and correlated features that simply co-occur with falsity in the training distribution. This limitation arises from inherent biases in real-world fake news datasets, which mirror historical disinformation patterns, platform-specific moderation rules, and cultural contexts. As a result, the model often learns superficial associations, such as treating content from certain regions or containing protest imagery as deceptive, rather than reasoning about whether the visual evidence genuinely contradicts the accompanying text. In essence, today’s LVLMs answer the question “What usually goes with fake news?” instead of “What proves this specific claim is false?”

To address this limitation, future LVLMs should be explicitly designed to perform causal and counterfactual reasoning, enabling them to identify which elements of the multimodal input contribute to the veracity judgment and to evaluate how modifying those elements would affect the final decision. A crucial step is to disentangle causally relevant signals, such as temporal inconsistencies, object mismatches, or semantic contradictions, from confounding factors like source bias, stylistic patterns, or demographic correlations. This can be accomplished by training LVLMs across diverse environments, including news from different countries, platforms, or time periods, and ensuring that their internal representations of evidence remain consistent even when confounding factors vary. Methods such as invariant risk minimization and domain-adversarial training can be adapted to reduce the model’s dependence on environment-specific cues while maintaining its ability to detect genuine cross-modal factual inconsistencies. Another effective strategy is to enhance training with counterfactual examples, which are modified versions of real multimodal samples where only non-essential attributes are changed. For example, an image in a fake news instance can be replaced with a visually similar but factually neutral photo while keeping the misleading text intact. A causally aware model should retain its “fake” prediction only if the textual claim remains unsupported by the new image. Conversely, if a true claim is paired with a manipulated image implying falsity, the model should detect the inconsistency and revise its judgment accordingly. Creating such counterfactual examples through image editing, text paraphrasing, or retrieval from verified media archives offers direct supervision for teaching the model to focus on the truly causal factors underlying veracity assessment. Finally, advancing in this direction requires the development of evaluation metrics that go beyond traditional accuracy. A causally competent model should demonstrate counterfactual consistency, meaning its predictions change only when causally relevant evidence is altered and remain stable when irrelevant features such as font style or background scenery are modified. Furthermore, the model’s explanations, including the image regions it highlights and the textual phrases it emphasizes, should align with human judgments of what constitutes decisive evidence. Collaborations with fact-checking organizations can support the creation of such evaluation benchmarks, ensuring that models are evaluated on genuine reasoning ability rather than statistical pattern recognition.

\subsection{Adversarially Robust and Out-of-Distribution Generalization}

Fake news is not a static phenomenon but an adversarial and adaptive process. Malicious actors continuously refine their strategies to evade detection, exploiting the very patterns that current LVLMs depend on. Early misinformation often involved simple image–text mismatches, whereas modern disinformation campaigns employ more sophisticated techniques such as context swapping, where real images are reused with fabricated captions; deepfakes with semantic alignment, which generate synthetic media that coherently supports false narratives; and multimodal prompt injection, which crafts inputs that subtly mislead LVLMs toward incorrect conclusions. These tactics are intentionally designed to maintain surface-level plausibility while eroding factual integrity, revealing the inherent fragility of existing models. Compounding this issue is the distributional shift inherent in real-world deployment. LVLMs trained on historical datasets, which are often dominated by English-language content from a few major platforms, tend to struggle when confronted with news from underrepresented regions, emerging social media formats (e.g., short videos, memes), or novel event types (e.g., pandemics, geopolitical crises). In such out-of-distribution (OOD) scenarios, models frequently default to spurious heuristics or fail silently, producing high-confidence but incorrect predictions. The root cause lies in the passive learning paradigm of most LVLMs: they are optimized for average-case performance on fixed, curated datasets, not for worst-case robustness or adaptability. Their representations are highly sensitive to input perturbations that preserve semantics to humans but alter model-internal features, such as minor color shifts, object repositioning, or synonym substitutions. Moreover, because training data rarely includes examples of how fake news evolves over time, models lack mechanisms to recognize or adapt to new manipulation strategies. Crucially, standard evaluation protocols mask this vulnerability. Benchmarks like FakeNewsNet or Weibo21 consist of static snapshots of past misinformation, offering no test of a model’s ability to generalize to future or unseen attack vectors. As a result, reported performance often overestimates real-world effectiveness.

To address these challenges, future work must move beyond static training and embrace proactive robustness by design. Below are several interlinked strategies that can significantly improve generalization under adversarial pressure and distribution shifts. First, LVLMs should be trained not only on clean examples but also on realistic adversarial variants that mimic actual disinformation tactics. This requires developing perturbation models tailored to multimodal content. Specifically, for images, apply context-preserving modifications such as object removal or replacement using diffusion-based inpainting, adjustments to lighting or color tone, and the generation of subtle deepfakes that maintain overall scene coherence. For text, apply meaning-preserving manipulations such as paraphrasing that introduces subtle false implications, substituting key entities with misleading alternatives, or inserting plausible yet fabricated details to distort the narrative while maintaining fluency. For cross-modal alignment, introduce semantic drift in which the image and text remain individually coherent but become jointly deceptive, such as pairing a real photo of a flood with a claim about a different location. By training on such augmented data, ideally produced through an iterative red-team and blue-team process, LVLMs can gradually learn invariant features that remain robust against manipulation. A more principled strategy is to ensure that the model’s internal representation of “fakeness” relies only on features that remain consistent across diverse environments, such as different news sources, languages, or event types. This can be accomplished by partitioning the training data into multiple environments, for example based on platform, country, or time period, and applying algorithms such as invariant risk minimization or domain-conflict-aware optimization. These methods encourage the model to build decision boundaries that depend on signals consistently associated with veracity across all environments rather than on environment-specific artifacts. The resulting representation captures the fundamental essence of multimodal inconsistency rather than the peculiarities of any single dataset. Finally, robust systems must know when they don’t know. LVLMs should be equipped with calibrated uncertainty estimation, for instance, via ensemble methods, Monte Carlo dropout, or density-based out-of-distribution scoring. When encountering inputs that deviate significantly from the training manifold (e.g., a novel meme format or a deepfake with unusual artifacts), the model should flag high uncertainty and defer to human reviewers or external verification tools, rather than outputting a confident but erroneous verdict.

\subsection{Efficient and Modular Architectures for Real-Time Deployment}

While recent large vision-language models (LVLMs) have shown impressive performance on curated fake news detection benchmarks, their practical usefulness in real-world content moderation remains severely limited. The main challenge lies not only in accuracy but also in computational efficiency, latency, and adaptability under operational constraints. Social media platforms must process millions of multimodal posts every minute, including images, short videos, memes, and text captions, often on edge devices or within strict response-time limits of less than 500 milliseconds per post. Current monolithic LVLMs, which contain billions of parameters and rely on complex cross-modal fusion mechanisms, are poorly suited for such environments. They demand costly GPU clusters, consume large amounts of energy, and lack the flexibility and transparency required for timely updates or inspection when new manipulation tactics appear. Moreover, the “one-size-fits-all” architecture of most LVLMs ignores the heterogeneity of real-world content. A breaking news photo, a satirical meme, and a deepfake video demand fundamentally different verification strategies, yet today’s models apply the same computationally intensive pipeline to all inputs, wasting resources on low-risk or obviously benign content. This mismatch between research prototypes and deployment realities creates a critical gap: the most accurate models are too slow to be useful, while fast heuristics lack the reasoning depth needed for nuanced disinformation.

To bridge this gap, future systems must abandon the paradigm of “bigger is better” in favor of intelligent, modular, and adaptive architectures that allocate computational resources only when and where they are needed. This requires rethinking both the structure of the model and the flow of inference. Instead of applying a full LVLM to every post, a more scalable solution is to adopt a multi-stage verification cascade. In the first stage, lightweight screening models such as distilled transformers or vision-text co-occurrence filters quickly identify obviously benign content like personal photos and product advertisements, as well as clear violations such as known deepfake signatures or blacklisted sources. These models operate within milliseconds and can filter out more than 80 percent of total traffic. The second stage involves specialized verifiers that are activated only for ambiguous or high-risk cases. For example, a temporal inconsistency detector can be used when a claim references a recent event, while a semantic alignment analyzer focuses on narrative-heavy posts. Each module is trained for a narrow and well-defined subtask and can be independently updated or replaced. The third stage is a full LVLM fallback, reserved for rare cases that require comprehensive multimodal reasoning, ensuring that high-cost processing is applied only when necessary. This hierarchical design mirrors the workflow of human fact-checkers, where rapid preliminary screening is followed by more detailed analysis for uncertain or disputed claims. Furthermore, not all inputs demand the same level of processing. Early-exit mechanisms enable the model to produce confident predictions at intermediate layers when sufficient evidence has already been gathered, avoiding the need to engage deeper and more computationally expensive modules. For example, if a meme’s text includes a clearly false statistic and the accompanying image offers no conflicting visual evidence, the system can terminate the analysis early without performing resource-intensive cross-attention over visual features. Similarly, input-adaptive routing can direct different modalities through tailored subnetworks. A text-dominant post (e.g., a quote screenshot) might skip heavy visual encoding, while a video-based claim could activate a dedicated temporal consistency module. Such routing can be learned jointly with the main task or guided by lightweight meta-classifiers. Finally, efficiency must be co-designed with hardware constraints. Techniques such as model quantization (reducing numerical precision from 32-bit to 8-bit), structured pruning (removing redundant attention heads or feed-forward neurons), and knowledge distillation (training small student models to mimic large teachers) can drastically reduce model size and latency without significant accuracy loss. Crucially, these optimizations should preserve the model’s sensitivity to subtle multimodal inconsistencies, avoiding aggressive compression that erases fine-grained evidential signals.

\subsection{Inference-Time Mitigation of Multimodal Hallucination}

Multimodal hallucination represents a critical yet underexplored challenge in LVLM based fake news detection. Unlike general reasoning errors, hallucinated visual entities, events, or cross modal causal relations can artificially increase the plausibility of false claims, misleading both automated detectors and downstream users by providing high confidence yet unfounded evidence. This issue is especially concerning in misinformation scenarios, where models must reason beyond directly observable content. Hallucinated visual or textual content can lead to the generation of false narratives that appear credible, thereby undermining the reliability of detection systems. The problem is compounded by the seamless reinforcement of misleading textual claims through misinterpreted or manipulated visual content, creating more convincing but ultimately false narratives.

While most existing solutions focus on training time alignment or data curation, emerging research indicates that inference time mitigation offers a lightweight, model agnostic alternative to reduce hallucinated reasoning without the need for retraining large scale LVLMs. This approach is particularly advantageous for real world applications, where retraining large models on massive datasets can be computationally expensive and impractical. A prominent example is Instruction Contrastive Decoding (ICD) \cite{wang2024mitigating}, which suppresses generation paths dominated by language priors by explicitly contrasting faithful and unfaithful decoding trajectories. This approach ensures that the model’s output remains grounded in both textual and visual modalities, leading to more accurate visual grounding and reducing the risk of hallucinations. ICD is especially relevant for fake news detection, where even subtle hallucinations such as fabricated visual content or misrepresented events can distort veracity assessments by introducing fictitious evidence. Beyond ICD, innovative methods such as Cogsteer \cite{wang2025cogsteer} have shown considerable promise in mitigating hallucinations during inference. Cogsteer introduces a selective layer intervention mechanism that draws inspiration from cognitive processes of attention and filtering. This method guides the model to prioritize relevant, contextually accurate information while downplaying less reliable or hallucinated details. By intervening at specific layers of the model during inference, Cogsteer effectively steers large models toward more grounded, factually accurate outputs, significantly reducing the likelihood of generating hallucinated cross modal inferences without requiring large scale retraining or additional data curation. The combination of selective layer interventions and contrastive decoding offers a robust strategy for mitigating hallucinations across both visual and textual modalities. While ICD addresses hallucinated textual content, Cogsteer ensures that the model’s attention aligns with relevant visual evidence, thus creating a more holistic solution to hallucination mitigation. These techniques, when combined, form a powerful framework for enhancing the reliability of LVLMs in multimodal misinformation detection tasks. Furthermore, additional inference time mechanisms, such as evidence-conditioned generation  and self-consistency verification, provide complementary controls to enhance cross modal faithfulness. Evidence-conditioned generation verifies the consistency of generated content with available, trusted evidence during the generation process. Self-consistency verification enables models to cross check their outputs against multiple sources or versions of the input, ensuring that the generated content aligns with verified data. While these techniques have yet to be fully explored in the context of multimodal misinformation detection, they offer promising avenues for further reducing hallucination rates and improving model trustworthiness.

Future research should focus on integrating hallucination aware inference strategies such as ICD, Cogsteer, and self-consistency mechanisms into LVLM based fake news detectors. Evaluation metrics should prioritize faithfulness oriented measures that explicitly quantify evidence grounding and hallucination rates, rather than relying solely on end task detection accuracy. These faithfulness metrics would offer a more nuanced evaluation of model performance, particularly in complex real world scenarios where identifying hallucinations is just as critical as detecting false claims. Incorporating such metrics will significantly advance the development of more reliable, robust, and trustworthy LVLM based systems for multimodal fake news detection.

	\subsection{Knowledge-Enhanced LVLM Adaptation for Veracity Reasoning}
	
	The discussion of knowledge enhanced LVLM adaptation remains underdeveloped, despite mounting evidence that integrating contextual world knowledge can substantially enhance multimodal reasoning, particularly in tasks requiring complex, cross modal understanding. Existing frameworks often frame knowledge enhancement narrowly, focusing primarily on external fact retrieval or static knowledge graph augmentation. However, recent advancements suggest that LVLMs can not only leverage structured external knowledge but also dynamically generate and contextualize world knowledge to improve decision-making processes during inference.
	
	A prime example of this is WisdoM  \cite{wang2024wisdom}, which demonstrates the capacity of LVLMs to generate, contextualize, and fuse commonsense and background knowledge with both visual and textual modalities. By actively eliciting relevant contextual knowledge during inference, WisdoM improves multimodal sentiment analysis, especially under ambiguous conditions or when context is sparse. This approach goes beyond static knowledge integration by enabling models to adjust their knowledge based on the immediate reasoning context, leading to more accurate multimodal predictions. This dynamic knowledge generation and fusion approach has strong implications for veracity reasoning in fake news detection. Fake news often presents claims that, while visually plausible, may be contextually implausible. For instance, claims may involve geopolitical contexts or events that require implicit world knowledge, such as an understanding of regional political dynamics, typical event timelines, or physical plausibility. These forms of knowledge are often not directly observable in the visual modality, which makes their integration crucial for accurate veracity assessments. Building on these insights, future research should focus on adapting knowledge enhanced paradigms to veracity reasoning tasks. LVLMs must be able to dynamically incorporate both external and internal knowledge contextually relevant background information that aids in the identification of false claims while maintaining cross modal consistency. Unlike traditional approaches that treat knowledge as an isolated post hoc component, this vision advocates for the seamless integration of knowledge generation and grounding mechanisms throughout the model’s reasoning process. Such a holistic approach would enable LVLMs to not only detect inconsistencies between visual and textual cues but also assess whether a claim, while visually credible, aligns with contextual knowledge, leading to more robust and reliable fake news detection systems.
	
	Knowledge enhanced LVLMs, particularly those capable of dynamic knowledge generation and fusion, present significant potential for advancing multimodal reasoning in fake news detection. By enabling LVLMs to distinguish between visually plausible yet contextually false claims, this approach offers a promising path toward improving veracity assessment, providing a more comprehensive and accurate solution to the challenges posed by multimodal misinformation.

\section{Conclusion}
As one of the fastest-growing areas in artificial intelligence, large visual-language models (LVLMs) for multimodal fake news detection have made significant progress in recent years. Therefore, we provide a comprehensive review of this research area. First, we introduce its background and motivation and highlight the unique challenges posed by LVLMs. Second, we present some preliminary research results, including the definition of multimodal fake news, the evolution of traditional multimodal methods, and LVLM-based methods. Third, we propose a taxonomy of current methods, categorizing them into parameter-frozen paradigm and  parameter-tuning paradigm. Each paradigm provides a complementary perspective on the development of this field. We then review representative models within each paradigm, tracing their development history and methodological innovations. Fourth, We introduce commonly used evaluation metrics in multimodal fake news detection. Fifth, we compare the performance of different methods on various multimodal fake news detection datasets. Sixth, we summarize several commonly used multimodal fake news detection datasets. Finally, we outline unresolved challenges and future research directions.


\bibliographystyle{elsarticle-num}
\bibliography{refs}


%
%
%

\end{document}